\documentclass[final,5p,times,authoryear, twocolumn]{elsarticle}
\usepackage{caption}
\usepackage{subcaption}
\usepackage{amsmath,amssymb,amsfonts}
\usepackage{units}
\usepackage{xcolor}
\usepackage{booktabs} 
\usepackage[switch]{lineno} 
\usepackage[hidelinks,colorlinks = true,urlcolor=blue,linkcolor = blue,citecolor = blue]{hyperref}
\makeatletter
\def\ps@pprintTitle{%
	\let\@oddhead\@empty
	\let\@evenhead\@empty
	\let\@oddfoot\@empty
	\let\@evenfoot\@oddfoot
}
\title{Predictive Digital Twin for Condition Monitoring Using Thermal Imaging}

\begin{document}

\begin{frontmatter}
    \author[label1]{Daniel Menges\fnref{fn1}}
    \ead{daniel.menges@ntnu.no}
    \author[label1]{Florian Stadtmann\fnref{fn1}} 
    \ead{florian.stadtmann@ntnu.no}
    \author[label1]{Henrik Jordheim}
    \ead{henjord@hotmail.com}
     \author[label1,label2]{Adil Rasheed\corref{mycorrespondingauthor}}
    \cortext[mycorrespondingauthor]{Corresponding author}
    \ead{adil.rasheed@ntnu.no}
    \fntext[fn1]{Equal contribution}
    \affiliation[label1]{
    organization={Department of Engineering Cybernetics, NTNU},
            addressline={O.S.Bragstads plass 2}, 
            city={Trondheim},
            postcode={7034},
            country={Norway}
            }
    \affiliation[label2]{
    organization={Mathematics and Cybernetics, SINTEF Digital},
            addressline={Klæbuveien 153}, 
            city={Trondheim},
            postcode={7031},
            country={Norway}
            }
\begin{abstract}
This paper explores the development and practical application of a predictive digital twin specifically designed for condition monitoring, using advanced mathematical models and thermal imaging techniques. Our work presents a comprehensive approach to integrating Proper Orthogonal Decomposition (POD), Robust Principal Component Analysis (RPCA), and Dynamic Mode Decomposition (DMD) to establish a robust predictive digital twin framework. We employ these methods in a real-time experimental setup involving a heated plate monitored through thermal imaging. This system effectively demonstrates the digital twin's capabilities in real-time predictions, condition monitoring, and anomaly detection. Additionally, we introduce the use of a human-machine interface that includes virtual reality, enhancing user interaction and system understanding. The primary contributions of our research lie in the demonstration of these advanced techniques in a tangible setup, showcasing the potential of digital twins to transform industry practices by enabling more proactive and strategic asset management.
\end{abstract}
\end{frontmatter}

\section{Introduction}
\label{sec:introduction}
Digital twin \citep{Rasheed2020dtv} technology is widely regarded as a game-changing innovation, offering the potential to revolutionize how industries manage, monitor, and optimize their assets. With a capability scale \citep{Stadtmann2023dti} ranging from 0-5 (0-standalone, 1-descriptive, 2-diagnostic, 3-predictive, 4-prescriptive, 5-autonomous), digital twins enable deeper integration between physical assets and their virtual counterparts. Among the most valuable types of digital twins for industrial applications are those operating at the diagnostic and predictive levels, which are essential for condition monitoring. These digital twins provide actionable insights, facilitating predictive maintenance and proactive decision-making to mitigate potential failures before they occur.

To ensure that a digital twin remains closely synchronized with its physical counterpart and avoids operational drift, it must be continuously fed with real-time data. This requirement emphasizes the importance of designing assets with the potential digital twin in mind, ensuring they are properly instrumented for seamless data integration. Thermal imaging \citep{Bagavathiappan2013itf,Menges2024rtp}, in particular, offers a promising solution for condition monitoring, given its non-intrusive nature, high-resolution measurements, and cost efficiency. As temperature is often a key indicator of asset health, thermal imaging allows for accurate monitoring without disrupting system operations.

However, thermal imaging generates highly complex, high-dimensional datasets, making it necessary to employ computationally efficient algorithms to process the data and extract meaningful insights. A shortage of such high-dimensional datasets has historically hindered the validation of new algorithms for digital twin applications. While many recent methods have been proposed, several rely heavily on black-box models \citep{Sandhu2024acs,Jiang2024aob,Belay2024mmo}, such as neural networks, which can lack transparency and interpretability, posing challenges in high-stakes environments where understanding the decision-making process is critical. To address this aspect of modeling, recent works \citep{Menges2024rtp,Belay2024mts,Menges2024cam} have built upon methods with a robust mathematical foundation. The current work combines two such methods \citep{Menges2024rtp,Menges2024cam} originally tested independently on offline data and integrates them into a unified online digital twin framework for condition monitoring applied to a heated plate monitored using thermal imaging. The results are communicated using a human machine interface consisting of virtual reality. The main contributions of this work are: 
\begin{itemize}
    \item the development of a digital twin-ready extendable physical framework for condition monitoring based on high dimensional thermal imaging.
    \item the creation and demonstration of a corresponding extendable real-time diagnostic and predictive digital twin powered by predictive models based on a robust mathematical foundation.
    \item blurring the distinction between the physical and the digital twin frameworks using a virtual reality interface.
\end{itemize} 

These contributions highlight a practical step forward from theoretical concepts to real-world applications, advancing proactive and strategic asset management. While this work demonstrates the effectiveness of digital twin technology, it also lays the groundwork for future developments, with the potential to improve industrial operations.

The structure of this paper systematically addresses the development and implementation of a predictive digital twin for condition monitoring. Section \ref{sec:theory} presents the required theories to understand the work. Section \ref{sec:dtreadyphysicalasset} presents the development of the physical framework for condition monitoring. Section \ref{sec:methodology} includes all the information required to reproduce the content presented in this work. Results and Discussions are presented in Section \ref{sec:results}. Finally, the paper concludes (Section \ref{sec:conclusionandfuturework}) with a summary of the key contributions and outlines future work aimed at enhancing the digital twin framework and extending its applicability to other use cases.

\section{Theory}
\label{sec:theory}
This section begins with a clear definition of a digital twin, outlining its role in mirroring physical systems to enhance real-time operational capabilities and predictive maintenance across various industries. Following this foundational understanding, we delve into the mathematical and computational methodologies that underpin our predictive digital twin. This includes an exploration of key techniques such as Proper Orthogonal Decomposition (POD), Robust Principal Component Analysis (RPCA), and Dynamic Mode Decomposition (DMD). The methodologies and the required theories for this study follow the approach outlined in \cite{Menges2024rtp}. These methods are crucial for accurately modeling complex behaviors and ensuring the reliability and efficiency of the digital twin in practical applications.

    \subsection{Digital twin and capability level}
\begin{figure*}[!ht]
    \centering
    \includegraphics[width=\linewidth]{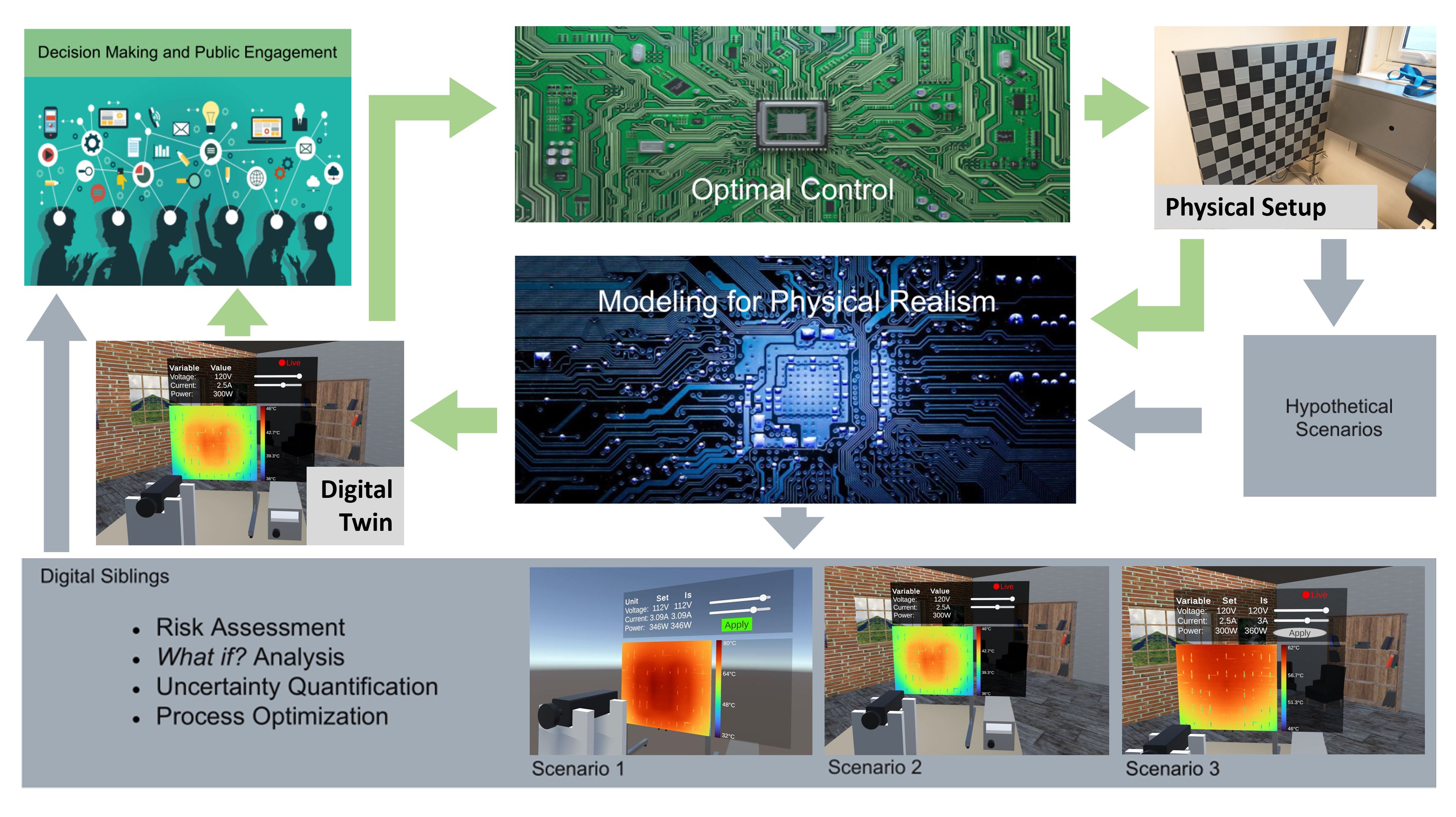}
    \caption{Concept of a digital twin and digital sibling. The physical asset is shown in the top right. Data from the asset is collected and enhanced with models (middle) to instill physical realism into the digital twin (middle left). The digital twin can be used for decision-making and public engagement (top left) and for optimal control (top middle). By devising and modeling hypothetical scenarios, risk assessment, what-if analysis, uncertainty quantification, and process optimization can be performed. This is often referred to as a digital sibling. Green arrows represent real-time actions and information flow, while grey arrows and boxes may be executed offline.}
    \label{fig:dt1}
    \end{figure*}

    As mentioned earlier digital twin is a virtual replica of a physical asset, enabled through data and simulations, that can be used for real-time monitoring, optimization, and decision-making \citep{Rasheed2020dtv}. The motivation to study DTs is rooted in the potential cost savings and efficiency gains they offer. In Fig.~\ref{fig:dt1}, we can see the concept of a digital twin. The physical asset is located in the top right side of the figure, equipped with various sensors that provide real-time big data. However, this data has limited spatio-temporal resolution and does not tell about the future state of the asset. To complement the measurement data, models are used to create a digital representation of the asset. If the digital twin can provide the same information as the physical asset, it can be utilized for informed decision-making and optimal control. The green arrows in the figure show real-time data exchange and analysis. To perform risk assessment, what-if analysis, uncertainty quantification, and process optimization, the digital twin can be run in an offline setting for scenario analysis. It is then called digital siblings. The gray box and arrows represent the digital sibling. Additionally, the digital twin predictions can be archived during the lifetime of the asset and can be used for designing a next generation of assets, in which the concept is referred to as digital threads.
    \begin{figure*}
    \centering
    \includegraphics[width=\linewidth]{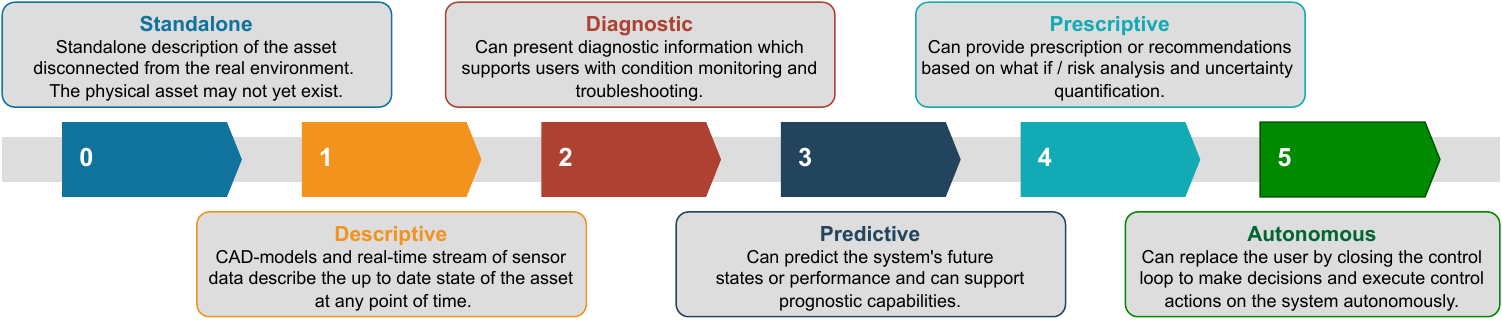}
    \caption{Description of capability levels of a digital twin.}
    \label{fig:cap_level_framework}
    \end{figure*}
    \cite{San2021haa} and \cite{Stadtmann2023dif} present a digital twin capability level scale adapted from a DNV GL report \citep{GL2020dra} that divides a DT into six distinct levels. These levels are 0-Standalone, 1-Descriptive, 2-Diagnostic, 3-Predictive, 4-Prescriptive and 5-Autonomous (Fig. \ref{fig:cap_level_framework}). A standalone DT can exist even before the asset is built and can consist of solid models. When the asset is in place and is equipped with sensors, data can be streamed in real-time to create a descriptive digital twin, giving more insight into the state of the asset. When analytic tools are applied to the incoming data stream to diagnose anomalies, the digital twin advances to a diagnostic level. At the first three levels, the digital twin can provide information only about the past and present. However, a predictive digital twin can describe the future state of the asset. Using the predictive digital twin, one can do scenario analysis to provide recommendations to push the asset to the desired state. This is then referred to as prescriptive level. Lastly, the asset updates the digital twin at the autonomous level, and the digital twin controls the asset autonomously. 
    
\subsection{Proper Orthogonal Decomposition} \label{POD}
    Proper Orthogonal Decomposition (POD) is a method used to simplify complex systems by reducing their dimensions and focusing on the most important features. It identifies a set of orthogonal (mutually independent) modes that efficiently capture the main characteristics of the system. 
    The POD modes can be extracted using Singular Value Decomposition (SVD) \citep{Chatterjee2000AnIT}.    
    Given a data matrix $\mathbf{X}\in \mathbb{R}^{n\times k}$, the SVD of this matrix leads to
    \begin{equation}
        \mathbf{X} = \mathbf{U}\mathbf{\Sigma}\mathbf{V}^\top,
    \end{equation}
    where $\mathbf{U}\in \mathbb{R}^{n\times n}$, $\mathbf{\Sigma}\in \mathbb{R}^{n\times k}$, and $\mathbf{V}\in \mathbb{R}^{k\times k}$. The diagonal elements of $\mathbf{\Sigma}$ represent the singular values, $\Sigma_{ii}$, arranged in descending order.
    A reduced representation of the original data is captured by a small set of orthonormal eigenmodes $\mathbf{\Psi}_r$ (POD modes) \citep{Manohar2018}. This is achieved through a lower-rank SVD approximation, expressed as
    \begin{equation}
        \mathbf{X}\approx \mathbf{U}_r\mathbf{\Sigma}_r\mathbf{V}^\top_r. \label{eq:lower_rank_SVD}
    \end{equation}
    The matrices are defined as $\mathbf{U}_r \in \mathbb{R}^{n\times r}$, $\mathbf{\Sigma}_r \in \mathbb{R}^{r\times r}$, and $\mathbf{V}_r^\top~\in~\mathbb{R}^{r\times k}$, where $r$ represents the reduced dimensionality. The POD modes are then given by $\mathbf{\Psi}_r = \mathbf{U}_r$.

\subsection{Robust Principal Component Analysis} \label{sec:RPCA}
    Robust Principal Component Analysis (RPCA) operates by decomposing the data matrix into two components: a low-rank matrix and a sparse matrix. The low-rank matrix represents the underlying dynamics of the data, while the sparse matrix captures outliers or anomalies  \cite{Menges2024cam}. This decomposition is particularly valuable in applications such as image and video processing, where the low-rank component often corresponds to the background, and the sparse component highlights moving objects or irregularities.
    The core concept involves decomposing the data matrix $\mathbf{X}$ into two distinct components, represented as
    \begin{equation}
        \mathbf{X} = \mathbf{L} + \mathbf{S},
    \end{equation}
    where the matrix $\mathbf{L}$ represents the low-rank component, encapsulating the primary structure of the data, while the matrix $\mathbf{S}$ is sparse, identifying outliers and irregularities. Hence, the objective is to determine the matrices $\mathbf{L}$ and $\mathbf{S}$ that fulfill the condition
    \begin{equation}
    \begin{split}
    & \underset{\mathbf{L}, \mathbf{S}}{\text{minimize}} \hspace{0.5cm}\mathrm{rank}(\mathbf{L}) + \|\mathbf{S}\|_0, \\
    & \text{subject to} \hspace{0.5cm} \mathbf{L} + \mathbf{S} = \mathbf{X},
    \end{split}
    \label{eq:RPCA_ideal}
    \end{equation}
    where $|\mathbf{S}|_0$ represents the zero norm of $\mathbf{S}$, indicating the number of nonzero elements, and $\mathrm{rank}(\mathbf{L})$ denotes the rank of the matrix $\mathbf{L}$. However, the nonconvexity of both $\mathrm{rank}(\mathbf{L})$ and $|\mathbf{S}|_0$ makes this optimization problem computationally intractable \citep{Scherl2020rpc}. To address this challenge, convex relaxation techniques \citep{Zhang2010aom} offer a means to approximate nonconvex problems by transforming them into convex ones. This approach allows the reformulation of Eq.~\eqref{eq:RPCA_ideal} into a more tractable optimization problem given by
    \begin{equation}
    \begin{split}
    & \underset{\mathbf{L}, \mathbf{S}}{\text{minimize}} \hspace{0.5cm}\|\mathbf{L}\|_* + \lambda \|\mathbf{S}\|_1, \\
    & \text{subject to} \hspace{0.5cm} \mathbf{L} + \mathbf{S} = \mathbf{X}.
    \end{split}
    \label{eq:PCP}
    \end{equation}
    The notation $\|\cdot\|_1$ is the $L_1$ norm, defined as the sum of the absolute values of the matrix entries, while $\|\cdot\|_*$ denotes the nuclear norm, defined as the sum of the singular values. The hyperparameter $\lambda$ controls the balance between the two terms. Minimizing $\|\mathbf{S}\|_1$ approximates the minimization of $\|\mathbf{S}\|_0$, promoting sparsity in $\mathbf{S}$, while minimizing $\|\mathbf{L}\|_*$ serves as an approximation for reducing the rank of $\mathbf{L}$, encouraging a low-rank structure. The formulation given in Eq.~\eqref{eq:PCP} represents a convex optimization problem, commonly referred to as Principal Component Pursuit (PCP). To address this convex optimization problem, the Augmented Lagrange Multiplier (ALM) algorithm provides a feasible approach \citep{lin_augmented_2010}. The ALM can be formulated as
    \begin{equation}
        \hspace{-0.7em}\resizebox{.93\hsize}{0.028\vsize}{$\mathcal{L}(\mathbf{L},\mathbf{S},\mathbf{\Lambda})=\|\mathbf{L}\|_* + \lambda \|\mathbf{S}\|_1+\langle \mathbf{\Lambda}, \mathbf{X} - \mathbf{L} - \mathbf{S} \rangle + \frac{\mu}{2}\|\mathbf{X}-\mathbf{L}-\mathbf{S}\|_{F}^2$}. \label{eq:ALM}
    \end{equation}
    In this formulation, $\mathbf{\Lambda}$ represents the matrix of Lagrange multipliers, $\mu$ denotes a hyperparameter, $\langle \cdot \rangle$ indicates the inner product, and $|\cdot|_F$ refers to the Frobenius norm (also known as the Euclidean norm), which quantifies the magnitude or length of a matrix. The expression $\mathcal{L}$ is minimized to determine $\mathbf{L}_k$ and $\mathbf{S}_k$ at timestep $k$, while the matrix of Lagrange multipliers is updated by
    \begin{equation}
        \mathbf{\Lambda}_{k+1} = \mathbf{\Lambda}_{k} + \mu(\mathbf{X}-\mathbf{L}_k-\mathbf{S}_k).
    \end{equation}
    Consequently, RPCA separates a data matrix $\mathbf{X}$ into a low-rank component $\mathbf{L}$ and a sparse component $\mathbf{S}$.
    
    \subsection{Optimal Sampling Location}\label{subsec:osp}
    Optimal Sampling Location (OSL) helps to identify the most informative points for data collection within a system, enhancing measurement accuracy and precision while minimizing the number of required sensors.
    It is expected that a single data frame $\boldsymbol{x} \in \mathbb{R}^n$ at a specific time can be approximated by
    \begin{equation}
        \boldsymbol{x} \approx \mathbf{\Psi}_r 	\boldsymbol{a},
    \end{equation}   
    where $\boldsymbol{a} \in \mathbb{R}^{r}$ denotes the time-varying coefficients. 
    Under the assumption that the optimal locations can be expressed by
    \begin{equation}
        \boldsymbol{y} = \mathbf{C}\boldsymbol{x},
    \end{equation}
    where $\mathbf{C} \in \mathbb{R}^{s \times n}$ denotes a sparse measurement matrix and $s$ indicates the number of samples, the values of $\boldsymbol{y}$ can be infererd by
    \begin{equation}
        \boldsymbol{y} \approx \mathbf{C}\mathbf{\Psi}_r \boldsymbol{a}.
    \end{equation}
    By defining $\mathbf{\Theta} = \mathbf{C}\mathbf{\Psi}_r$, the estimation of $\boldsymbol{a}$ can be obtained by
    \begin{equation}
        \boldsymbol{\hat{a}} = \mathbf{\Theta}^\dagger\boldsymbol{y}. \label{eq:a_est}
    \end{equation}
    Consequently, an estimation of $\boldsymbol{x}$ can be reconstructed using
    \begin{equation}
        \boldsymbol{\hat{x}} = \mathbf{\Psi}_r\boldsymbol{\hat{a}} = \mathbf{\Psi}_r(\mathbf{C}\mathbf{\Psi}_r)^\dagger\boldsymbol{y}.
    \end{equation}
    Given that $\mathbf{\Psi}_r$ is obtained from the lower-ranked SVD, the only unknown is the sparse measurement matrix $\mathbf{C}$. As stated in \cite{Manohar2018}, an optimal sensor placement can be realized by utilizing QR factorization with column pivoting on the POD modes $\mathbf{\Psi}_r$, with the condition that $s \geq r$.

    \subsection{Dynamic Mode Decomposition}
    Dynamic Mode Decomposition (DMD) offers a data-driven method for extracting coherent spatio-temporal patterns and capturing system dynamics through a linear model represented by
    \begin{equation}
        \mathbf{X}'=\mathbf{\Phi}\mathbf{X}, \label{eq:DMD_1}
    \end{equation}
    where $\mathbf{\Phi}$ expresses the transition matrix.
    To forecast future states, it is essential to first construct a time-shifted dataset $\mathbf{X}'$ based on
    \begin{align}
        &\mathbf{X} = [\boldsymbol{x}_0 \cdot \cdot \cdot \boldsymbol{x}_{w}],\\
        &\mathbf{X}' = [\boldsymbol{x}_l \cdot \cdot \cdot \boldsymbol{x}_{w+l}],
    \end{align}
    where $w$ represents the selected window size and $l$ indicates the number of future prediction steps.
    Considering a lower-rank SVD as given by Eq.~\eqref{eq:lower_rank_SVD}, the linear model Eq.~\eqref{eq:DMD_1} can be rearranged to approximate the transition matrix by computing
    \begin{equation}
        \mathbf{\tilde{\Phi}} = \mathbf{U}^\top_{r}\mathbf{X}'\mathbf{V}_{r}\mathbf{\Sigma}^{-1}_{r}. \label{eq:DMD_start}
    \end{equation}
    The matrix $\mathbf{\tilde{\Phi}}$ represents a linear approximation of a dynamical system characterizing the time evolution of the POD modes. Subsequently, the eigenvalues $\lambda$ of $\mathbf{\tilde{\Phi}}$ with $\mathbf{\Lambda} = \mathrm{diag}(\lambda_1 \cdot \cdot \cdot \lambda_r)$ and the eigenvectors $\boldsymbol{v}$ of $\mathbf{\tilde{\Phi}}$ \citep{Tu2014} are determined to compute
    \begin{align}
        \mathbf{P} &= \mathbf{X}'\mathbf{V}_{r}\mathbf{\Sigma}^{-1}_{r}\boldsymbol{v},\\
        \mathbf{\Phi} &= \mathbf{P}\mathbf{\Lambda}\mathbf{P}^\dagger.
    \end{align}
    The predicted states with $l$ number of prediction steps are finally given by
    \begin{equation}
        \mathbf{X}_{\mathrm{pred}} = \mathbf{\Phi}\mathbf{X}. \label{eq:DMD_end}
    \end{equation}
    However, applying DMD requires significant computational power, making real-time application on large datasets challenging.
    
    \subsection{Support Vector Regression}\label{subsec:svm} 
    The objective of Support Vector Regression (SVR) is to find a function
    \begin{equation}
        f(\boldsymbol{x}) = \boldsymbol{\theta}^\top \boldsymbol{x}+\theta_{o}, 
    \end{equation}
    minimizing the prediction error $|y - f(\boldsymbol{x})|$, where $\boldsymbol{\theta}$ represents a weight vector, $\theta_{o}$ denotes a bias term, and $\boldsymbol{y}$ characterizes the target data. 
    In this context, a linear kernel is utilized. Nonetheless, the kernel trick enables nonlinear mappings into higher-dimensional spaces without significantly increasing computational expenses. Consequently, the standard inner product can be substituted with any nonlinear kernel \citep{Yan2022}. 
    For a set of training vectors $\boldsymbol{x_i}$ where $i=1,...,q$, the optimization problem is expressed as
    \begin{equation}\label{eq:safety-filter}
    \begin{aligned}
        \min_{\boldsymbol{\theta},\theta_{o}} &\left(\boldsymbol{\theta}^\top \boldsymbol{\theta}+c\sum_{i=1}^{q}(\xi_i+\xi_i')\right) \\  
        \text{s.t. \ \ \ }
        &y_i - f(\boldsymbol{x}_i)\leq \varepsilon + \xi_i \\
        &f(\boldsymbol{x}_i)-y_i\leq \varepsilon + \xi_i' \\
        &\xi_i, \ \xi_i' \geq 0, \ \text{for} \ \  i = 1,...,q
    \end{aligned}
    \end{equation}
    Here, $c > 0$ acts as a penalty parameter, $\varepsilon$ defines the margin within which deviations are acceptable, and $\xi_i$ and $\xi_i'$ are slack variables that permit certain training points to exceed the margin constraints. 
    Quadratic programming is employed to solve the optimization problem, producing the weight vector $\boldsymbol{\theta}$ and the bias term $\theta_{o}$ that establish the final model.
    The weight vector is obtained by a weighted sum of the support vectors $\boldsymbol{x}_i$, the target values $y_i$, and the corresponding Lagrange multipliers $\alpha_i$, given by
    \begin{equation}
        \boldsymbol{\theta}=\sum_{i=1}^{q} \alpha_i y_i \boldsymbol{x}_i.
    \end{equation}
    By applying a nonlinear kernel function $K(\boldsymbol{x}, \boldsymbol{x}_i)$, the regression model finally yields
    \begin{equation}
        f(\boldsymbol{x}) = \sum_{i=1}^{q} \alpha_i y_i  \boldsymbol{\theta}^\top K(\boldsymbol{x},\boldsymbol{x}_i)+\theta_{o}.
    \end{equation}

\section{Digital twin ready physical asset}
\label{sec:dtreadyphysicalasset}
Digital twins use real-time measurements to synchronize the virtual representation with the physical asset. A physical asset is prepared for the digital twin through the installation of sensors and devices for data accumulation. Here, the experimental setup and data pipeline are described.
    
    \subsection{Experimental setup}
\begin{figure}[!ht]
    \centering
    \includegraphics[width=\linewidth]{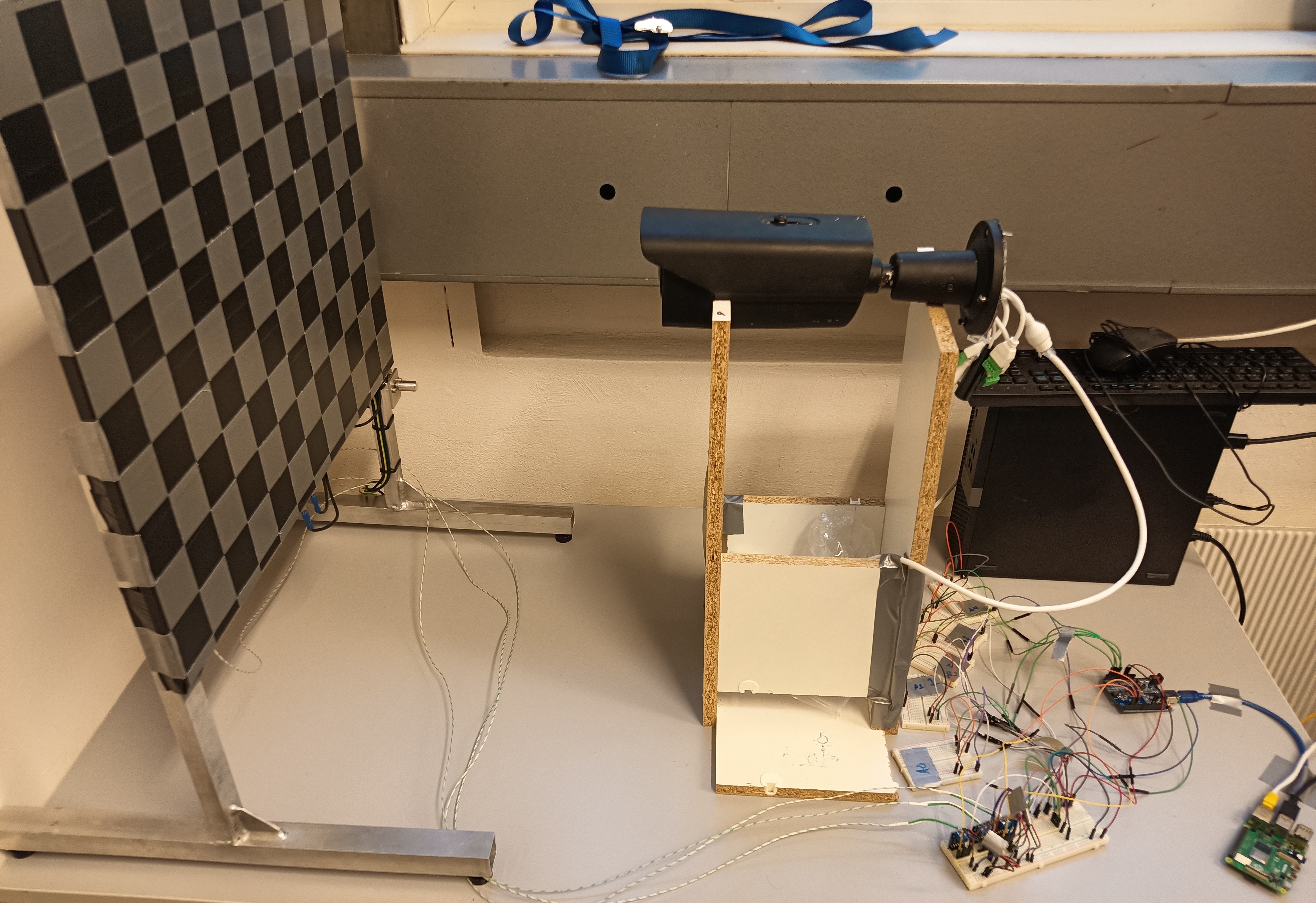}
    \caption{Setup: A back heated aluminum plate observed using a thermal camera}
    \label{fig:SetupSideView}
    \end{figure}

    The experimental setup is shown in Fig.~\ref{fig:SetupSideView}. The asset consists of a $\unit[60.5]{cm}\times \unit[52.5]{cm}$ rectangular aluminum plate with thickness $\unit[1]{cm}$, which is heated from the backside using a heating coil powered with a remote-controlled programmable power supply. A thermal camera is pointed at the front side of the aluminum plate to measure the temperature distribution on the plate in real-time without disturbing the system. The camera used in the setup is a Hikvision DS-2TD2166-7/V1 Thermal Network Bullet Camera that features a measurement range of -20 to 150 $^\circ$C at $640\times512$ pixel resolution and supports remote operation through the internet protocol. The thermal camera captures a wide field of view with 88.5$^\circ \times 73.2^\circ$ through a fisheye lens. The camera temperature measurements are calibrated using three thermocouples that are mounted on the aluminum plate and operated through an Arduino Mega board. Five thermistors are connected to the Arduino board to measure the temperature of the environment. A Dell OptiPlex 7040 x64-based desktop computer with an Intel i7-6700 CPU, 32 GB RAM, and 162 GB of storage capacity is used as an edge device to collect data from the different components of the setup.  
    
    \begin{figure}[b!]
    \centering
    \includegraphics[width=\linewidth]{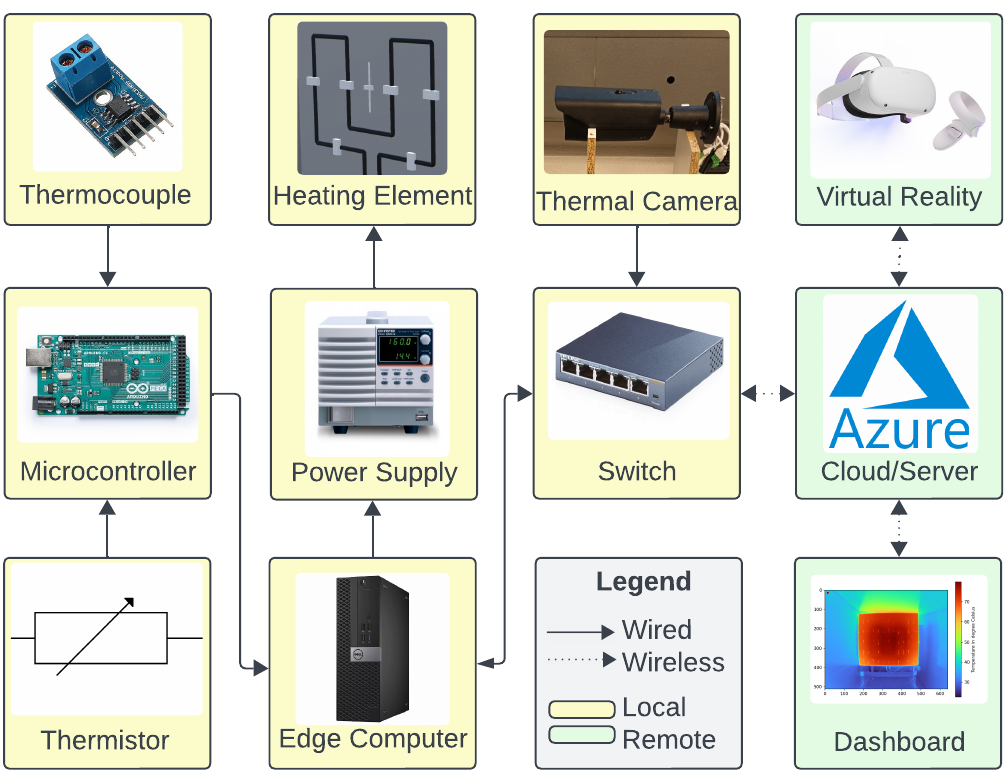}
    \caption{The data flowchart for physical setup and digital twin. All components can be monitored and controlled remotely through a dashboard and a virtual reality application.}
    \label{fig:HW-network}
    \end{figure}
    
    \subsection{Data pipeline}
    The data pipeline is sketched in Fig.~\ref{fig:HW-network}. A programmable power supply and an Arduino board are connected via USB to a local desktop computer, acting as an edge device. The edge device controls both the voltage and current of the programmable power supply, as well as the Arduino. This setup enables the Arduino to sample voltage from its I/O pins, indirectly measuring the temperature of thermocouples and thermistors. The open-source Arduino integrated development environment is used to program the setup, with the sequence loop sampling data from the thermistors and thermocouples at \unit[2]{Hz}. To measure the temperature at the thermistors, the Steinhart-Hart equation is implemented. For the thermocouples, temperature is measured using the Adafruit MAX31855 library from the Arduino library manager. 
    A network switch establishes a subnet connecting the thermal camera and the edge device via ethernet, forming part of a larger local private network. The thermal camera captures the temperature distribution on the aluminum plate at \unit[20]{Hz}, with raw images at a resolution of $640\times512$ pixels, which are then forwarded to the edge device. For further processing, the area of interest is cropped, reducing the image dimensions to $260\times300$ pixels.

    The storage space of the edge device is limited, but the data is continuously pushed to the cloud and can be accessed on-demand by devices connected to the local network through a virtual private network. There, the data can be utilized to continuously update a predictive model designed for condition monitoring of the system.  Additionally, the infrastructure allows for real-time data streaming to a VR headset, enabling immersive visualization of the system's thermal state. A dashboard is also available for on-demand data analysis, offering a user-friendly interface for more detailed and customized assessments of the experimental results.

    \subsection{Experiments}
    The remote access infrastructure was used to conduct a number of experiments. A data set describing the normal operation of the heatplate was constructed in the following way. Starting with the aluminum plate at room temperature, the heating element was powered at a constant voltage of \unit[120]{V} until the temperature distribution on the aluminum plate reached an equilibrium while recording thermal images at \unit[3.5]{s} intervals. Once the equilibrium was reached, the heating element was turned off, and the cooling process of the aluminum plate was recorded at the same interval until the temperature of the plate returned to room temperature. The experiment was repeated with different voltages ranging from \unit[10]{V} to \unit[120]{V} at \unit[10]{V} intervals. The distortions of the recorded images from the fisheye optic were removed with a pre-calibrated correction matrix, and the images were cropped to only include the aluminum plate. The resulting data set was used as training data for the normal operation of the setup. A second data set was created in a similar fashion, but the voltage was set to \unit[85]{V}, which is not included in the training data, and during the recording of the experiment, three anomalies were introduced to the system in two ways. Water was splashed onto the back of the aluminum plate twice, and a piece of metal was fixed to the back side of the aluminum plate for 20 minutes.

    \subsection{Interface for remote monitoring and control of the asset} \label{ssec:interface}
    A Computer-Aided Design (CAD) model of the experimental setup was initially developed to communicate the concept of the condition monitoring test rig, serving as a visual representation before the physical setup was constructed. At this early stage, the CAD model functioned as a standalone digital twin, allowing stakeholders to understand the design and functionality of the rig in the absence of the actual hardware. This virtual model was instrumental in refining the overall concept and ensuring alignment among project members. As the project progressed, the CAD model evolved to meet higher capability levels, particularly for its integration into virtual reality (VR) environments. 
    The enhanced model enables real-time visualization of the asset's current state and facilitates projections of the asset's future condition through VR, offering an immersive, interactive experience for the user. The integration of the VR interface also allows for direct control of the physical setup through the VR headset, enabling users to interact with and manage the system in an intuitive and engaging way. In addition to VR-based control, the experimental setup can also be monitored and managed via a more conventional dashboard interface, developed using the Plotly Dash library in Python. This dashboard, accessible from a computer, provides an alternative means of system control and condition monitoring, making the interface adaptable to various user preferences and operational needs. 
    The concept of VR is not recent and can be defined as a model of the real world that is maintained in real-time, sounds, and feels real with the possibility to directly and realistically manipulate the environment. Today, there exist many affordable VR solutions, generally consisting of a headset and complimentary controllers, that either utilize their own internal hardware or external processing power to render the virtual environment. Such a representation of a virtual environment through the usage of VR hardware compliments very well the visualization of a DT asset, which allows for a more realistic representation and feel of the asset.
    
\section{Methodology}
\label{sec:methodology}
The core algorithms for real-time predictive condition monitoring are adopted from \cite{Menges2024rtp} and briefly described below. 

    \subsection{Preprocessing framework}
    In this study, a target sampling rate of \unit[3.5]{s} was chosen, which is suitably small for slow-varying processes like the thermodynamics of a heating plate. However, due to signal delays and other factors, including network latency, the sampling intervals were not entirely consistent. Given the common occurrence of irregular time intervals in time series data, interpolation was applied to standardize the dataset.
    The data collected for this study contained minor noise but were not corrupted. However, when data cleaning is necessary, such as for offline analysis or data-driven modeling, RPCA can be a highly effective tool \citep{Menges2024cam}. It can also filter out noise for more accurate modeling of a system's true dynamics. In this work, only POD followed by OSL was applied for data preprocessing and modeling since OSL inherently possesses denoising capabilities within its reconstructions. Nevertheless, the performance of RPCA is additionally demonstrated in Section~\ref{sec:results} to showcase its effectiveness.
    Moreover, data-driven models of the few optimal sampled pixels obtained from OSL are derived by SVR using nonlinear Gaussian kernels. Since we rely on the optimally sampled locations, these additional models can then substitute a measurement in case an erroneous sample is detected.
 
    \subsection{Anomaly detection framework}
    \label{sec:anomaly_detection}
    The anomaly detection approach is based on OSL, utilizing the first three POD modes of the trained data. Given $p=78,000$ pixels per frame, it is assumed that the original image can be accurately reconstructed by sampling only three specific locations. For an image sample $\boldsymbol{x} \in \mathbb{R}^p$ and its reconstructed version $\boldsymbol{\hat{x}} \in \mathbb{R}^p$ from a few pixel measurements, the reconstruction error is defined as
    \begin{equation}
    \boldsymbol{e}=\boldsymbol{\hat{x}}-\boldsymbol{x}.
    \end{equation}
    To detect anomalies, the mean of the $m$ largest values in the error image $\boldsymbol{e}$ is calculated, expressed as
    \begin{equation}
    \bar{\boldsymbol{e}}_{\mathrm{max,m}} = \frac{1}{m}\sum_{i=1}^{m} \max(\boldsymbol{e}_{\mathrm{abs}},i).
    \end{equation}
    Here, $\boldsymbol{e}_{\mathrm{abs}} = \begin{bmatrix} |e_1|, & |e_2|, & \dots & |e_n| \end{bmatrix}^\top$ represents the vector of absolute values of the elements in $\boldsymbol{e}$, and $\max(\boldsymbol{e}_{\mathrm{abs}},i)$ denotes the $i$-th largest value within $\boldsymbol{e}_{\mathrm{abs}}$.
    If $\bar{\boldsymbol{e}}_{\mathrm{max,m}}$ exceeds a predefined threshold $\gamma_1$, the area of interest is identified by locating pixels where the reconstruction error surpasses this threshold. We define the set $\mathcal{S}$, which contains the indices of these pixels, corresponding to values in $\boldsymbol{e}_{\mathrm{abs}}$ that exceed $\gamma_1$, as follows
    \begin{equation}
    \mathcal{S} = \left\{ i \mid e_i \in \boldsymbol{e}_{\mathrm{abs}}, e_i > \gamma_1 \right\}. \label{eq:anomaly_set}
    \end{equation}
    The set $\mathcal{S}$ is a subset of the full image, representing the pixels that have been identified as anomalies. An additional mechanism for detecting rapid pixel changes is implemented using the temporal gradient of $\bar{\boldsymbol{e}}_{\mathrm{max,m}}$ given by
    \begin{equation}
        \nabla \bar{\boldsymbol{e}}_{\mathrm{max,m}} = \frac{\partial \bar{e}_{\mathrm{max,m}}}{\partial t}.
    \end{equation}
    For a discrete time approximation, the gradient can be expressed as
    \begin{equation}
    \nabla \bar{\boldsymbol{e}}_{\mathrm{max,m}} \approx \frac{\bar{e}_{\mathrm{max,m}}(t) - \bar{e}_{\mathrm{max,m}}(t-\Delta t)}{\Delta t},
    \end{equation}
    where $\bar{e}_{\mathrm{max,m}}(t)$ represents the value at time $t$, $\bar{e}_{\mathrm{max,m}}(t-\Delta t)$ is the value from the previous time step, and $\Delta t$ is the time step interval. This allows for identifying sudden shifts in individual pixels, which typically deviate from the slower dynamics observed in processes like the thermodynamics of a heating plate.
    To enhance this gradient estimation, a weighted moving average (WMA) is employed to smooth peaks due to noise. The WMA is defined as 
    \begin{equation}
    \bar{e}_{\mathrm{max,m}}^{\mathrm{WMA}}(t) = \frac{\sum_{i=1}^{N} i \cdot \bar{e}_{\mathrm{max,m}}(t - (N-i) \Delta t)}{\sum_{i=1}^{N} i},
    \end{equation}
    where $N$ specifies the considered window size.
    This adaptive mechanism improves sensitivity to rapid changes by focusing on more recent pixel values while suppressing the effects of noise. 
    The final expression for detecting this type of anomaly is formulated using the absolute values of the gradient of the WMA, resulting in
    \begin{equation}
    \left|\nabla \bar{\boldsymbol{e}}_{\mathrm{max,m}}^{\mathrm{WMA}}\right| = \left|\frac{\partial \bar{e}_{\mathrm{max,m}}^{\mathrm{WMA}}}{\partial t}\right| \approx \left|\frac{\bar{e}_{\mathrm{max,m}}^{\mathrm{WMA}}(t) - \bar{e}_{\mathrm{max,m}}^{\mathrm{WMA}}(t-\Delta t)}{\Delta t}\right|.
    \end{equation}
    If this value exceeds a predefined threshold $\gamma_2$, the detection algorithm is activated, identifying the pixels classified as anomalies as defined in Eq.~\eqref{eq:anomaly_set}.

    \subsection{Anomaly prediction framework}
    DMD forms the basis of the prediction framework. However, applying DMD to the full-dimensional image data results in computation times that exceed acceptable limits, making it unsuitable for real-time applications.
    
        \subsubsection{State Prediction}
        To extract hidden dynamics, we assume that they are represented by the time-varying coefficient vector $\boldsymbol{a}$, as described in Section~\ref{sec:theory}. Rather than applying DMD to the full-dimensional image data $\mathbf{X} \in \mathbb{R}^{n\times w}$ with window size $w$, we apply DMD to the lower-dimensional time-varying coefficient vector $\boldsymbol{a}$, stacked over the window size $w$. This stacked matrix is denoted as $\mathbf{A} \in \mathbb{R}^{r\times w}$. 
        
        Once the POD modes $\mathbf{\Psi}_r$ and the optimal sampling locations $\mathbf{C}$ are determined, SVD is performed on $\mathbf{A}$. Subsequently, DMD is computed according to Eq.~\eqref{eq:DMD_start}-Eq.~\eqref{eq:DMD_end}, resulting in
        \begin{equation}
        	\mathbf{A}_{\mathrm{pred}} = \mathbf{\Phi}\mathbf{A}.
        \end{equation}
        Finally, the estimated predicted images are given by
        \begin{equation}
            \mathbf{X}_{\mathrm{OSL,pred}} = \mathbf{\Psi}_r\mathbf{A}_{\mathrm{pred}}.
        \end{equation} 
        
        \subsubsection{Anomaly Prediction}
        DMD is applied in parallel to the subset $\mathcal{S}$ when an anomaly is detected. This approach models the underlying dynamics of the anomalies, enabling predictive capabilities by projecting them into the future. For clarity, the anomaly predictions for subset $\mathcal{S}$ are denoted as $\mathbf{X}_{\mathcal{S},\mathrm{pred}}$.
            
        \subsubsection{Prediction Framework}
        Accurate predictions of the actual states, which encompass the modeled dynamics through OSL as well as the dynamics of detected anomalies, are achieved by substituting the predictions for the subset $\mathcal{S}$ with the corresponding positions of the predicted images $\mathbf{X}_{\mathrm{OSL,pred}}$. This results in the following expression
        \begin{equation}
        \mathbf{X}_{\mathrm{pred},i} = 
        \begin{cases} 
          \mathbf{X}_{\mathrm{OSL,pred}} & \text{if } i \notin \mathcal{S}, \\
          \mathbf{X}_{\mathcal{S},\mathrm{pred}} & \text{if } i \in \mathcal{S},
        \end{cases} \label{eq:predictino_framework}
        \end{equation}
        where $i$ denotes the pixel index of an entire image.
    
    \subsection{Hyperparameter settings}
    This section introduces the hyperparameters utilized in the model training process and the real-time application, given in Table \ref{tab:hyperparameters}. It includes the choice of tuning parameters for RPCA, POD, OSL, SVR, and the methodologies introduced in this section.
    \begin{table}[h!]
        \centering
        \caption{Hyperparameters and their tuned values for the applied algorithms.}
        \begin{tabular}{@{}lll@{}}
            \toprule
            \textbf{Algorithm} & \textbf{Symbol} & \textbf{Value} \\ \midrule
            RPCA  & $\lambda$ & 0.001\\
            RPCA  & $\mu$ & $10^{-5}$\\
            POD  & $r$ & 3\\
            OSL  & $s$ & 3\\
            SVR  & $c$ & 1\\
            SVR  & $\epsilon$ & 0.1\\
            Anomaly detection  & $m$ & 100\\
            Anomaly detection  & $\gamma_1$ & 1\\
            Anomaly detection  & $\gamma_2$ & 0.01\\
            Anomaly detection (WMA)  & $N$ & 100\\
            Anomaly detection & $\Delta t$ & \unit[3.5]{s} \\
            State prediction (DMD)  & $l$ & \{100, 300\}\\
            State prediction (DMD)  & $w$ & \{100, 200, 300\}\\
            Anomaly prediction (DMD)  & $l$ & 100\\
            Anomaly prediction (DMD)  & $w$ & \{20, 50, 100\}\\
     \bottomrule
        \end{tabular}
        \label{tab:hyperparameters}
    \end{table}

    \subsection{Virtual reality interface}
            The virtual reality application is created similarly to the implementation in \cite{Stadtmann2023doa} and \cite{Stadtmann2023sda}. The application is created in the Unity Engine using CAD models of the setup created in the Blender design software and the Oculus Interaction SDK to integrate control through virtual reality.
            Since the data processing and analysis are coded in Python, they are not directly integrated into the virtual reality application, which is coded in C\#. Instead, the Python backend is deployed on a server (or in the cloud), and the virtual reality application accesses the data stream on demand through HTTP requests. Similar to the description in \cite{Stadtmann2023dif}, the Python side uses the FastAPI library, while the client side uses UnityWebRequests for two-way communication.
\section{Results and discussions}
\label{sec:results}
\begin{figure}[b!]
	\centering
	\includegraphics[width=\linewidth]{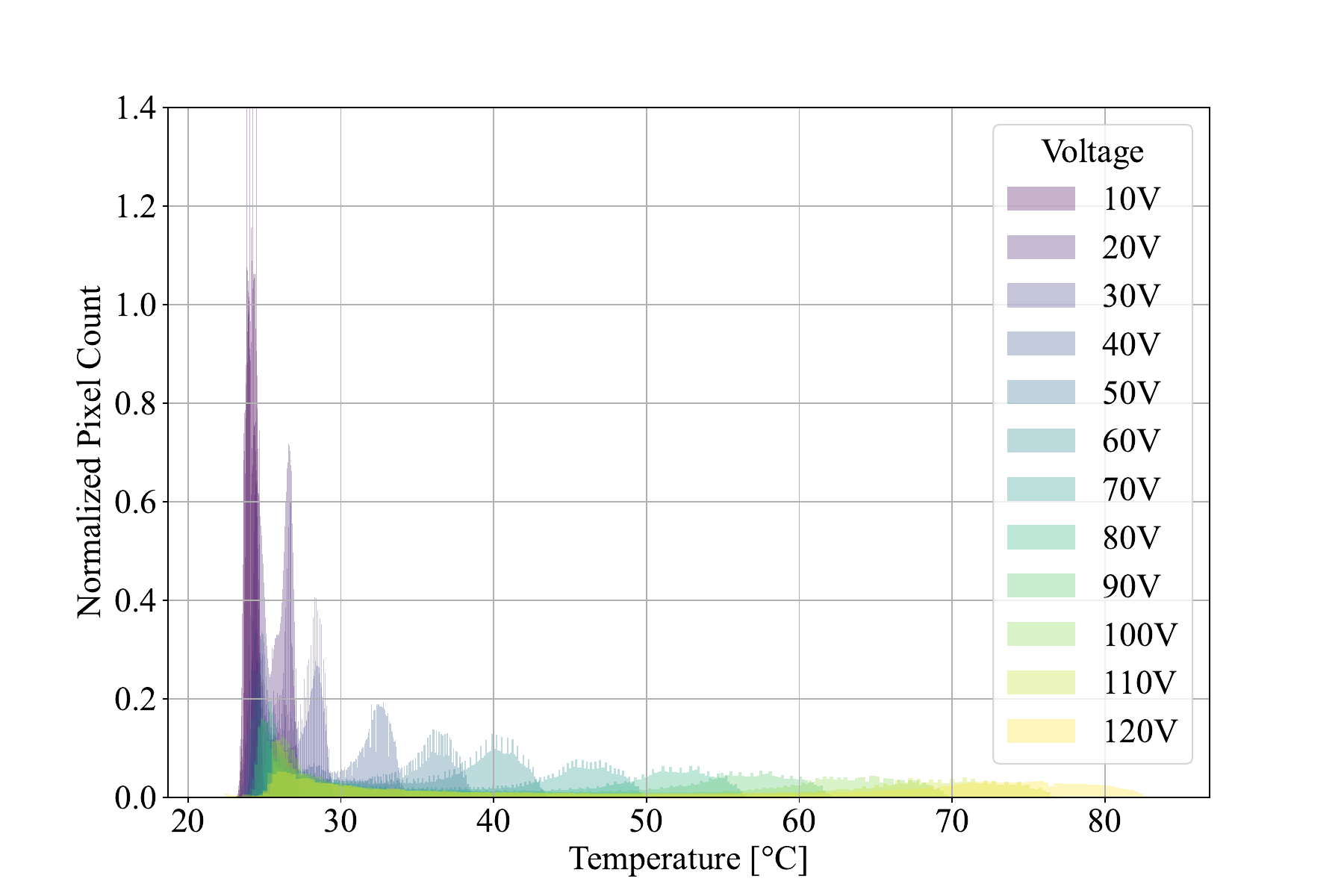}
	\caption{Temperature distribution of the training data ranging from \unit[10]{V} to \unit[120]{V}.}
	\label{fig:distribution_training_data}
\end{figure}
In this section, we present the results of image reconstruction, data imputation, anomaly detection, and state prediction. As mentioned earlier, the training set included voltages ranging from \unit[10]{V} to \unit[120]{V} in \unit[10]{V} increments. Each voltage step consisted of both a heating and a cooling phase, resulting in 24 distinct configurations. The temperature distributions for the 12 different voltage levels are shown in Fig.~\ref{fig:distribution_training_data}. It was observed that experiments conducted at higher power levels improved the signal-to-noise ratio. Consequently, in cases of limited training data, using higher-power data resulted in better denoising performance, as observed in multiple tests. The OSL reconstruction is demonstrated using only one training set (\unit[120]{V}) to showcase its effectiveness even with minimal data.

    \subsection{Automatic data cleaning using Robust PCA} \label{sec:results_RPCA}
\begin{figure}[t!]
    	\centering
    	\includegraphics[width=\linewidth]{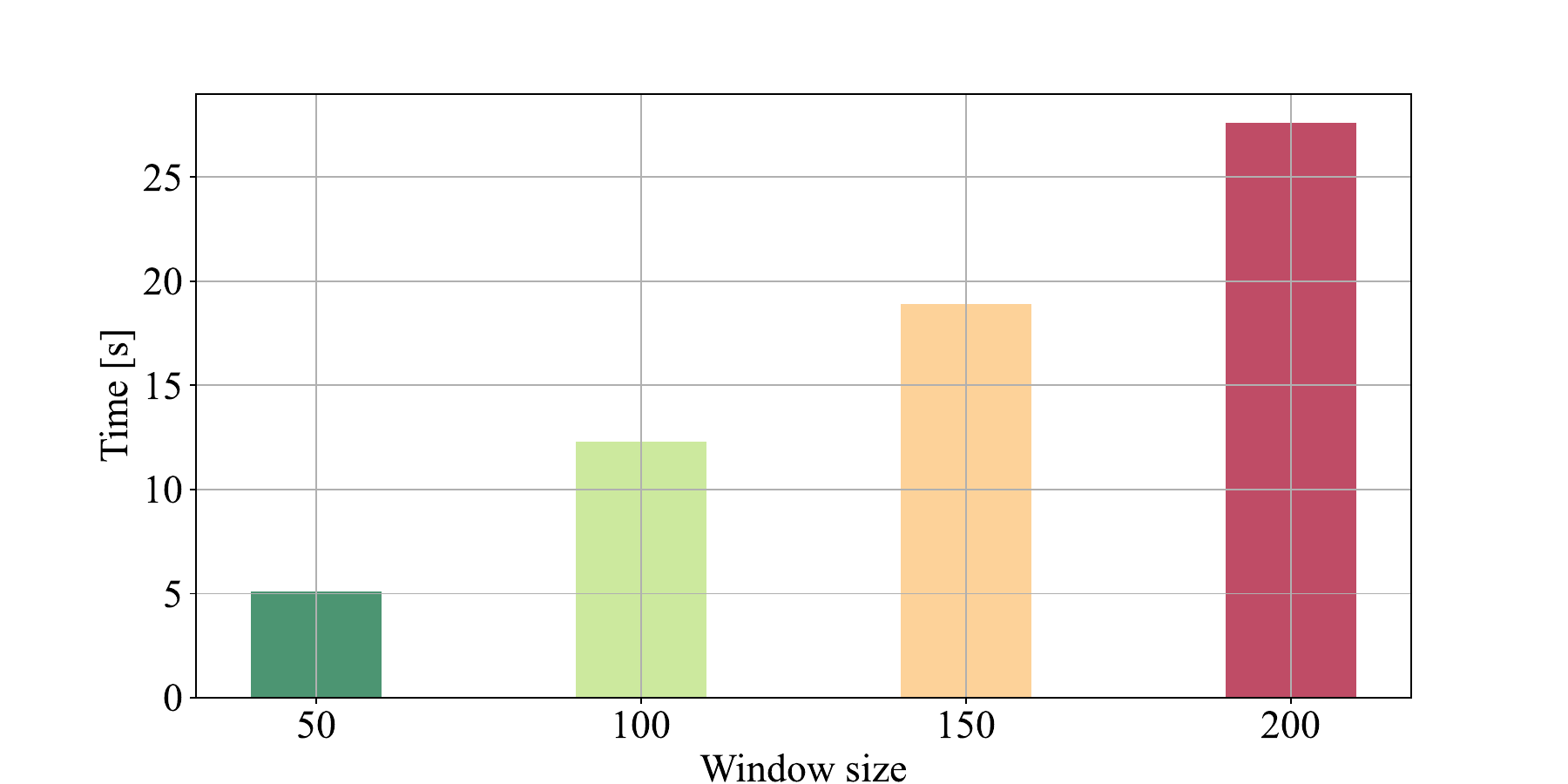}
    	\caption{Computation time of RPCA depending on the data dimension and the chosen window size. The data dimension is $n\times w$, where $n$ represents the number of pixels (78,000), and $w$ denotes the window size.}
    	\label{fig:RPCA_windows_time}
    \end{figure}

\begin{figure}[b!]
    \centering
    \begin{subfigure}[t]{0.3\linewidth}
    \includegraphics[width=\linewidth]{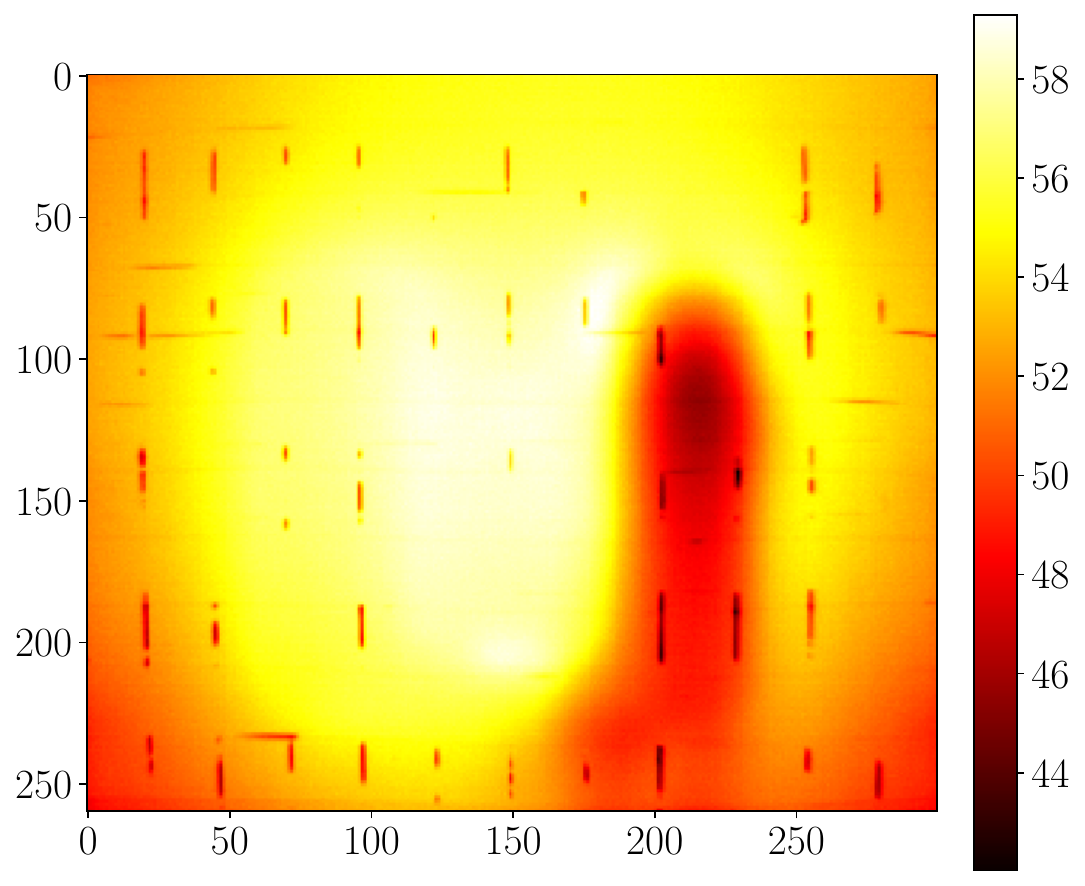} 
    \caption{Original image}
    \end{subfigure}
    \hspace{0.02\linewidth}
    \begin{subfigure}[t]{0.3\linewidth}
    \includegraphics[width=\linewidth]{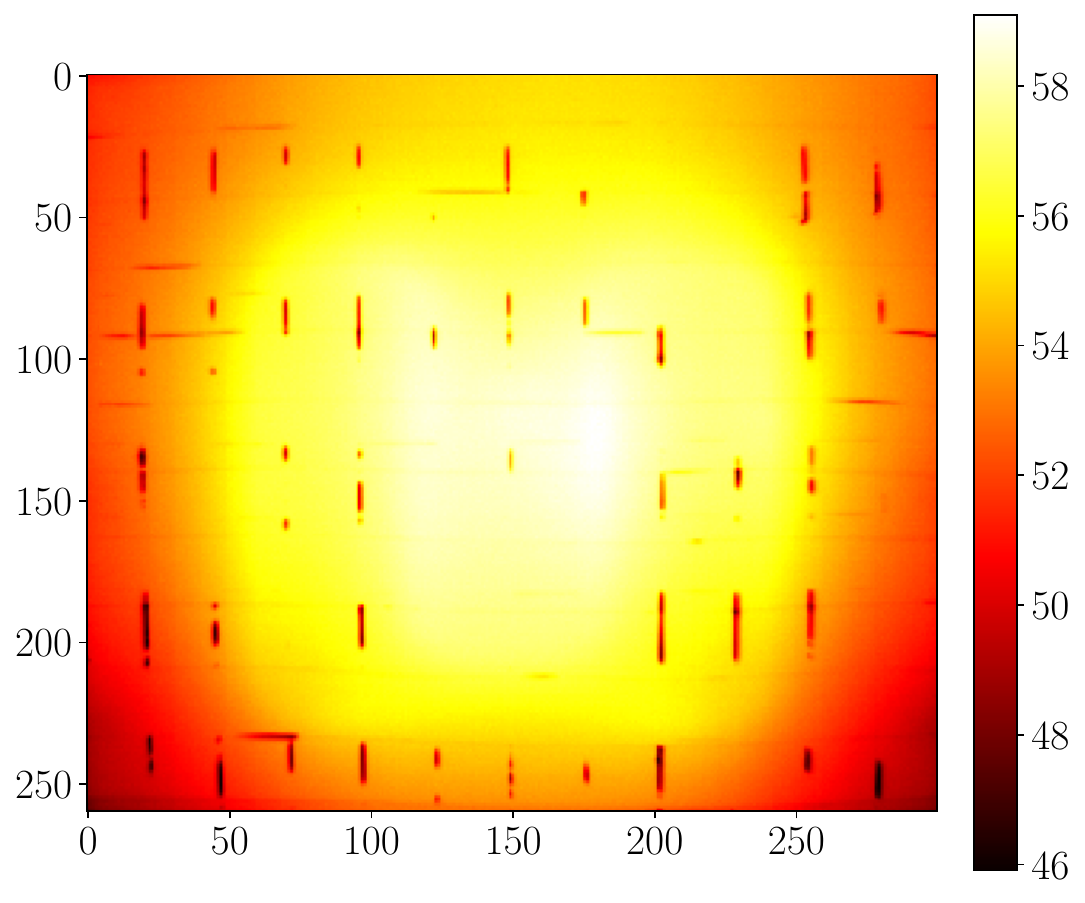}
    \caption{Matrix L from RPCA}
    \label{subfig:L_w50}
    \end{subfigure}
    \hspace{0.02\linewidth}
    \begin{subfigure}[t]{0.31\linewidth}
    \includegraphics[width=\linewidth]{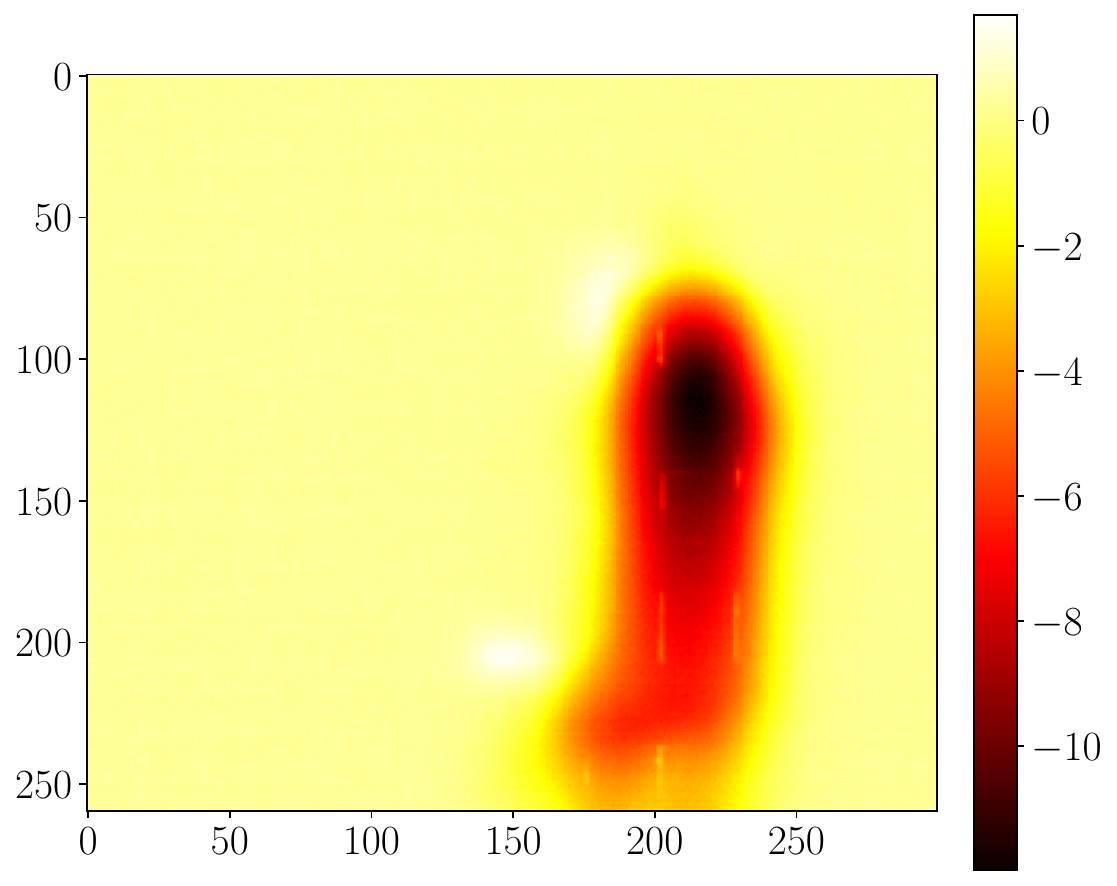} 
    \caption{Matrix S from RPCA}
        \label{subfig:S_w50}
    \end{subfigure}
    \caption{Illustration of RPCA applied to data corrupted by a water splash at the rear of the heating plate. These results were obtained using a window size of 50, with a computation time of approximately \unit[5]{s}.}
    \label{fig:RPCA}
    \end{figure}

    Since an optimization problem must be solved, RPCA can become too time-consuming for real-time cleaning, depending on the hardware, the data dimension (78,000 pixels in our example), and the selected window size. In Fig.~\ref{fig:RPCA_windows_time}, the computation time for different window sizes is compared. The longer the window size, the better the results at the cost of computational time.
    However, a window size of only 50 yields a robust representation of the underlying modes extracted in the matrix $\mathbf{L}$. Simultaneously, all corruptions are captured by the matrix $\mathbf{S}$, as demonstrated in Fig.~\ref{fig:RPCA}. This approach can be utilized for either anomaly detection or data cleaning.

    \begin{figure*}[t!]
    \centering
    \begin{subfigure}{0.24\linewidth}
    \includegraphics[width=\linewidth]{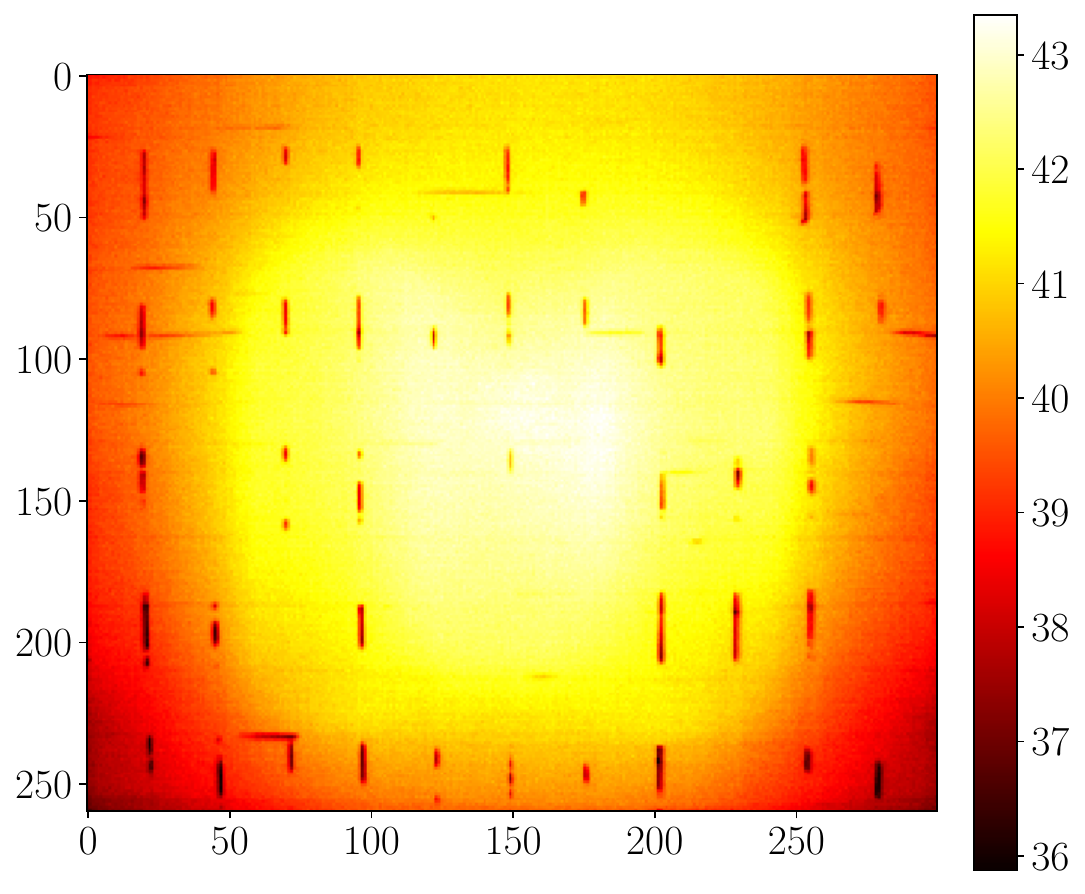} 
    \caption{Original image}
    \end{subfigure}
    \begin{subfigure}{0.211\linewidth}
    \includegraphics[width=\linewidth]{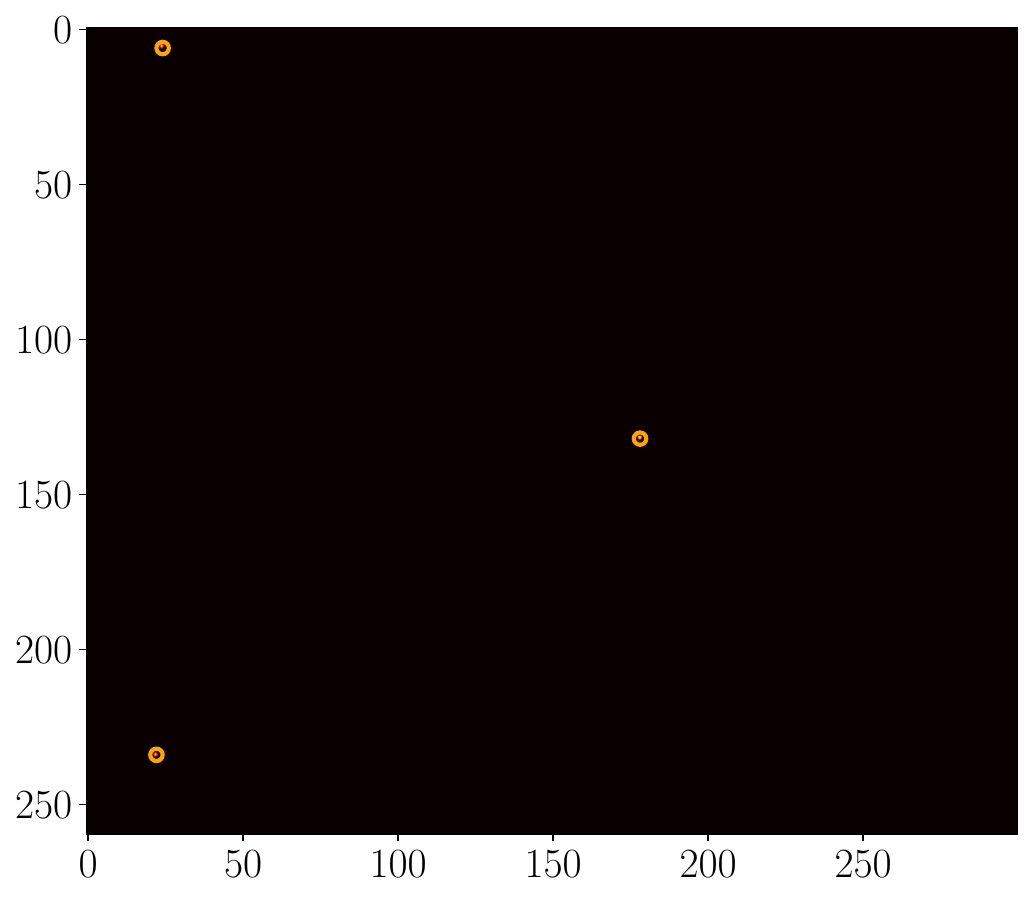}
    \caption{3 measured pixels}
    \label{subfig:osp}
    \end{subfigure}
    \begin{subfigure}{0.24\linewidth}
    \includegraphics[width=\linewidth]{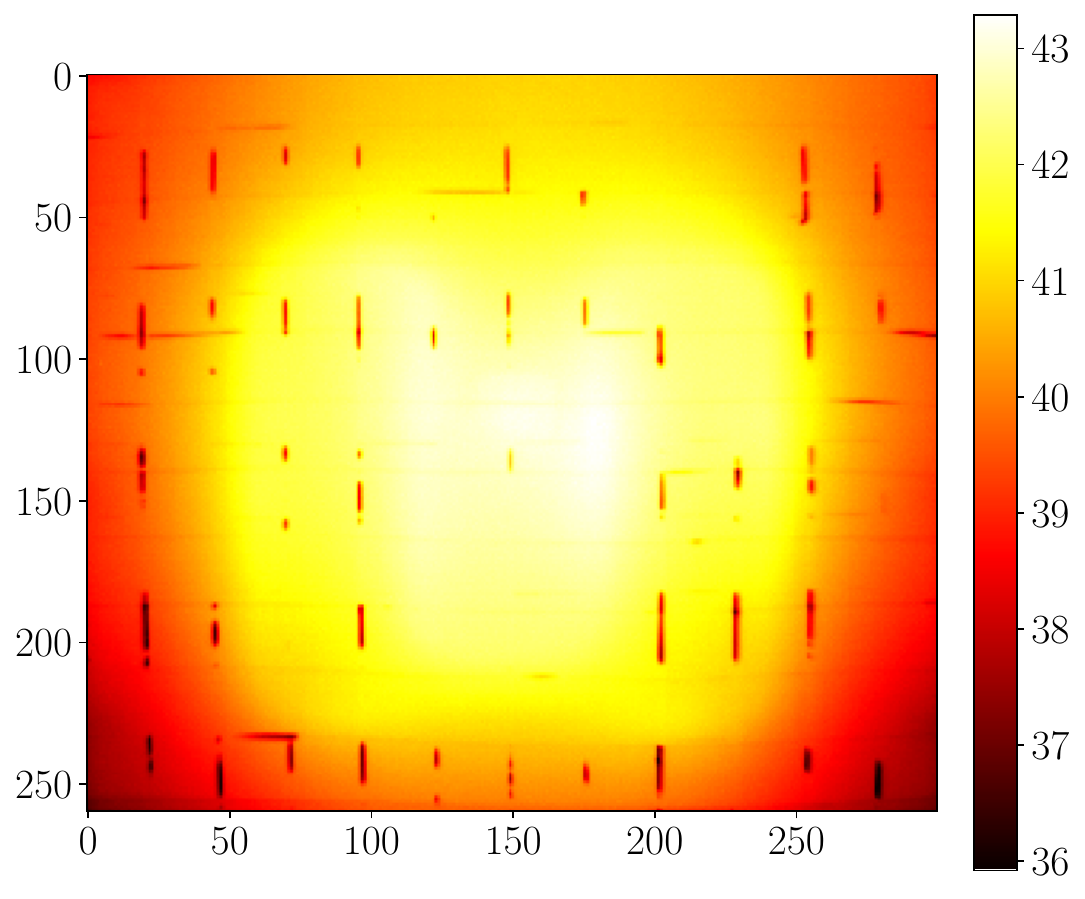} 
    \caption{Reconstructed image}
        \label{subfig:osp_reconstructed_image}
    \end{subfigure}
    \begin{subfigure}{0.24\linewidth}
    \includegraphics[width=\linewidth]{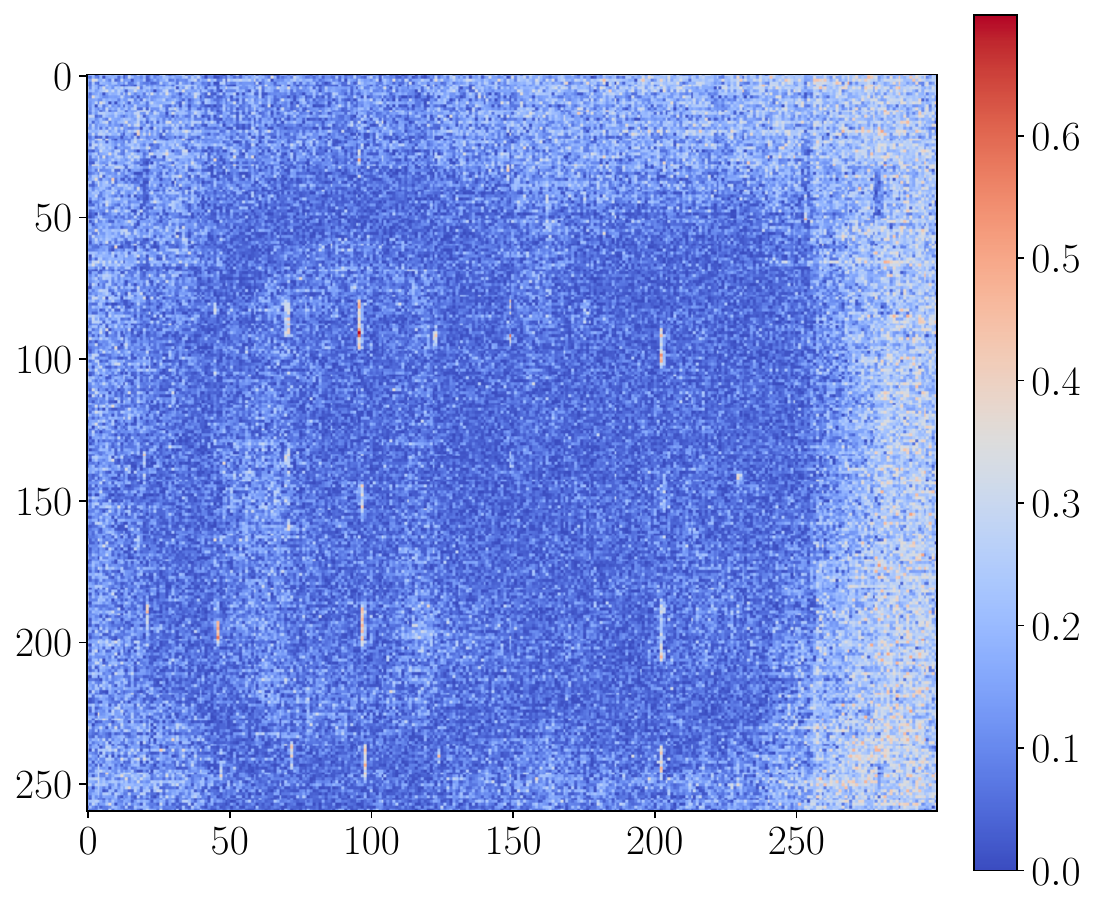} 
    \caption{Reconstruction error}
    \label{subfig:Error_image_OSP}
    \end{subfigure}
    \caption{Dimensionality reduction and image reconstruction using OSL with only three pixel measurements. A randomly selected time sample from the test set was used, with training conducted exclusively at \unit[120]{V} and testing at \unit[60]{V}, to demonstrate the model's capability even with limited training data.}
    \label{fig:Optimal_sensor_placement}
    \end{figure*}

    \subsection{3D reconstruction from sparse data}
    To compress the data, OSL is applied using a reduced order of $r=3$ since the first three POD modes already capture 99.9968\% of the total variance. Fig.~\ref{fig:Optimal_sensor_placement} illustrates the effectiveness of the OSL algorithm in reconstructing the original images by utilizing only a small subset of three pixel measurements. The optimal sampling locations are shown in Fig.~\ref{subfig:osp}, while the reconstruction error is depicted in Fig.~\ref{subfig:Error_image_OSP}, highlighting the precision of the method. 
    A randomly selected time sample from the test set was used, with training conducted solely at \unit[120]{V}, while testing was carried out at \unit[60]{V}. Although the error could be further reduced with more training data, this example demonstrates the method’s effectiveness even with limited training data. The largest errors occur in areas where adhesive strips were applied to create non-uniform thermal behavior, while the lowest errors are found in regions near the heating elements, indicating that the dynamical modes in those areas are well understood.
    Additionally, OSL offers a denoising effect, which can be observed by comparing the reconstructed image with the original.

    \subsection{Data Imputation}
\begin{figure}[b!]
    	\centering
    	\includegraphics[width=\linewidth]{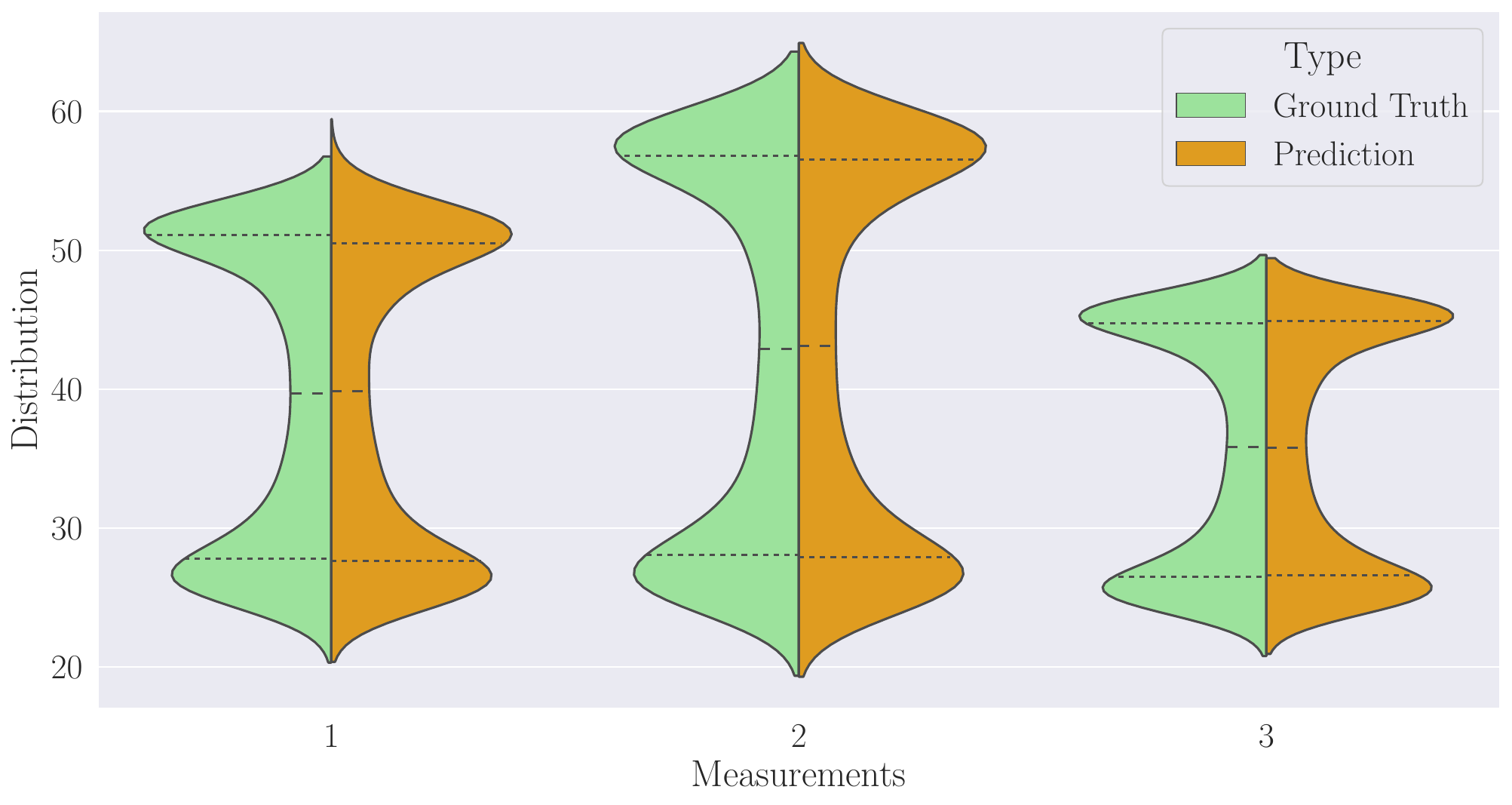}
    	\caption{Distribution of the three SVR measurement model predictions compared to the ground truth. The testing was conducted using a dataset including anomalies and a voltage setup of \unit[85]{V}, a parameter configuration not included in the training phase.}
    	\label{fig:Imputation}
    \end{figure}

    The imputation models built by SVR are used for improving the robustness in case one of the measured pixels is erroneous. Erroneous measurements can be tracked by, e.g., using the distribution from training and identifying whether a measured pixel lies within that expected range. Otherwise, the pixel can be imputed using its SVR model. The prediction performance of the data-driven models is presented in Fig.~\ref{fig:Imputation} showing well-performing predictions demonstrated by distributions via violin plots.
     
    \subsection{Anomaly detection}
\begin{figure}[b!]
        \centering
        \begin{subfigure}[b]{\linewidth}
            \centering
        \includegraphics[width=0.98\linewidth]{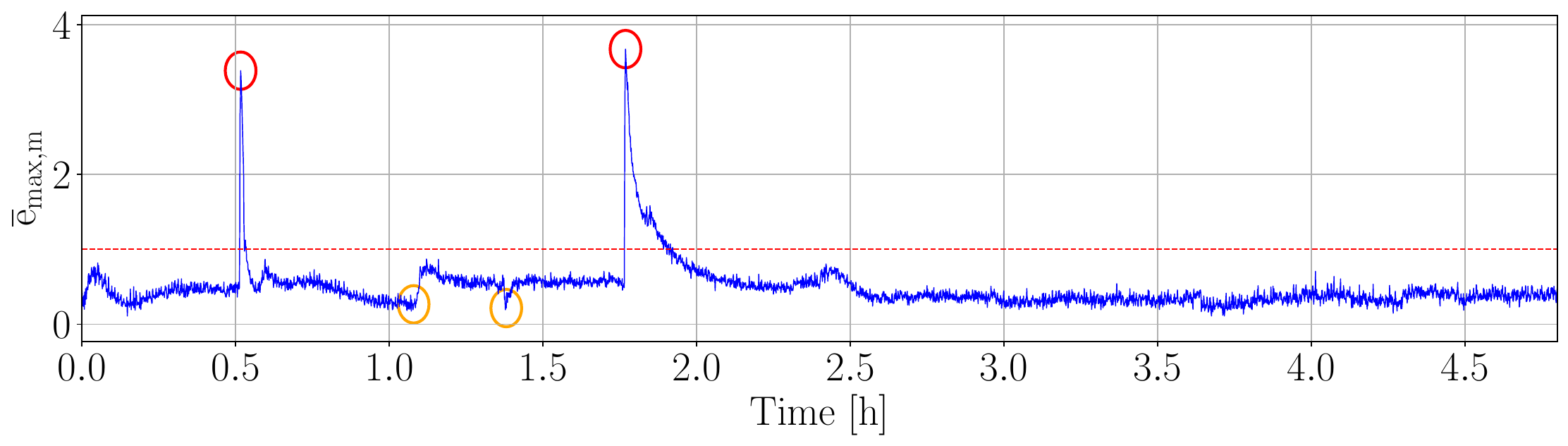}
            \caption{Maximum error values. The red dashed line represents $\gamma_1$.}
            \label{fig:operation_error}
        \end{subfigure}
        \begin{subfigure}[b]{\linewidth}
            \centering
            \includegraphics[width=\linewidth]{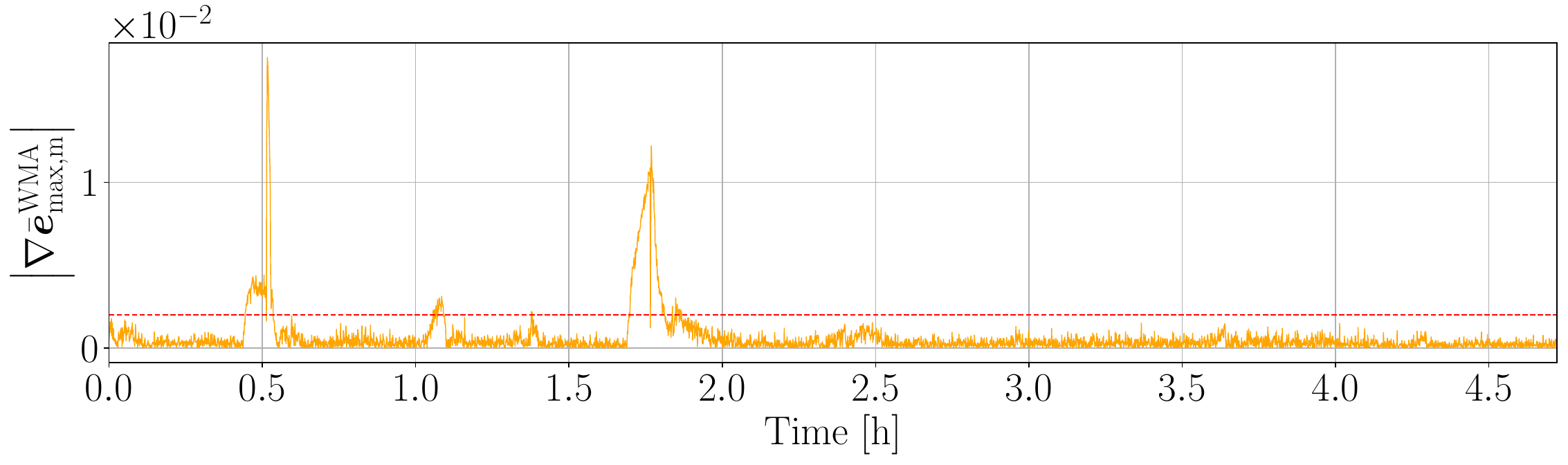}
            \caption{WMA gradient of the maximum error values. The red dashed line represents $\gamma_2$.}
         \label{fig:gradient_operation_error}
        \end{subfigure}
        \caption{Anomaly detection triggered by large values of $\bar{\boldsymbol{e}}_{\mathrm{max,m}}$ (red circles), or large values of $\left|\nabla \bar{\boldsymbol{e}}_{\mathrm{max,m}}^{\mathrm{WMA}}\right|$ (orange circles).}
        \label{fig:combined_operations}
    \end{figure}

    As mentioned earlier, two distinct anomaly types were introduced into the experiments for the test data set. The heating plate was perturbed twice by splashing water on its rear, and once, a metallic object was placed at the rear, in proximity to the heating plate, for approximately \unit[20]{min}. Subsequently, the metallic object was removed.
    The anomaly detection process over time is illustrated in Fig.~\ref{fig:combined_operations}, where the red circles indicate how the anomaly detection is activated by perturbations caused by water splashes. The orange circles represent the anomaly detection in action during the introduction or removal of the metallic object.
    
\begin{figure}[t!]
    \centering
        \begin{subfigure}[t]{0.3\linewidth}
        \includegraphics[width=\linewidth]{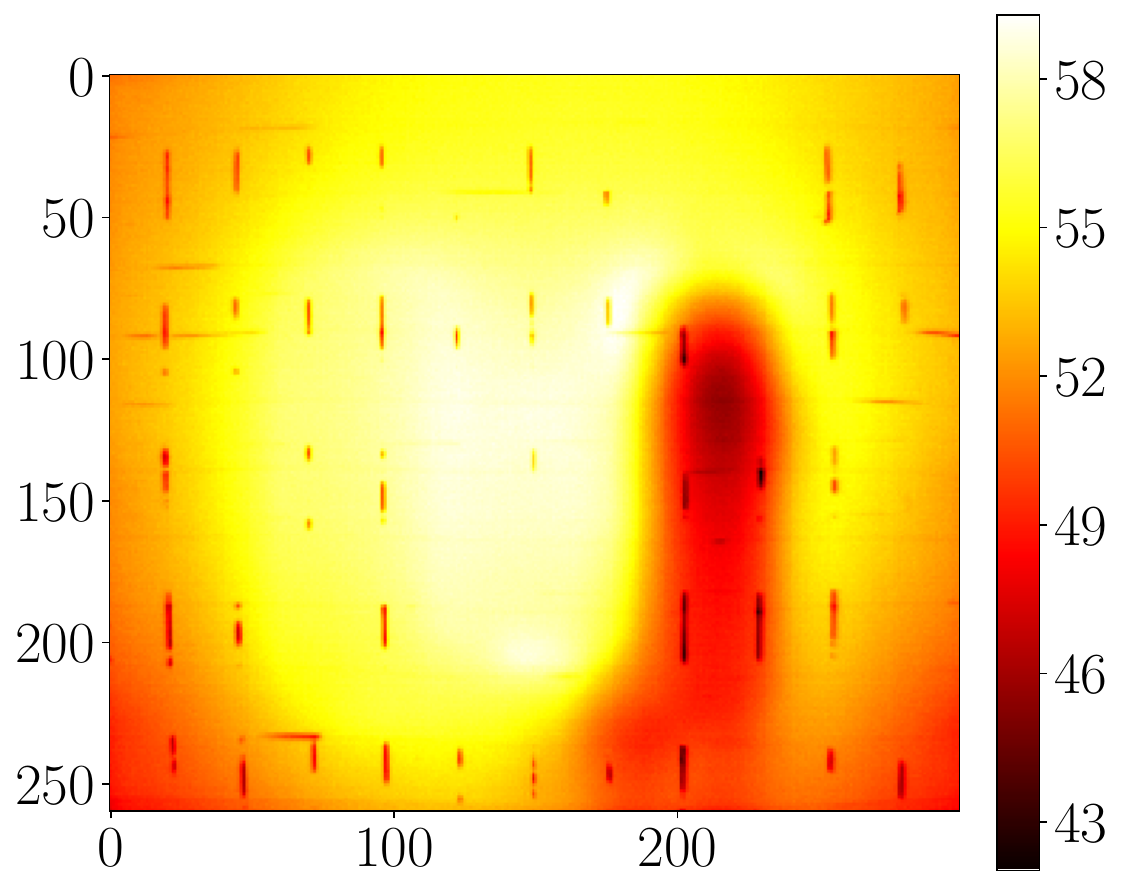} 
        \caption{Original image}
        \end{subfigure}
        \hspace{0.02\linewidth}
        \begin{subfigure}[t]{0.3\linewidth}
        \includegraphics[width=\linewidth]{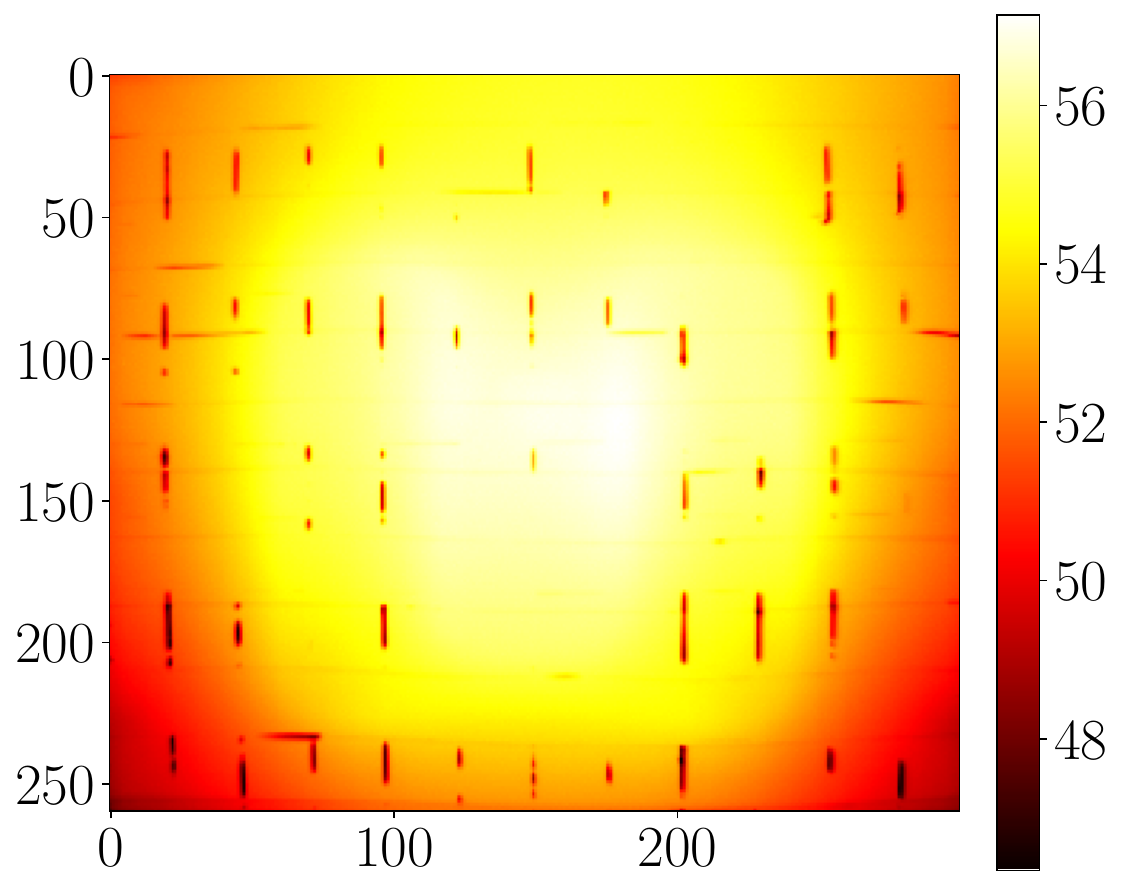}
        \caption{OSL reconstruction}
        \label{subfig:Anomaly_reconstructed_image}
        \end{subfigure}
        \hspace{0.02\linewidth}
        \begin{subfigure}[t]{0.31\linewidth}
        \includegraphics[width=\linewidth]{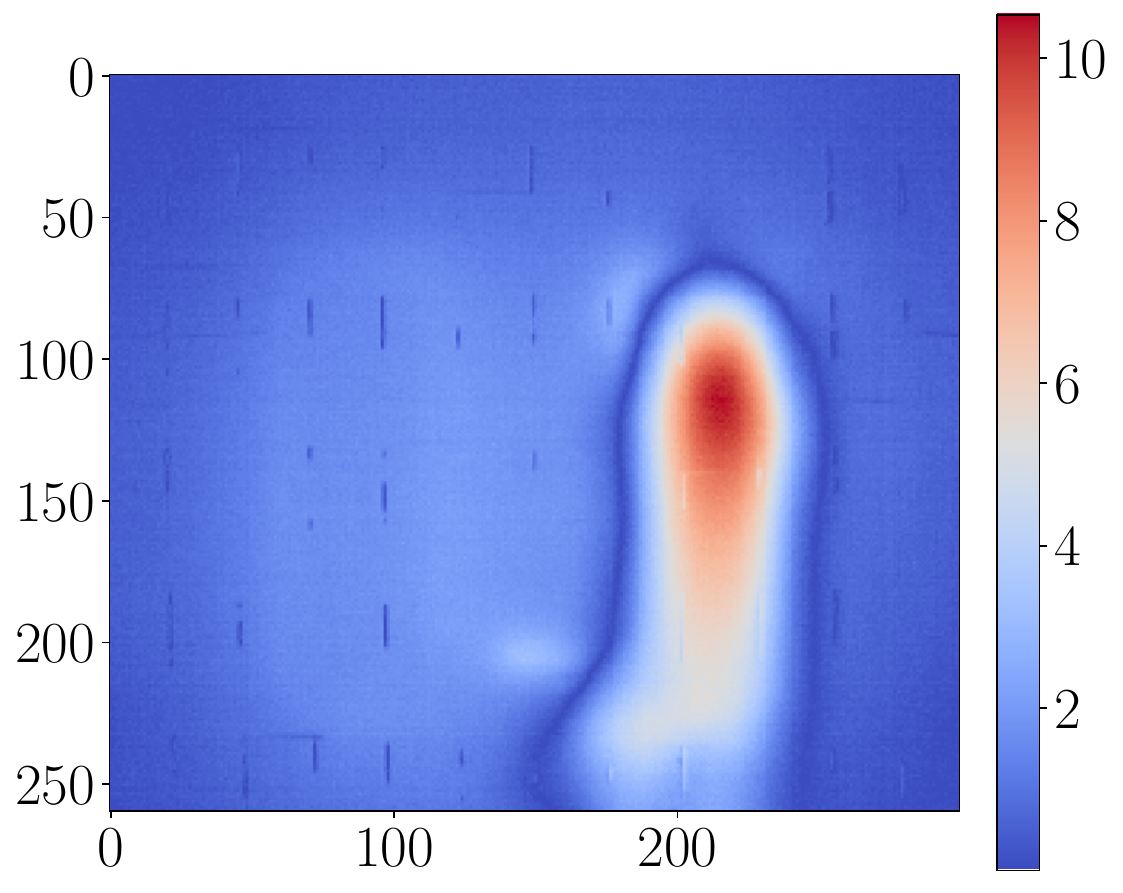} 
        \caption{Error image}
            \label{subfig:error_water}
        \end{subfigure}
    \caption{Anomaly detection while identifying a water splash at the rear of the heating plate.}
    \label{fig:anomaly_detection_water}
    \end{figure}

    The corruption discussed in Section~\ref{sec:results_RPCA}, which applies RPCA, is associated with the second red circle in the figure. In comparison, the same anomaly detected using the approach described in Section~\ref{sec:anomaly_detection} is showcased in Fig.~\ref{fig:anomaly_detection_water}.
    
\begin{figure}[htb!]
    \centering
        \begin{subfigure}[t]{0.19\linewidth}
        \includegraphics[width=\linewidth]{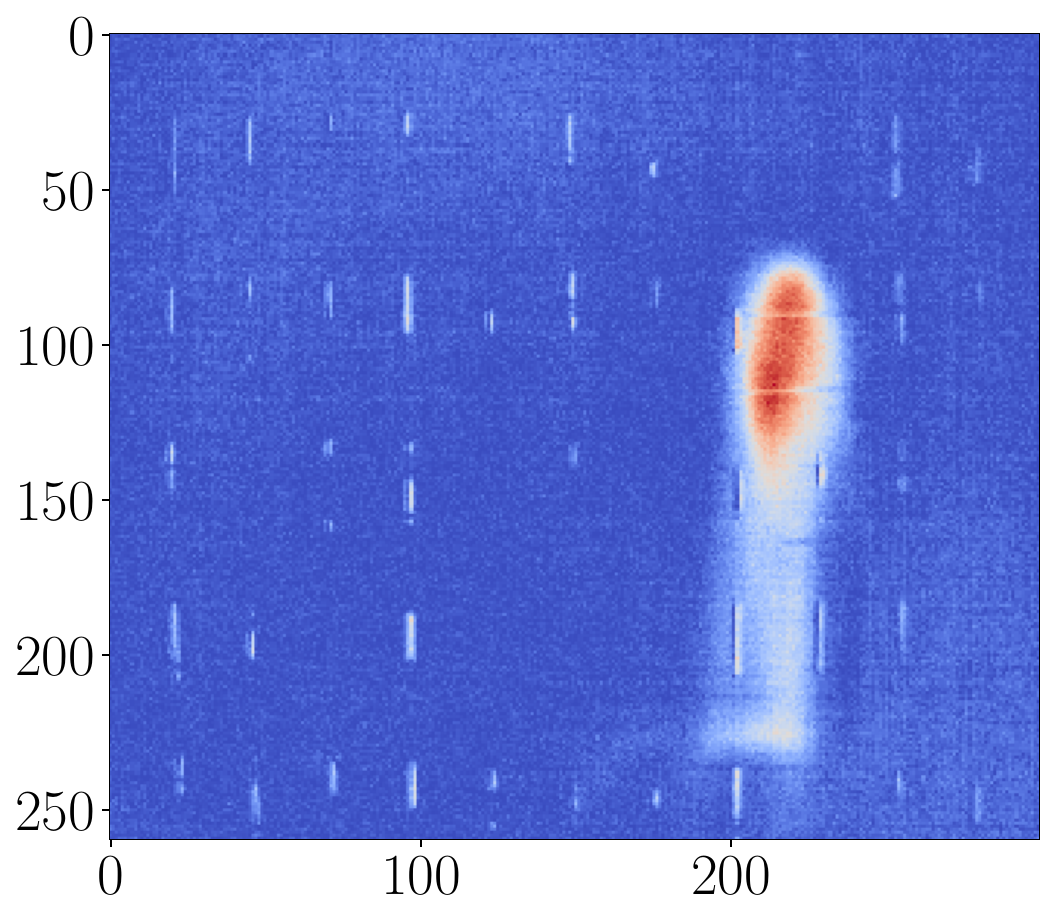} 
        \end{subfigure}
        \begin{subfigure}[t]{0.19\linewidth}
        \includegraphics[width=\linewidth]{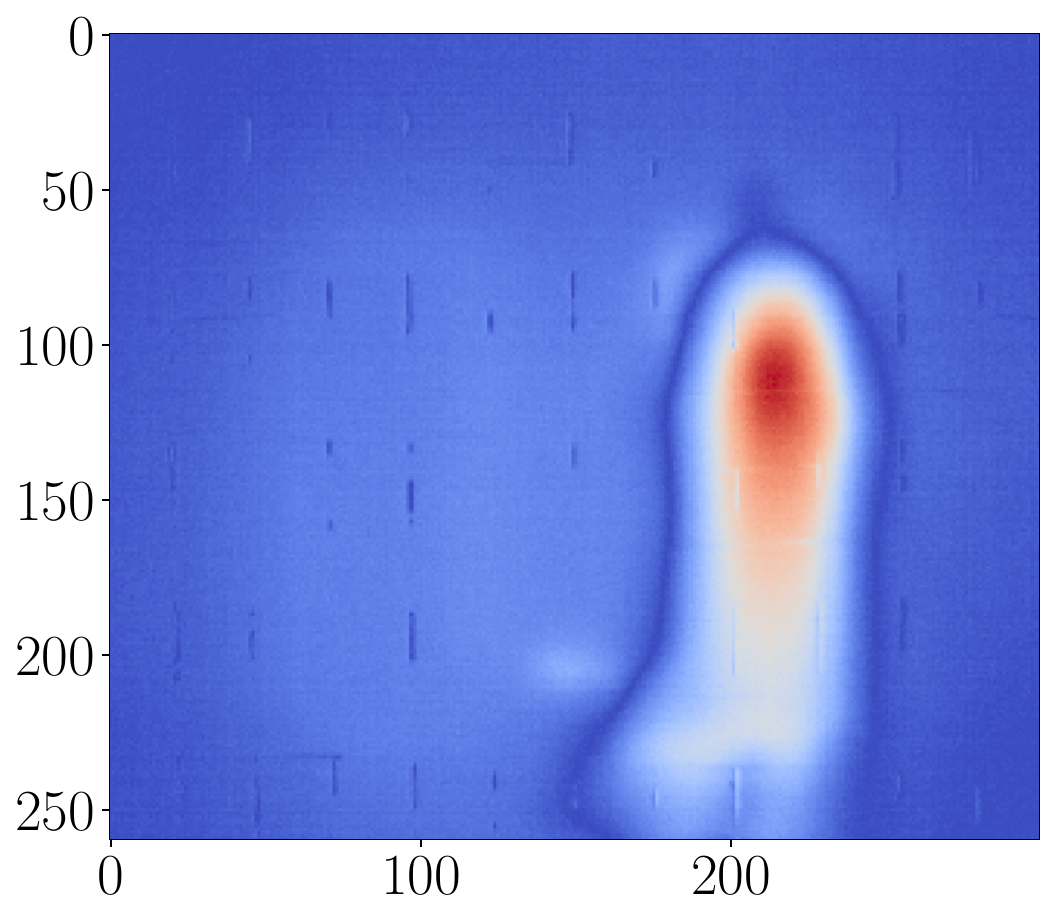}
        \end{subfigure}
        \begin{subfigure}[t]{0.19\linewidth}
        \includegraphics[width=\linewidth]{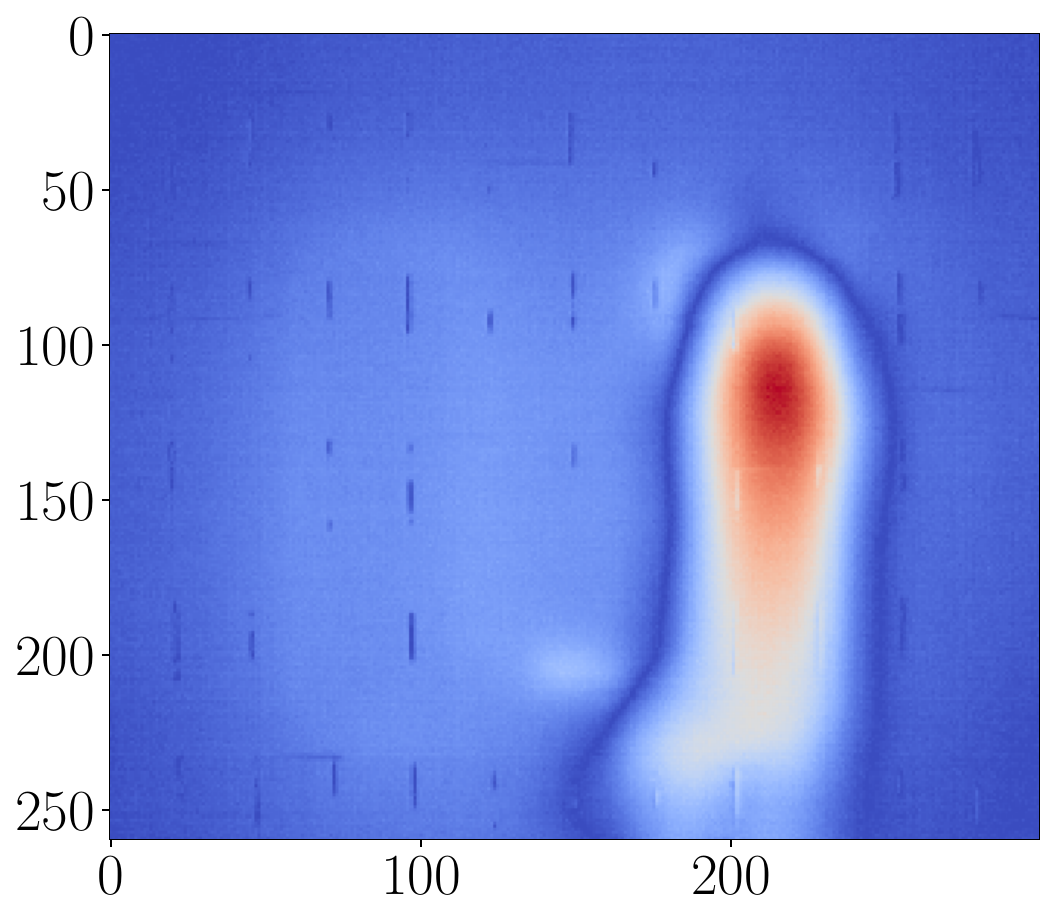} 
        \end{subfigure}
        \begin{subfigure}[t]{0.19\linewidth}
        \includegraphics[width=\linewidth]{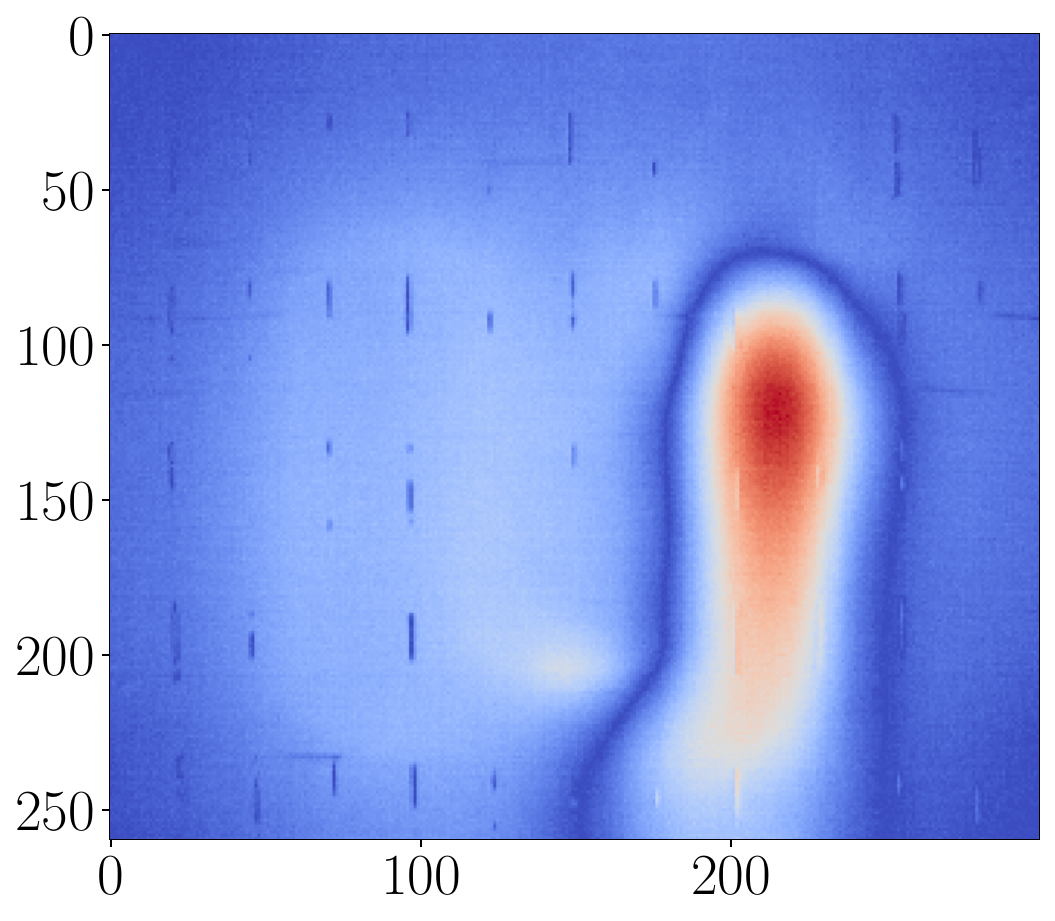}
        \end{subfigure}
        \begin{subfigure}[t]{0.19\linewidth}
        \includegraphics[width=\linewidth]{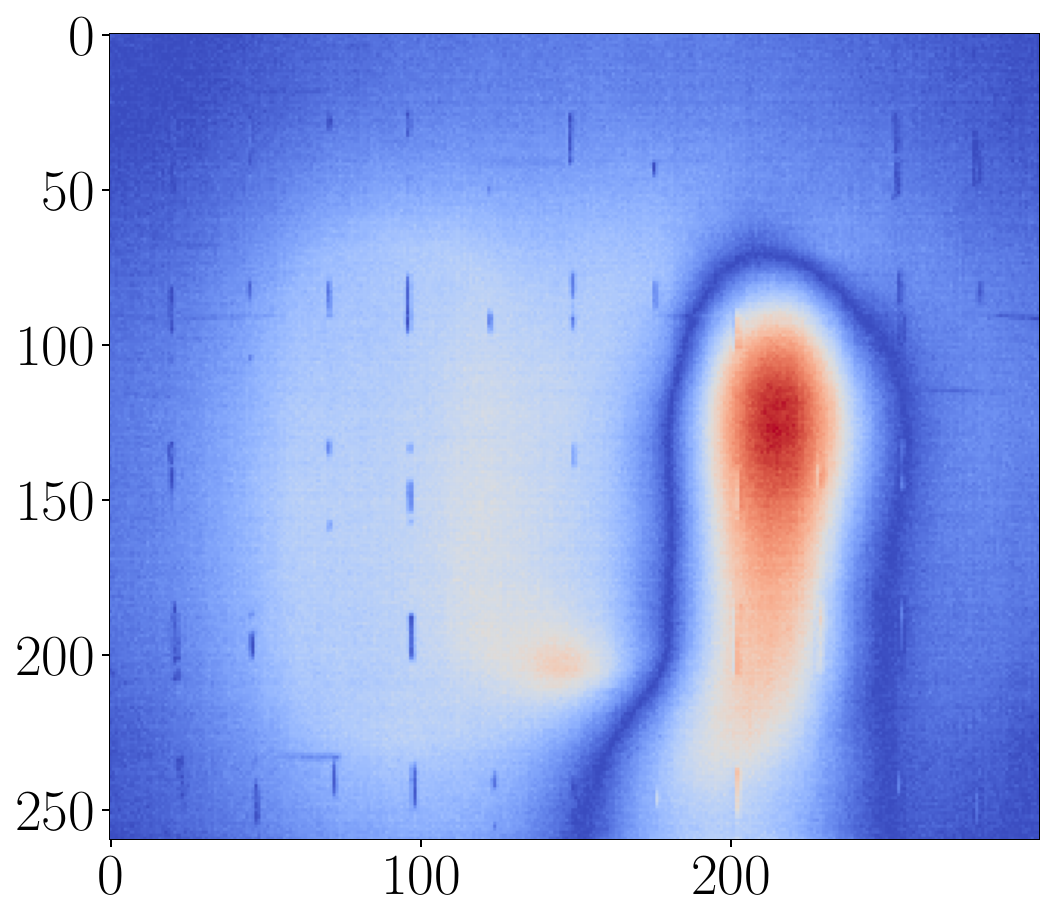} 
        \end{subfigure}
        \\
        \begin{subfigure}[t]{0.19\linewidth}
        \includegraphics[width=\linewidth]{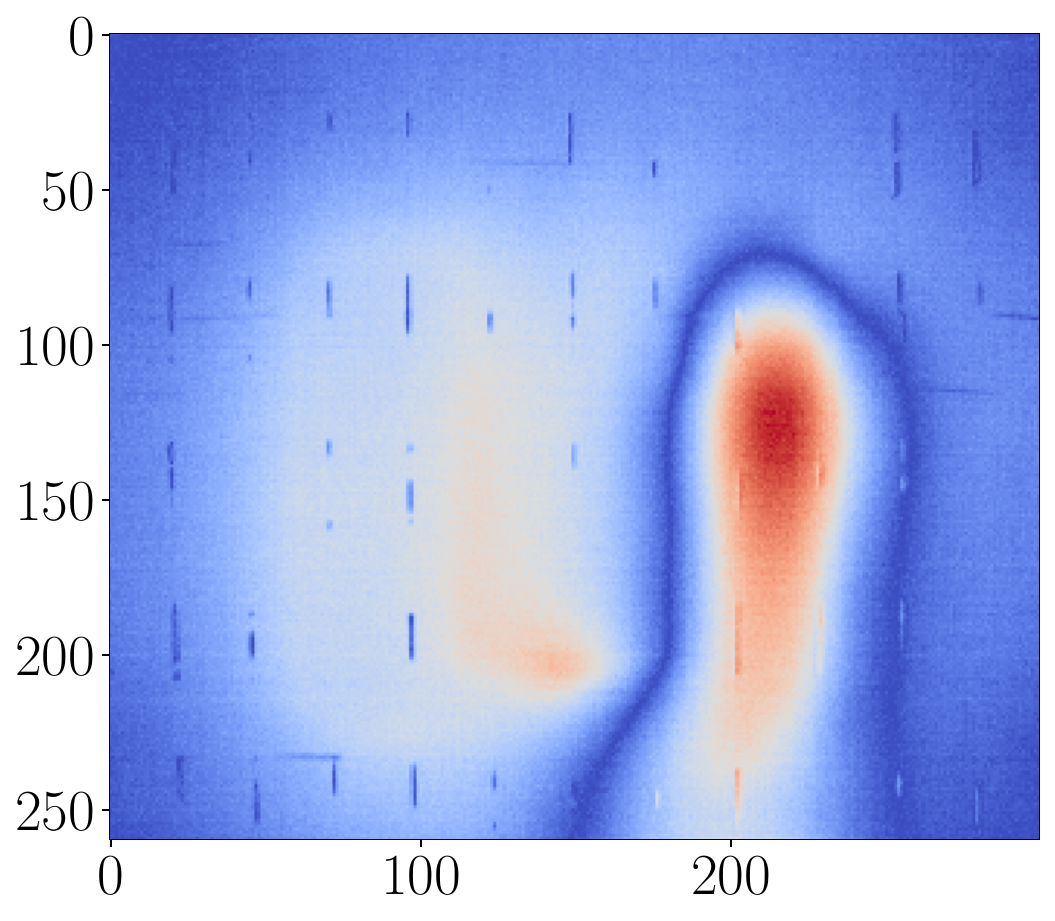} 
        \end{subfigure}
        \begin{subfigure}[t]{0.19\linewidth}
        \includegraphics[width=\linewidth]{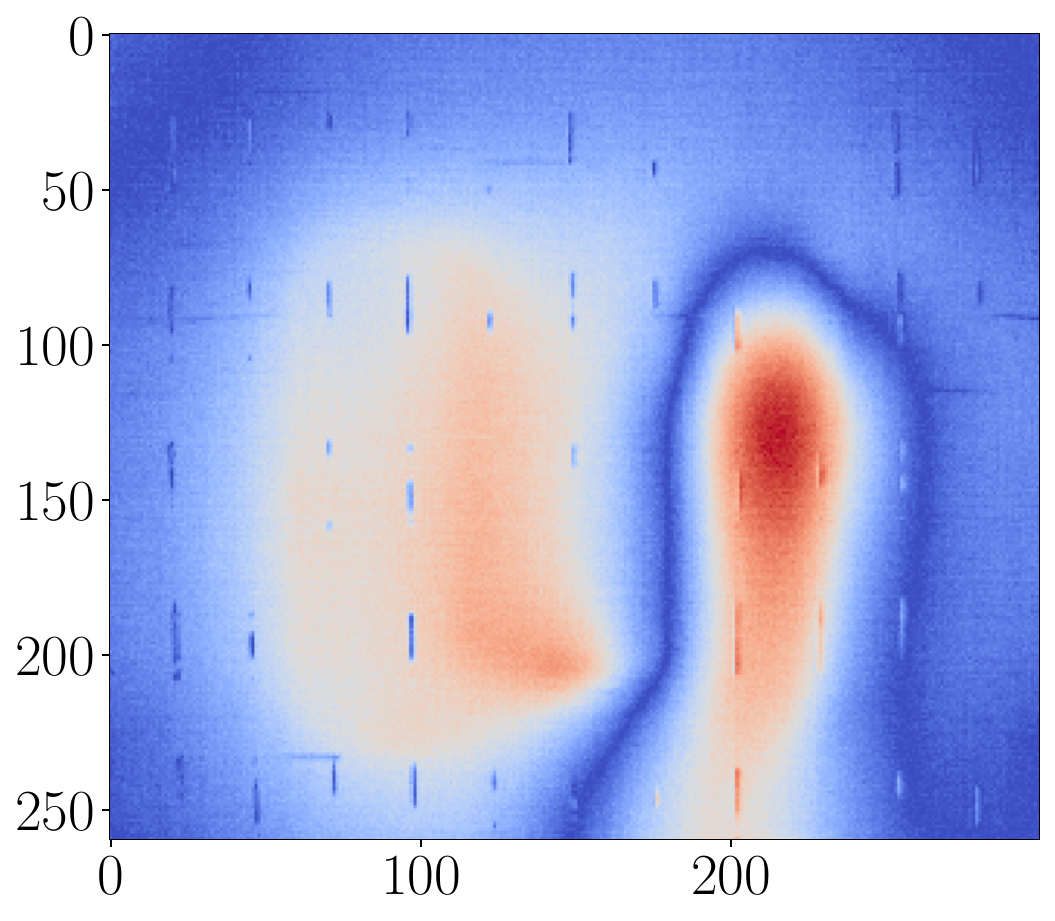}
        \end{subfigure}
        \begin{subfigure}[t]{0.19\linewidth}
        \includegraphics[width=\linewidth]{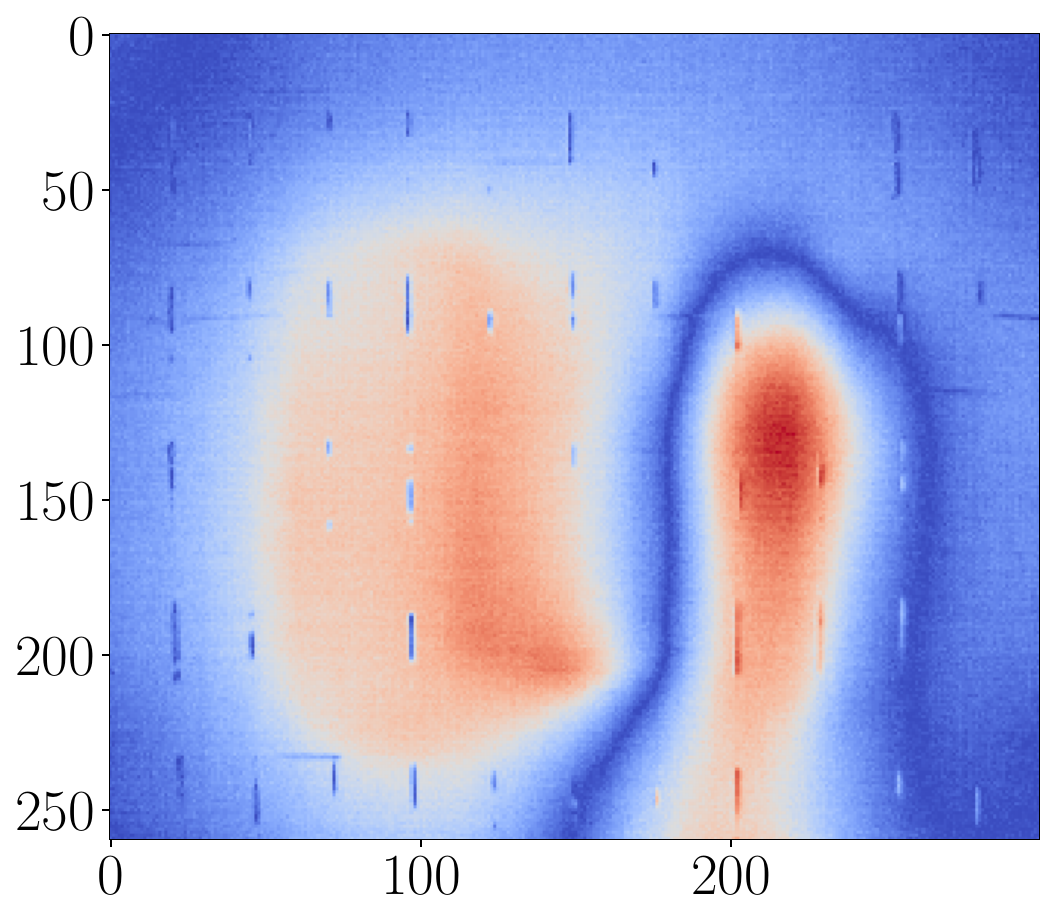} 
        \end{subfigure}
        \begin{subfigure}[t]{0.19\linewidth}
        \includegraphics[width=\linewidth]{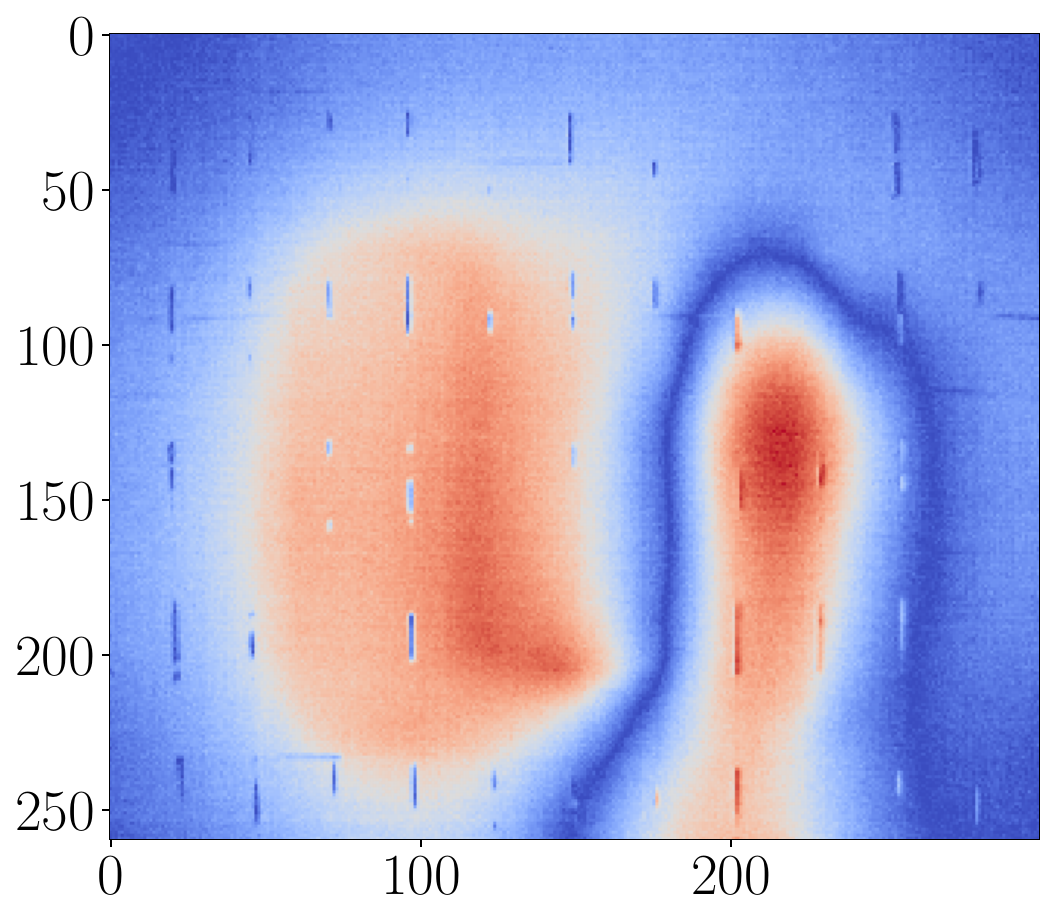}
        \end{subfigure}
        \begin{subfigure}[t]{0.19\linewidth}
        \includegraphics[width=\linewidth]{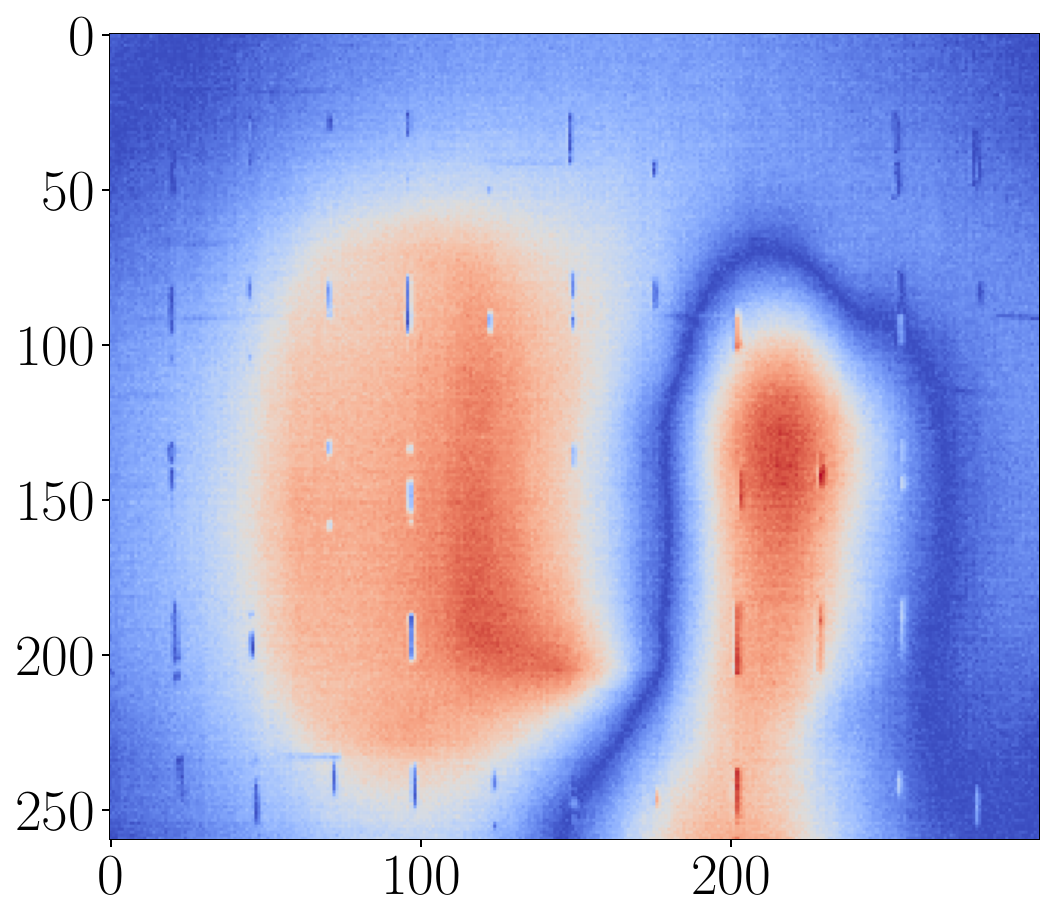} 
        \end{subfigure}
    \caption{Time series data of real-time anomaly detection.}
    \label{fig:anomaly_detection_timeseries}
    \end{figure}

\begin{figure}[htb!]
    \centering
        \begin{subfigure}[t]{0.3\linewidth}
        \includegraphics[width=\linewidth]{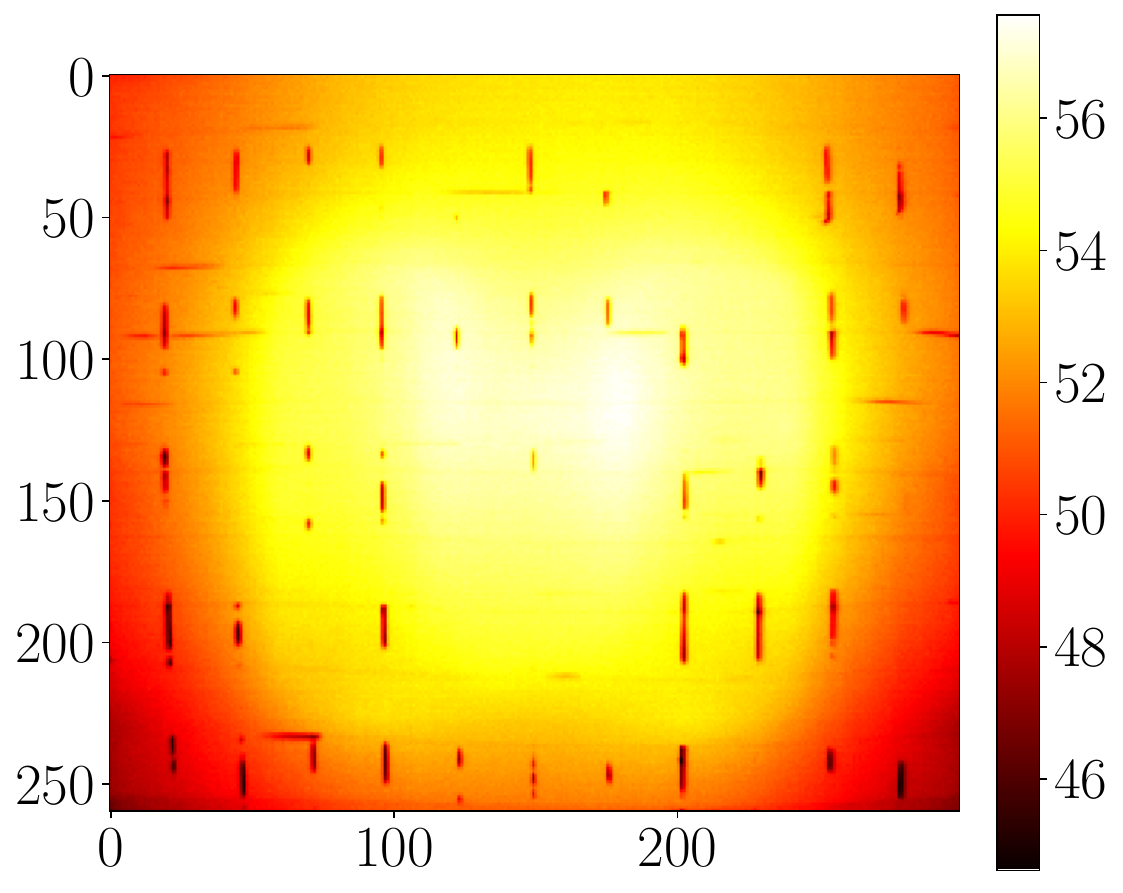} 
        \caption{Original image}
        \end{subfigure}
        \hspace{0.02\linewidth}
        \begin{subfigure}[t]{0.3\linewidth}
        \includegraphics[width=\linewidth]{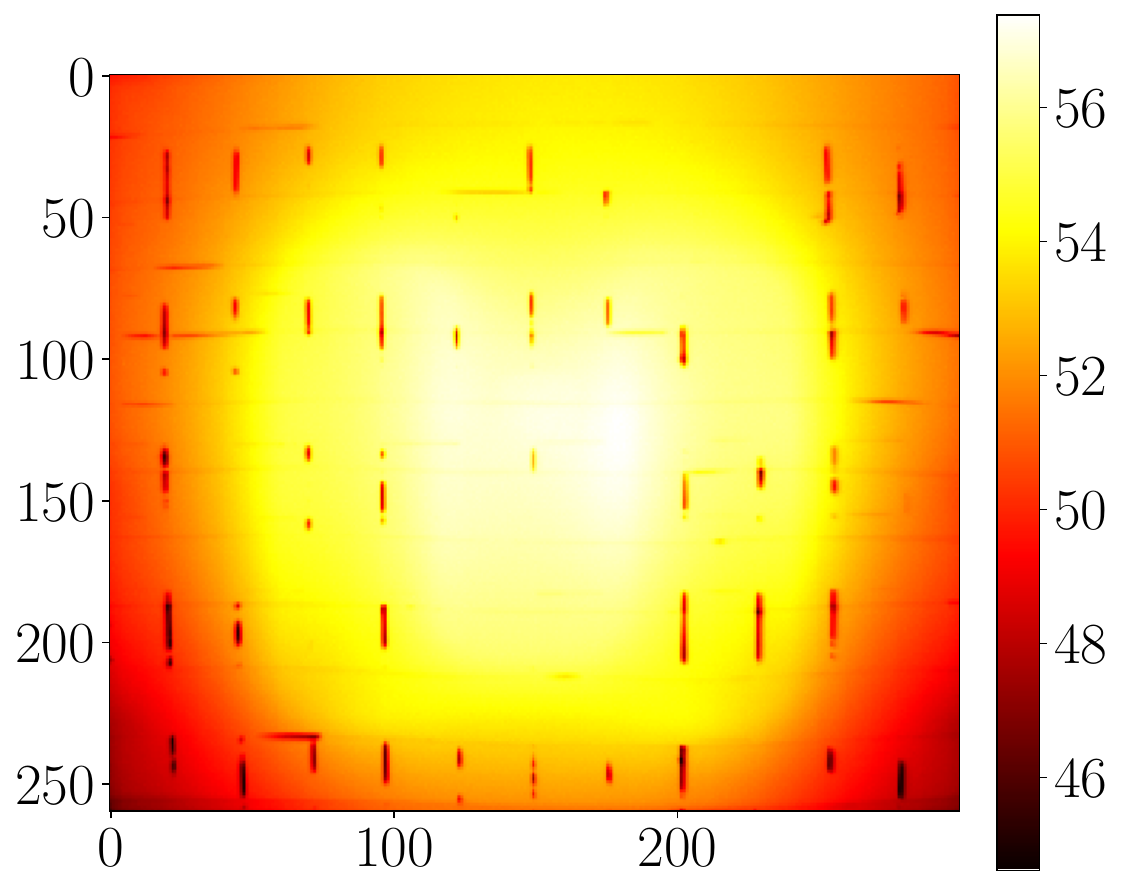}
        \caption{OSL reconstruction}
        \label{subfig:Anomaly_reconstructed_image_knife}
        \end{subfigure}
        \hspace{0.02\linewidth}
        \begin{subfigure}[t]{0.31\linewidth}
        \includegraphics[width=\linewidth]{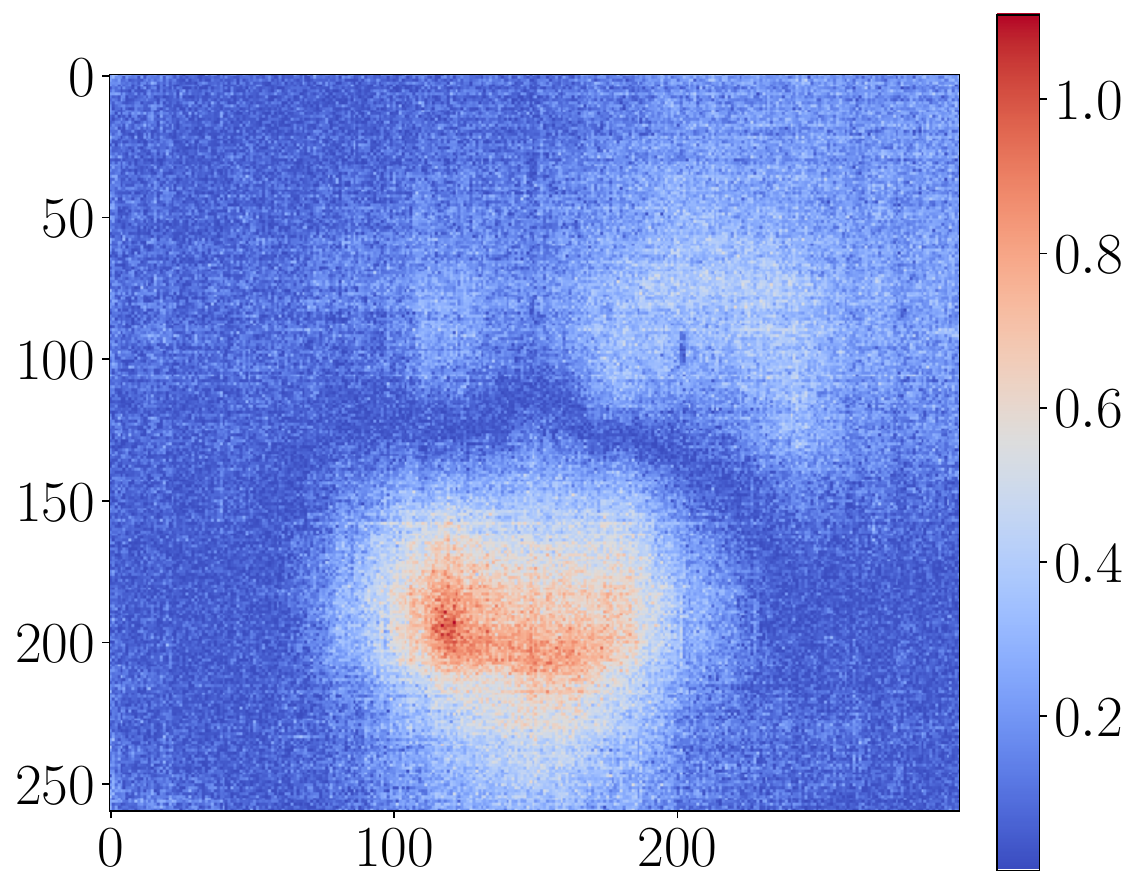} 
        \caption{Error image}
            \label{subfig:error_knife}
        \end{subfigure}
    \caption{Anomaly detection while identifying a metallic object at the rear of the heating plate. As demonstrated in the original image, a human eye can not identify any irregularities, while the anomaly detection algorithm is able to identify a variation.}
    \label{fig:anomaly_detection_knife}
    \end{figure}

    However, in comparison to RPCA, this anomaly detection approach functions in real-time without the delay constraints imposed by optimization processes. As a result, real-time monitoring can be effectively achieved, as demonstrated in Fig.~\ref{fig:anomaly_detection_timeseries}.
    Furthermore, whereas RPCA was unable to identify the anomalies introduced by the metallic object (represented by the orange circles in the figure), the approach outlined in Section~\ref{sec:anomaly_detection} effectively detects these anomalies, as illustrated in Fig.~\ref{fig:anomaly_detection_knife}.
    
    \subsection{Predicting the future state of the system}
\begin{figure}[t!]
    \centering
    \includegraphics[width=\linewidth]{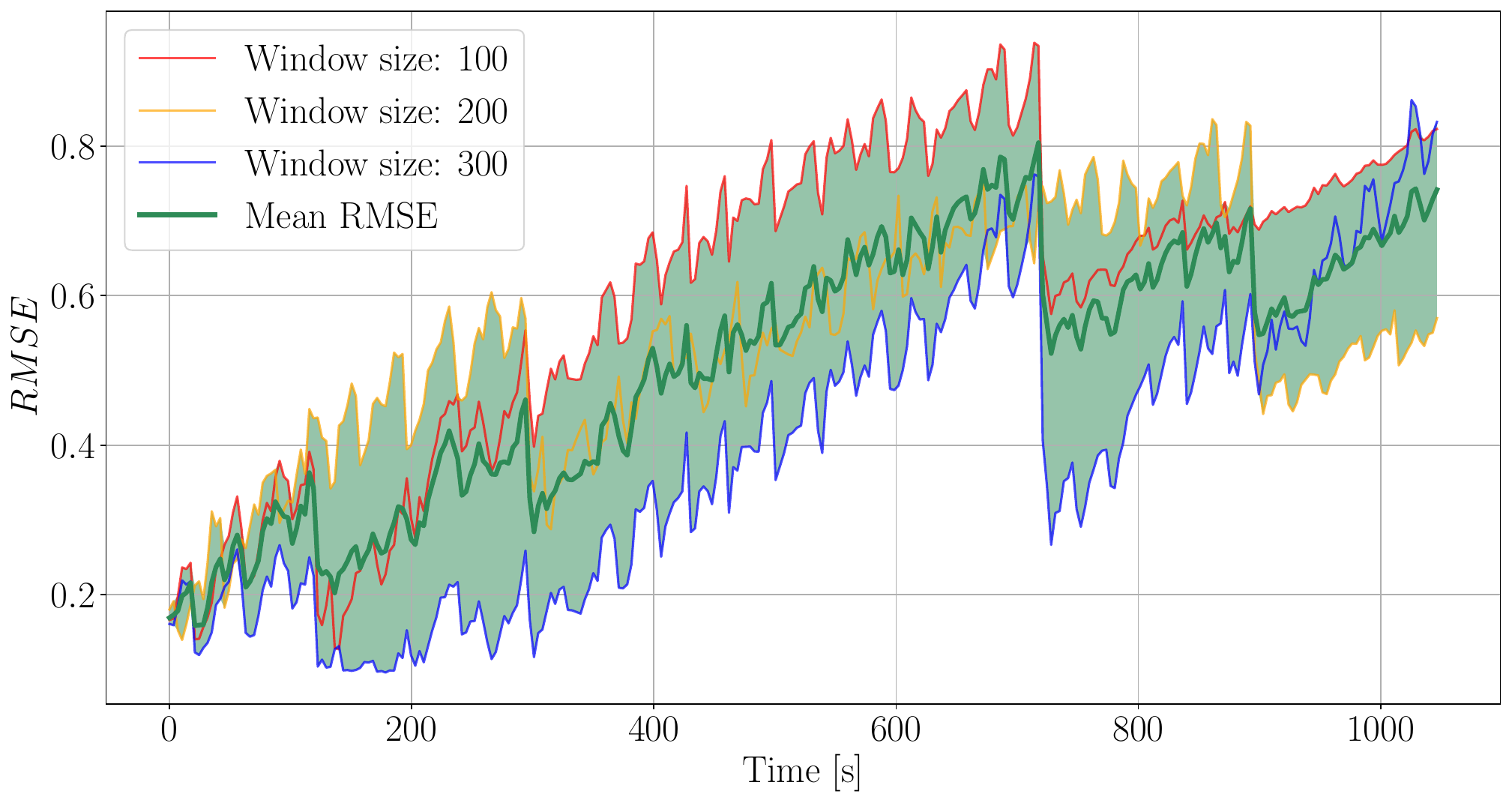}
    \caption{The root mean square error (RMSE) of the state predictions for the heating plate is presented, utilizing DMD with three different window sizes (100, 200, and 300) and a prediction horizon of 300 timesteps, equivalent to \unit[1050]{s}. In addition, the mean values and the corresponding distribution range are illustrated.}
    \label{fig:prediction_error}
    \end{figure}

\begin{figure}[htb!]
    \centering
    \includegraphics[width=\linewidth]{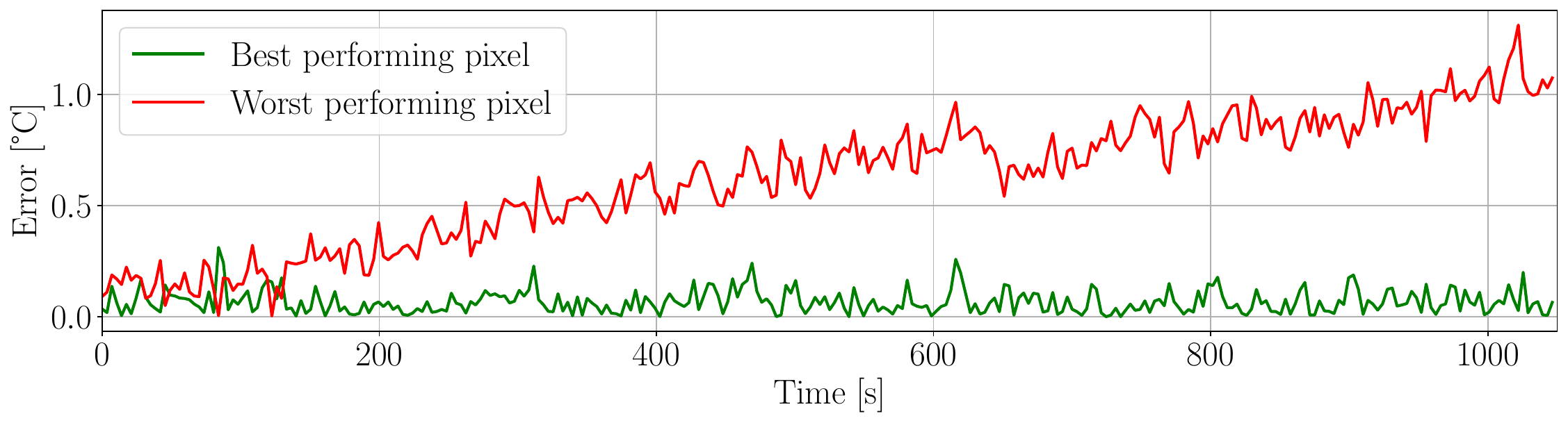}
    \caption{Prediction error of the best and worst performing pixel for a prediction horizon of 300 timesteps (\unit[1050]{s}) utilizing a window size of $w=200$. Considering the sensing range of \unit[-20]{°C} to \unit[150]{°C}, the relative prediction error for the worst-performing pixel in relation to the presented prediction horizon is approximately 0.6\%.}
    \label{fig:prediction_error_worst_pixel}
    \end{figure}

    Fig.~\ref{fig:prediction_error} presents the root mean square error (RMSE) of the state predictions for the heating plate using DMD with window sizes of 100, 200, and 300 over a prediction horizon of 300 timesteps (\unit[1050]{s}). In this context, the RMSE corresponds to the error in the image denoted by $\boldsymbol{e}$. The results show that the RMSE remains below 1, indicating accurate predictions within this horizon. However, increasing the window size or extending the prediction horizon introduces greater uncertainties, reflected by higher RMSE values.
    Since the performance of the individual pixels varies, in Fig.~\ref{fig:prediction_error_worst_pixel}, the prediction error of the best and worst performing pixel is shown.
    
\begin{figure*}[htb!]
\centering
    \begin{subfigure}[t]{0.19\linewidth}
    \includegraphics[width=\linewidth]{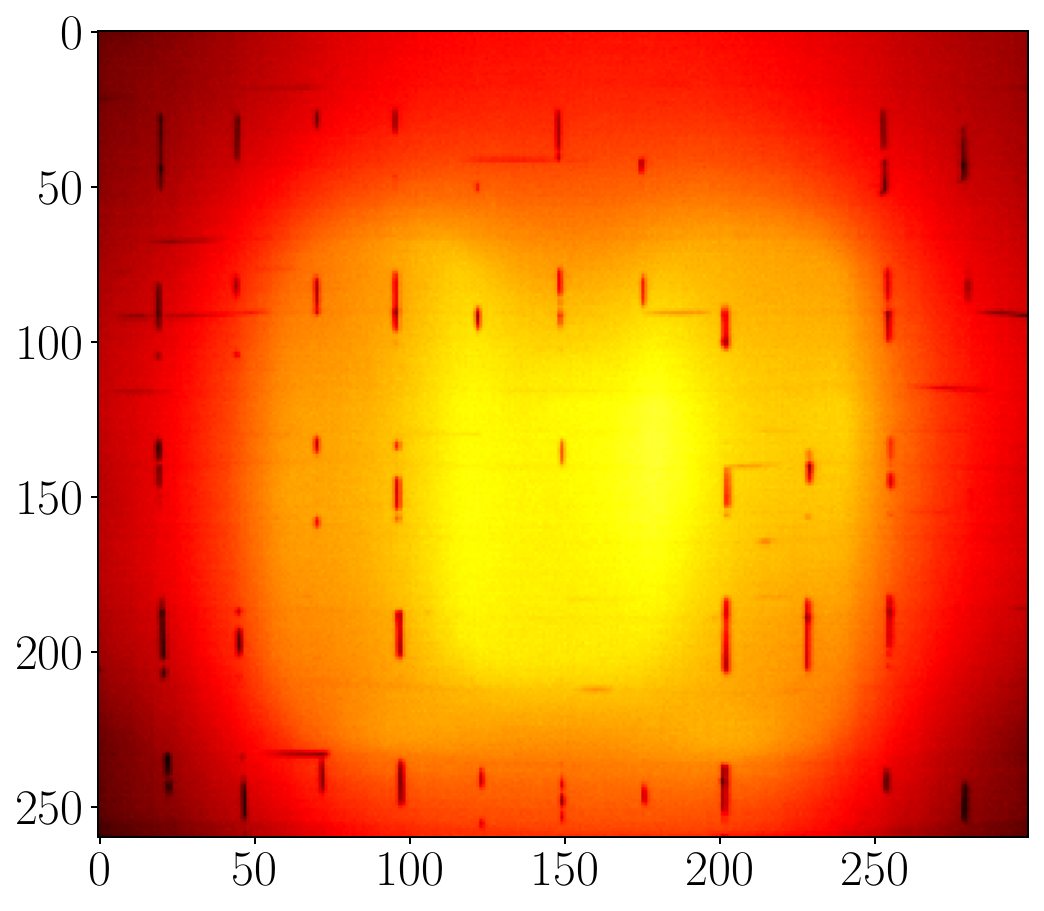} 
    \caption*{Ground truth (\unit[70]{s})}
    \end{subfigure}
    \begin{subfigure}[t]{0.19\linewidth}
    \includegraphics[width=\linewidth]{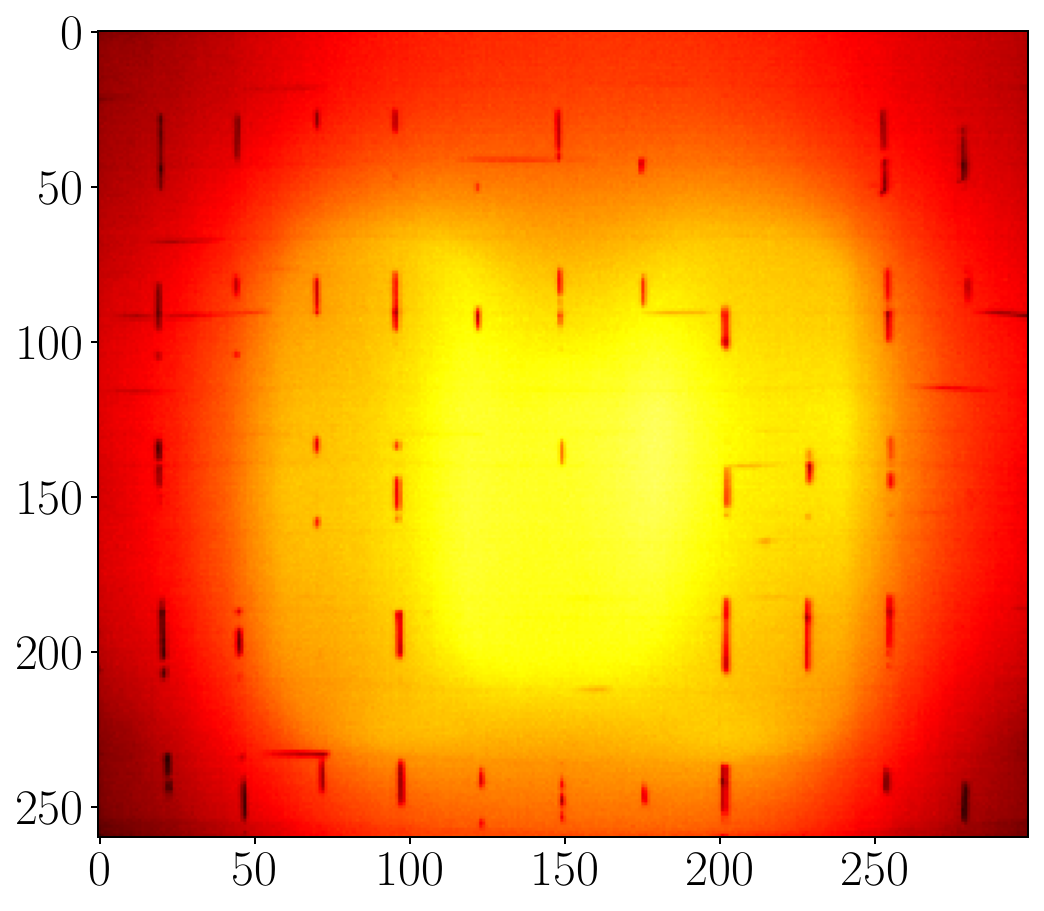}
    \caption*{Ground truth (\unit[140]{s})}
    \end{subfigure}
    \begin{subfigure}[t]{0.19\linewidth}
    \includegraphics[width=\linewidth]{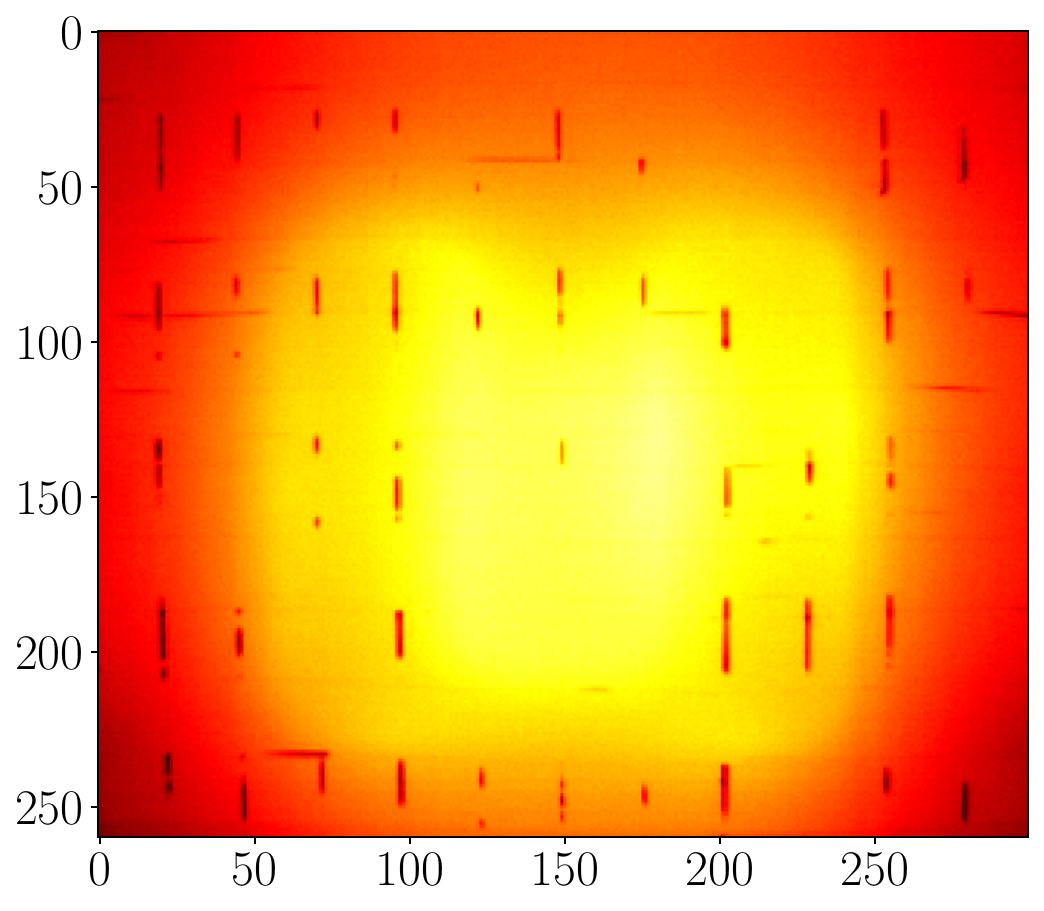} 
    \caption*{Ground truth (\unit[210]{s})}
    \end{subfigure}
    \begin{subfigure}[t]{0.19\linewidth}
    \includegraphics[width=\linewidth]{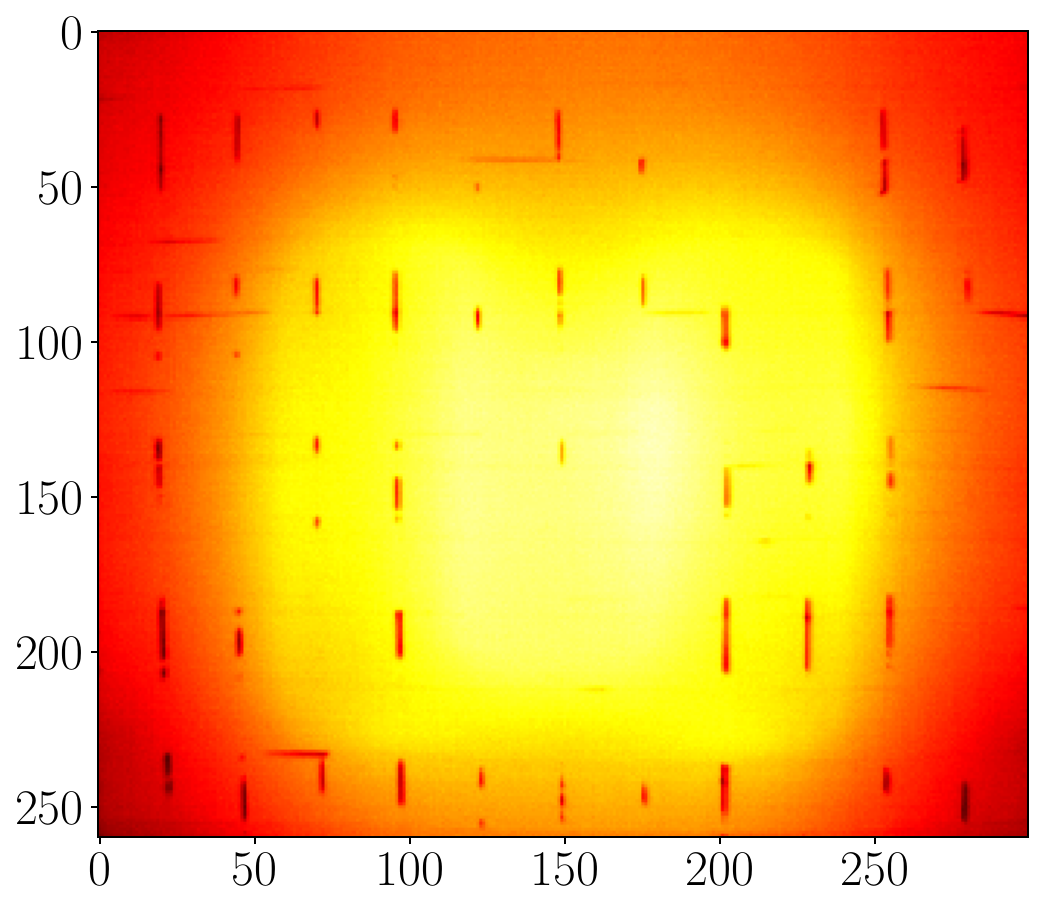}
    \caption*{Ground truth (\unit[280]{s})}
    \end{subfigure}
    \begin{subfigure}[t]{0.22\linewidth}
    \includegraphics[width=\linewidth]{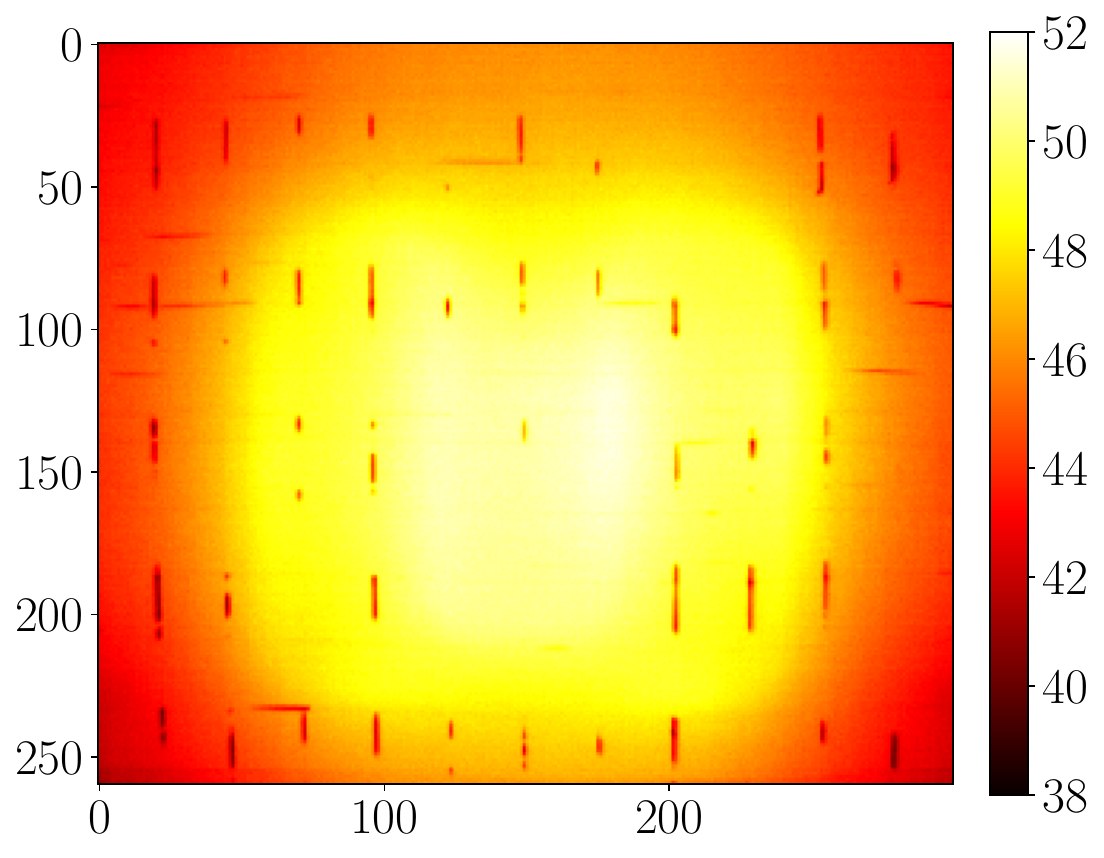} 
    \caption*{Ground truth (\unit[350]{s})}
    \end{subfigure}
    \\
    \begin{subfigure}[t]{0.19\linewidth}
    \includegraphics[width=\linewidth]{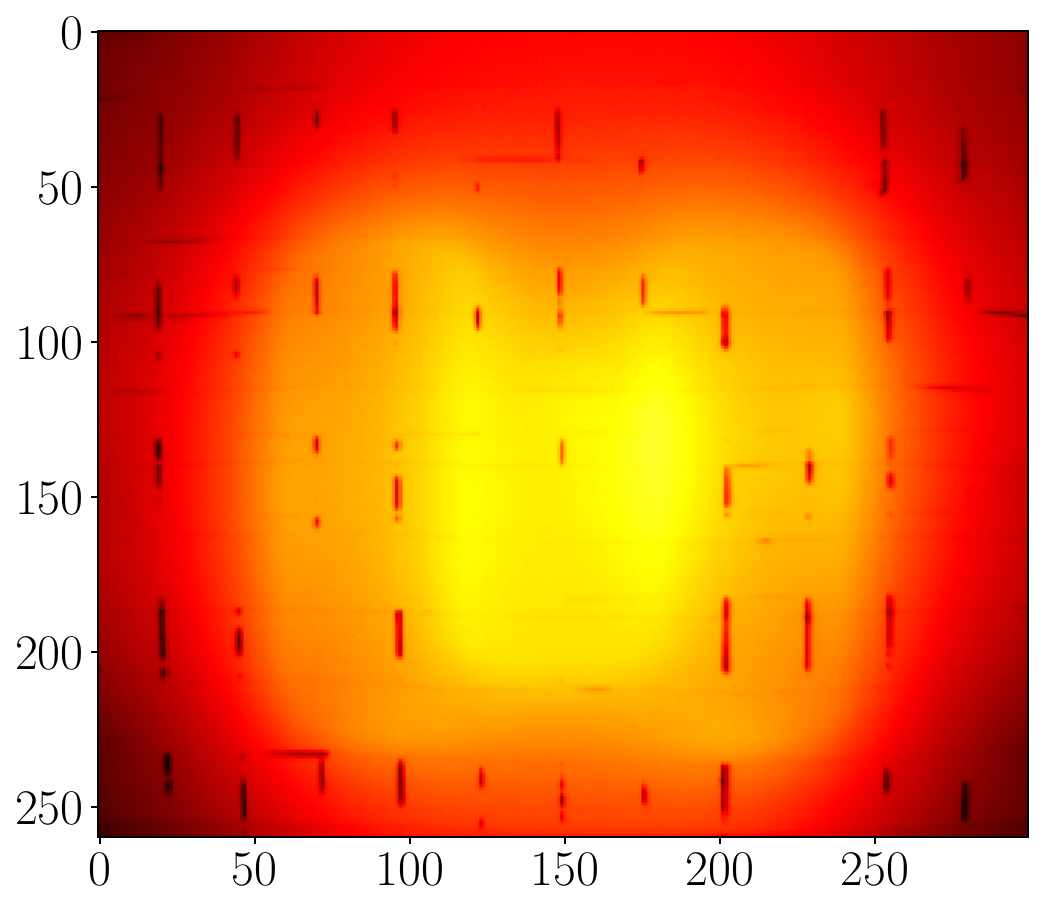}
    \caption*{Prediction (\unit[70]{s})}
    \end{subfigure}
    \begin{subfigure}[t]{0.19\linewidth}
    \includegraphics[width=\linewidth]{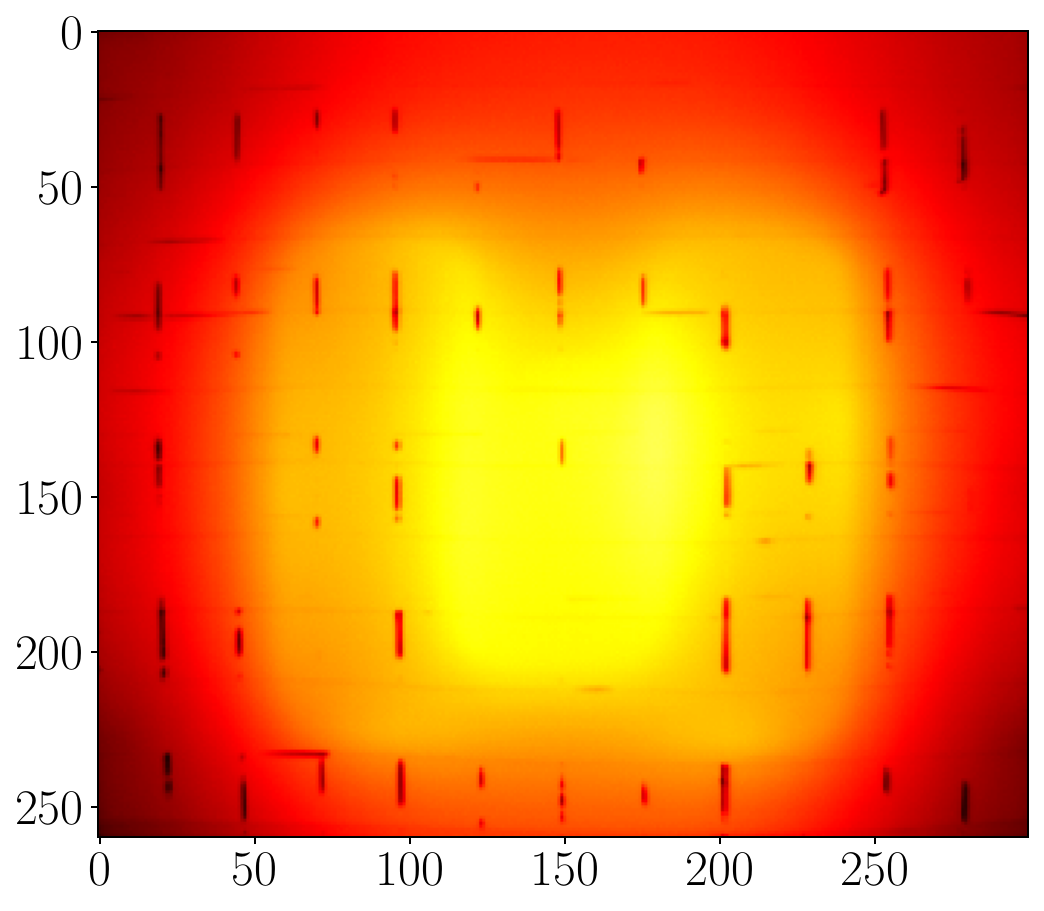} 
    \caption*{Prediction (\unit[140]{s})}
    \end{subfigure}
    \begin{subfigure}[t]{0.19\linewidth}
    \includegraphics[width=\linewidth]{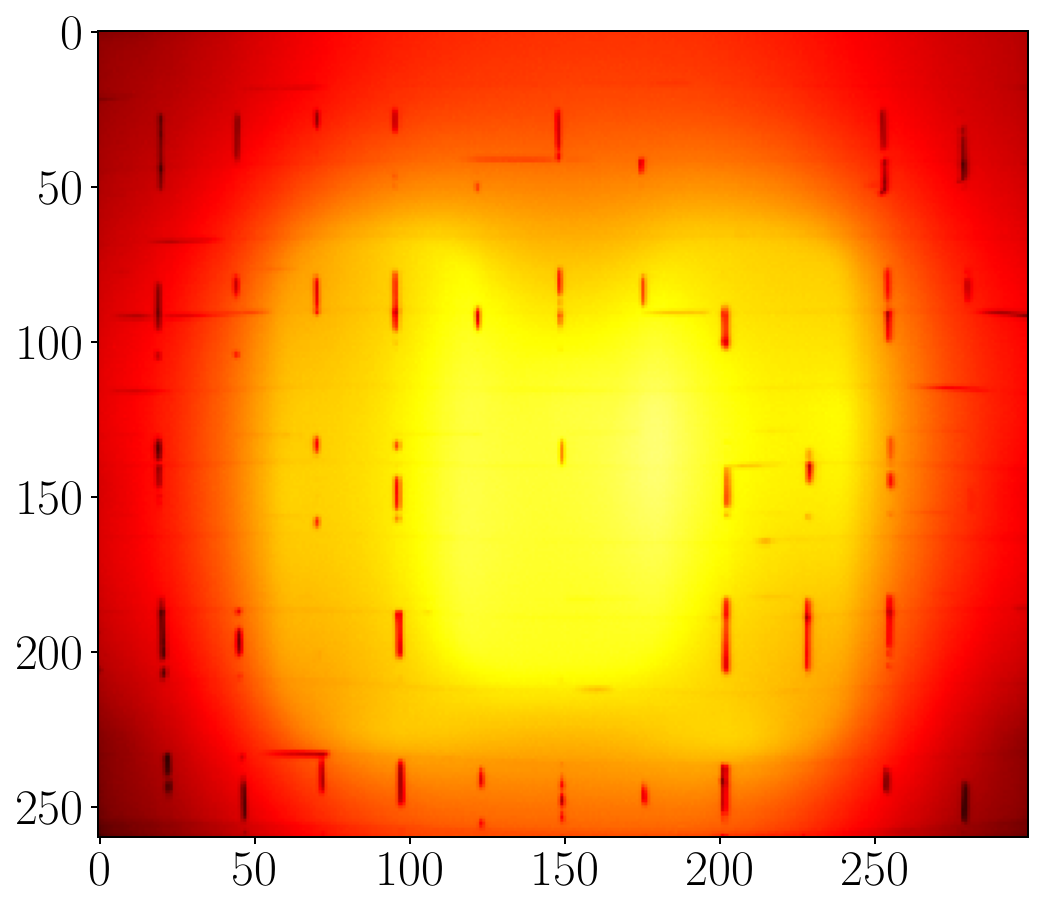}
    \caption*{Prediction (\unit[210]{s})}
    \end{subfigure}
    \begin{subfigure}[t]{0.19\linewidth}
    \includegraphics[width=\linewidth]{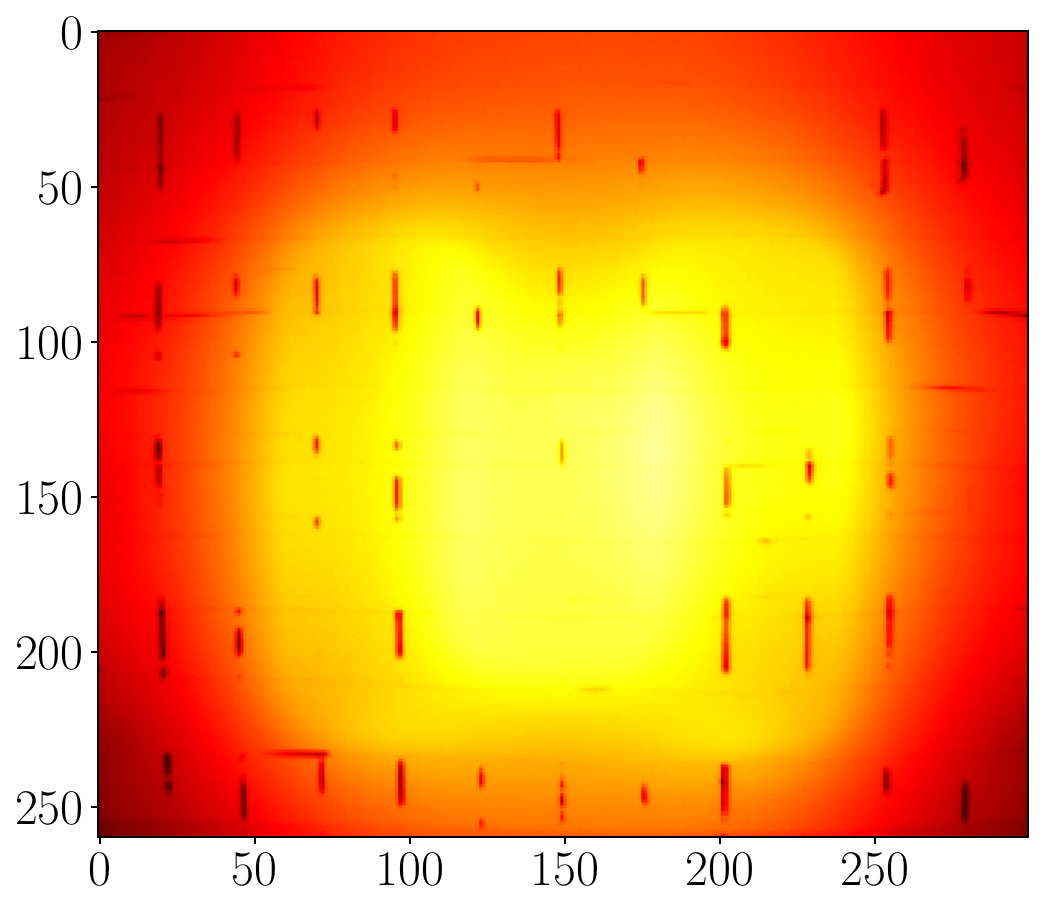}
    \caption*{Prediction (\unit[280]{s})}
    \end{subfigure}
    \begin{subfigure}[t]{0.22\linewidth}
    \includegraphics[width=\linewidth]{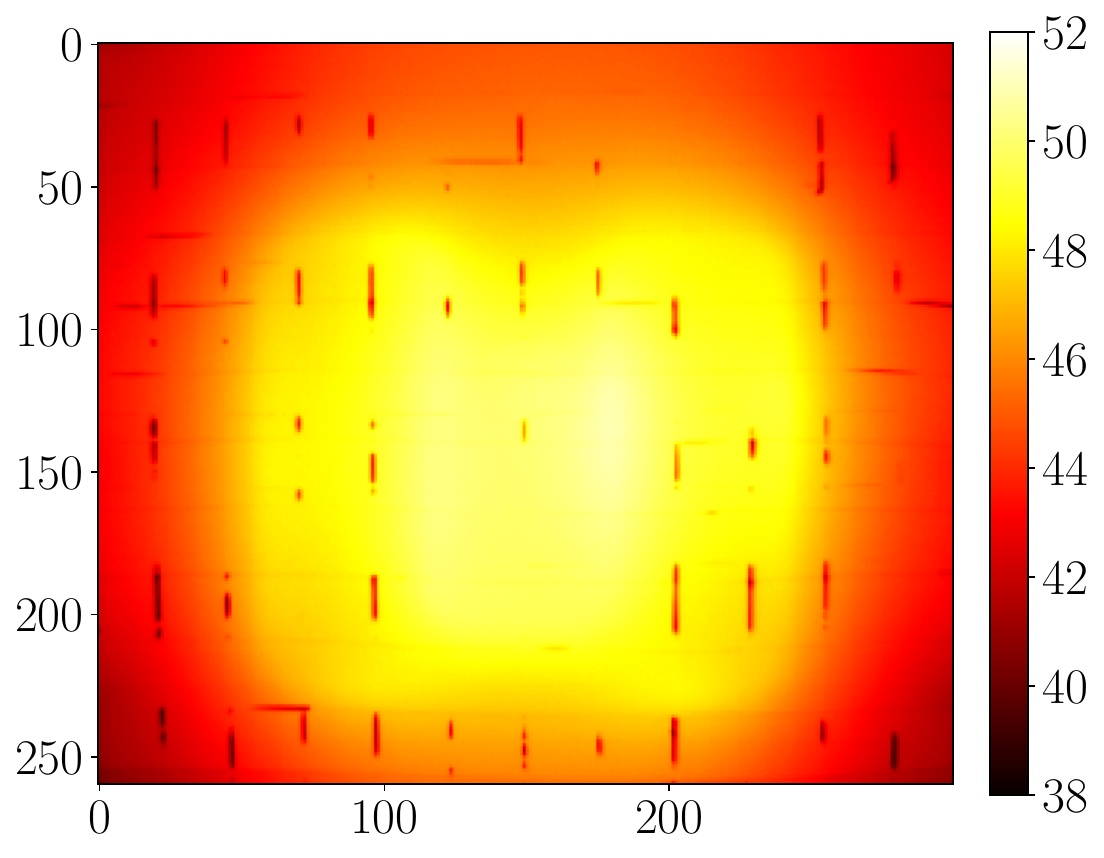}
    \caption*{Prediction (\unit[350]{s})}
    \end{subfigure}
    \\    
    \begin{subfigure}[t]{0.19\linewidth}
    \includegraphics[width=\linewidth]{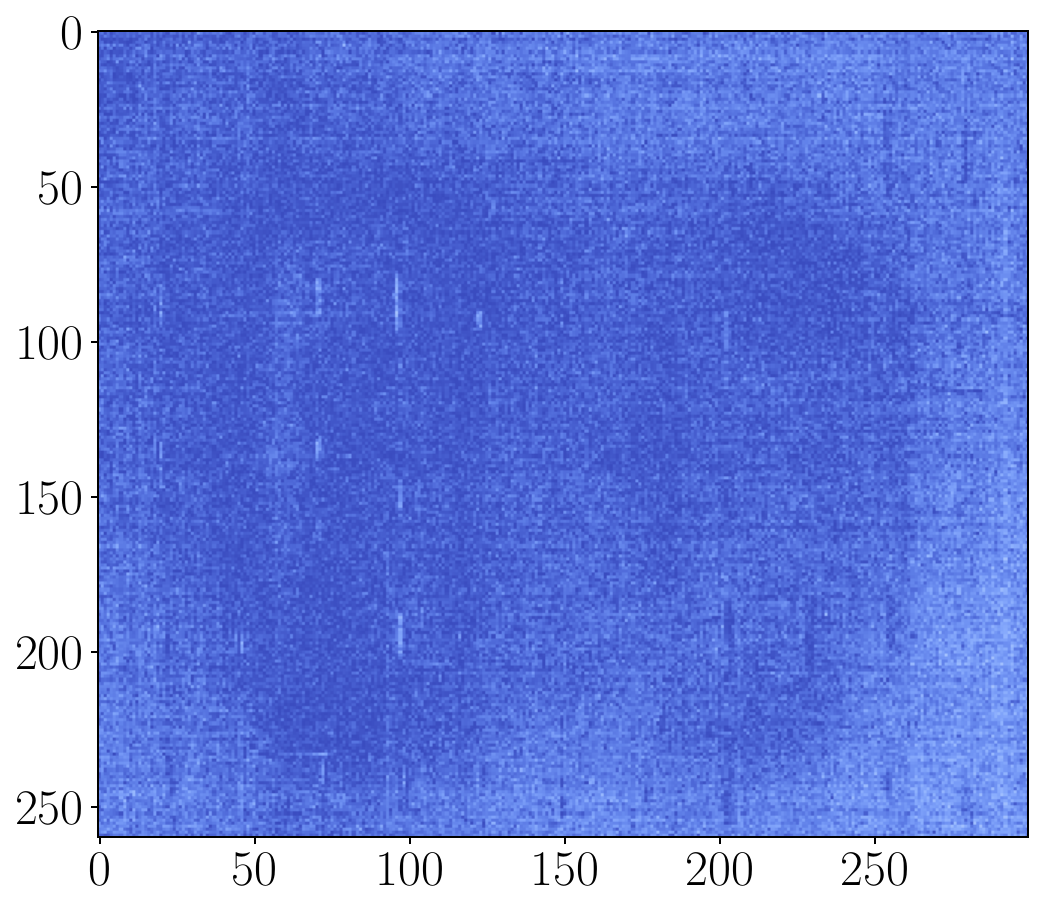} 
    \caption*{Error (\unit[70]{s})}
    \end{subfigure}
    \begin{subfigure}[t]{0.19\linewidth}
    \includegraphics[width=\linewidth]{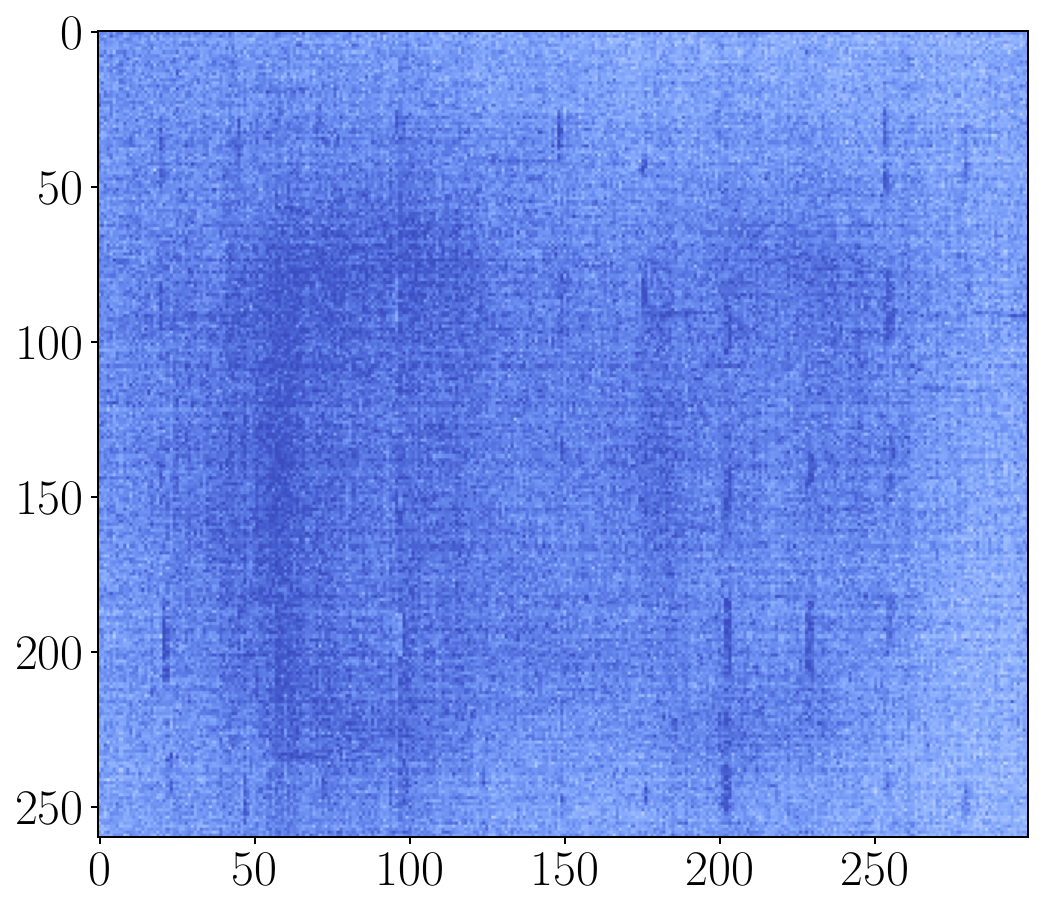} 
    \caption*{Error (\unit[140]{s})}
    \end{subfigure}
    \begin{subfigure}[t]{0.19\linewidth}
    \includegraphics[width=\linewidth]{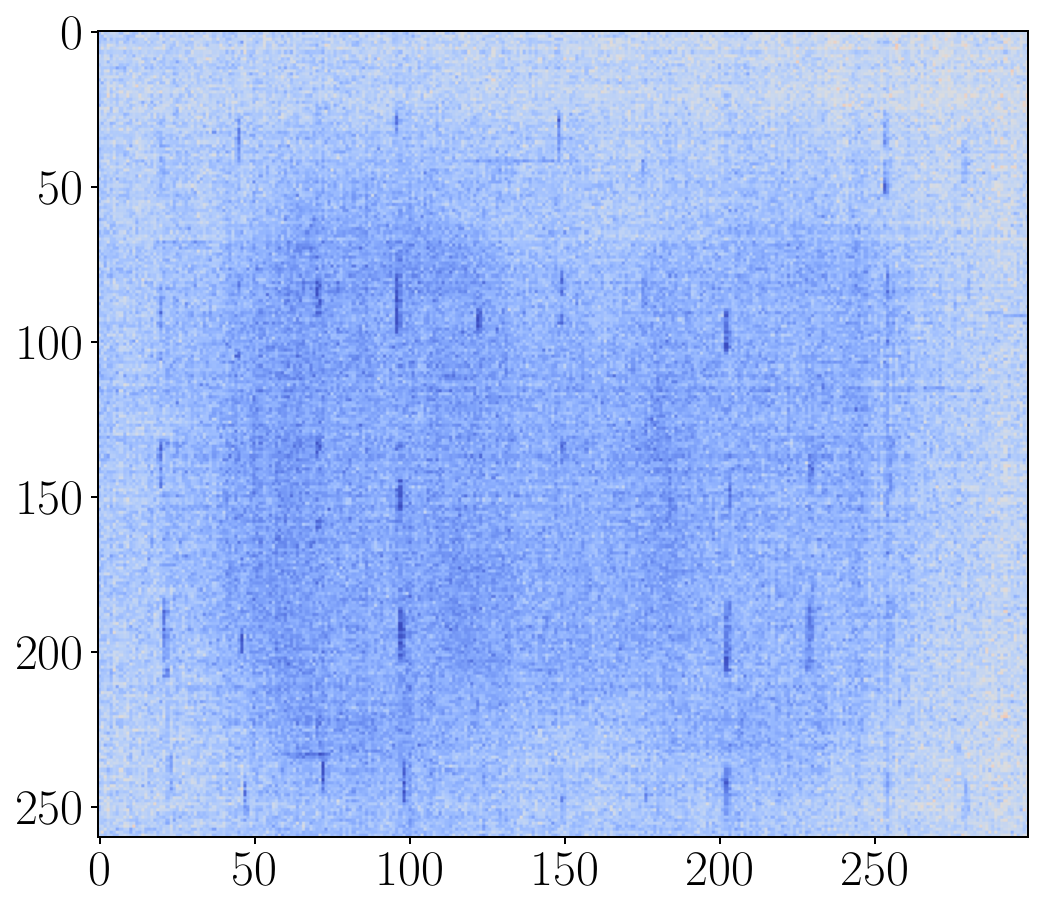} 
    \caption*{Error (\unit[210]{s})}
    \end{subfigure}
    \begin{subfigure}[t]{0.19\linewidth}
    \includegraphics[width=\linewidth]{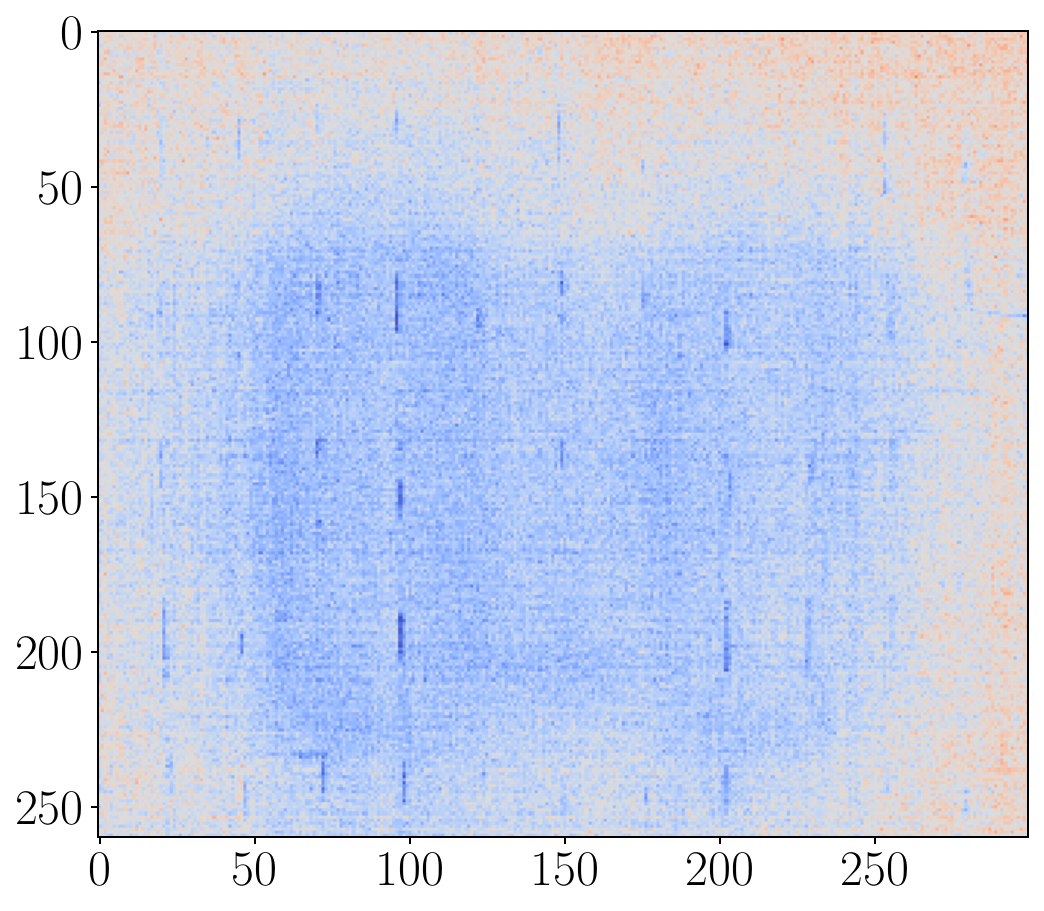} 
    \caption*{Error (\unit[280]{s})}
    \end{subfigure}
    \begin{subfigure}[t]{0.2215\linewidth}
    \includegraphics[width=\linewidth]{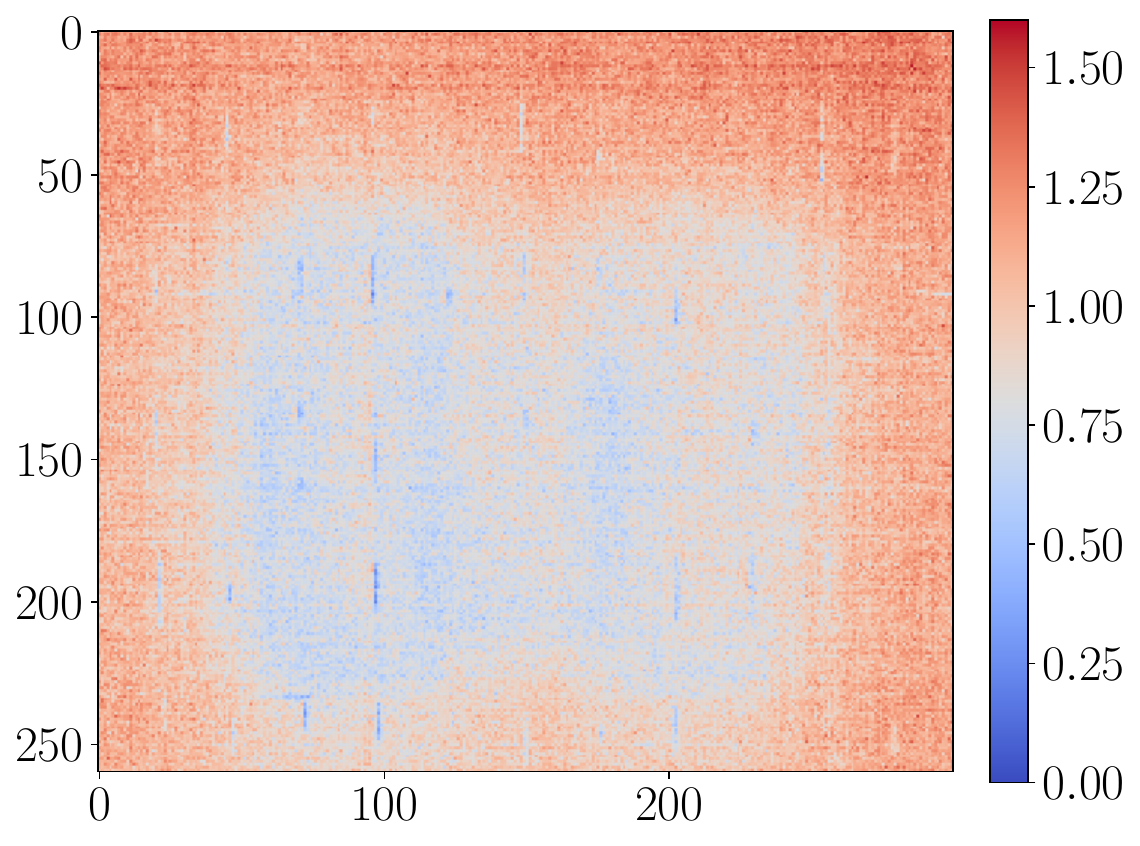} 
    \caption*{Error (\unit[350]{s})}
    \end{subfigure}
\caption{Prediction accuracy for various time horizons. The window size was chosen to be $w=100$. Snapshots of the ground truth future states (first row) and their predictions (second row) for 20 timesteps (\unit[70]{s}), 40 timesteps (\unit[140]{s}), 60 timesteps (\unit[210]{s}), 80 timesteps (\unit[280]{s}), and 100 timesteps (\unit[350]{s}) are demonstrated. The last row (third row) demonstrates the error evolving over time.}
\label{fig:prediction_timeseries}
\end{figure*}

    To demonstrate the performance of forecasting the thermal states within images, the predictions, the ground truth, and the related error images are visualized in Fig.~\ref{fig:prediction_timeseries}. Here, the window size was chosen to be $w=100$, and the predictions of 20 timesteps (\unit[70]{s}), 40 timesteps (\unit[140]{s}), 60 timesteps (\unit[210]{s}), 80 timesteps (\unit[280]{s}), and 100 timesteps (\unit[350]{s}) are presented. This demonstrates the evolution of prediction accuracy over increasing time horizons.

    \subsection{Predicting anomaly in real-time}
\begin{figure}[b!]
    \centering
    \includegraphics[width=\linewidth]{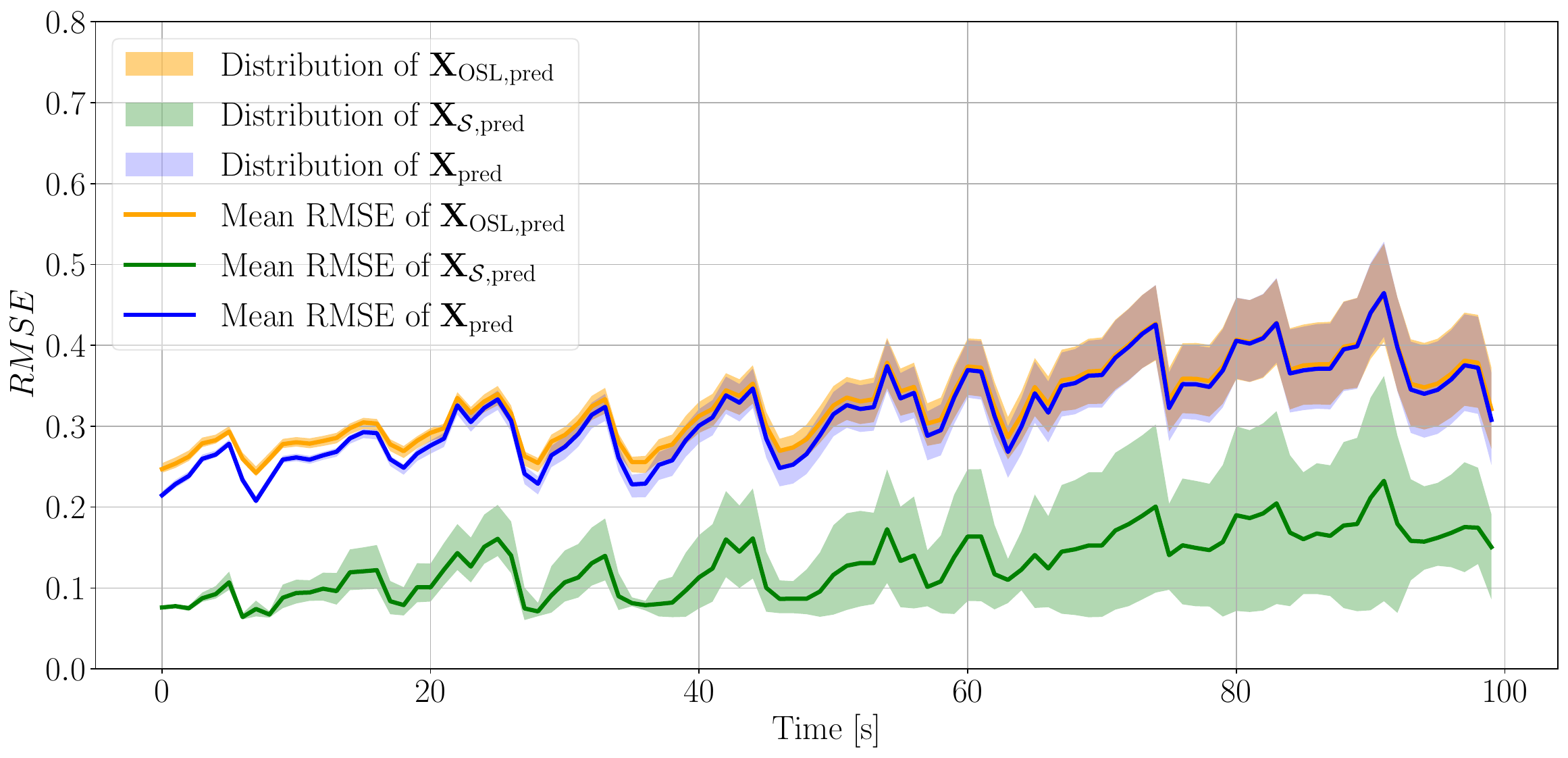}
    \caption{The root mean square error (RMSE) of the state predictions for the heating plate is presented, utilizing DMD with three different window sizes (20, 50, and 100) and a prediction horizon of 100 timesteps, equivalent to \unit[350]{s}. In addition, the mean values and the corresponding distribution range are illustrated.}
 \label{fig:RMSE_anomaly_predictions}
    \end{figure}

    Accurate state predictions are essential for prompt responses and interventions in real-time anomaly detection. Fig.~\ref{fig:RMSE_anomaly_predictions} illustrates the RMSE of the state predictions for the heating plate, utilizing DMD on both the time-varying coefficients obtained from OSL and the pixels identified as anomalies. Three different window sizes, 20, 50, and 100, were chosen to demonstrate the performance. The prediction horizon was set to 100 timesteps, corresponding to \unit[350]{s}.

    
    The results indicate that the predictions related to the anomalies are more accurate than the modeled time-varying parameters derived from the OSL. This is expected since the trained POD modes were unable to capture the behavior of these anomalies. Therefore, substituting the predicted anomaly values with their dedicated pixel positions, as described in Eq.~\eqref{eq:predictino_framework}, improves the overall accuracy of the state predictions.
    
    The figure shows the predictions adjusted by the anomaly predictions represented by the blue band. Although the improvement may not appear substantial when compared to the orange band representing the OSL predictions, this is attributed to the sparse nature of the anomaly vector, which has a minimal impact on the overall RMSE due to the high-dimensional data. However, the effectiveness of the anomaly predictions is clearly highlighted by the green band.
    
    \subsection{Visualization in virtual reality}
\begin{figure}
        \centering
        \includegraphics[width=\linewidth]{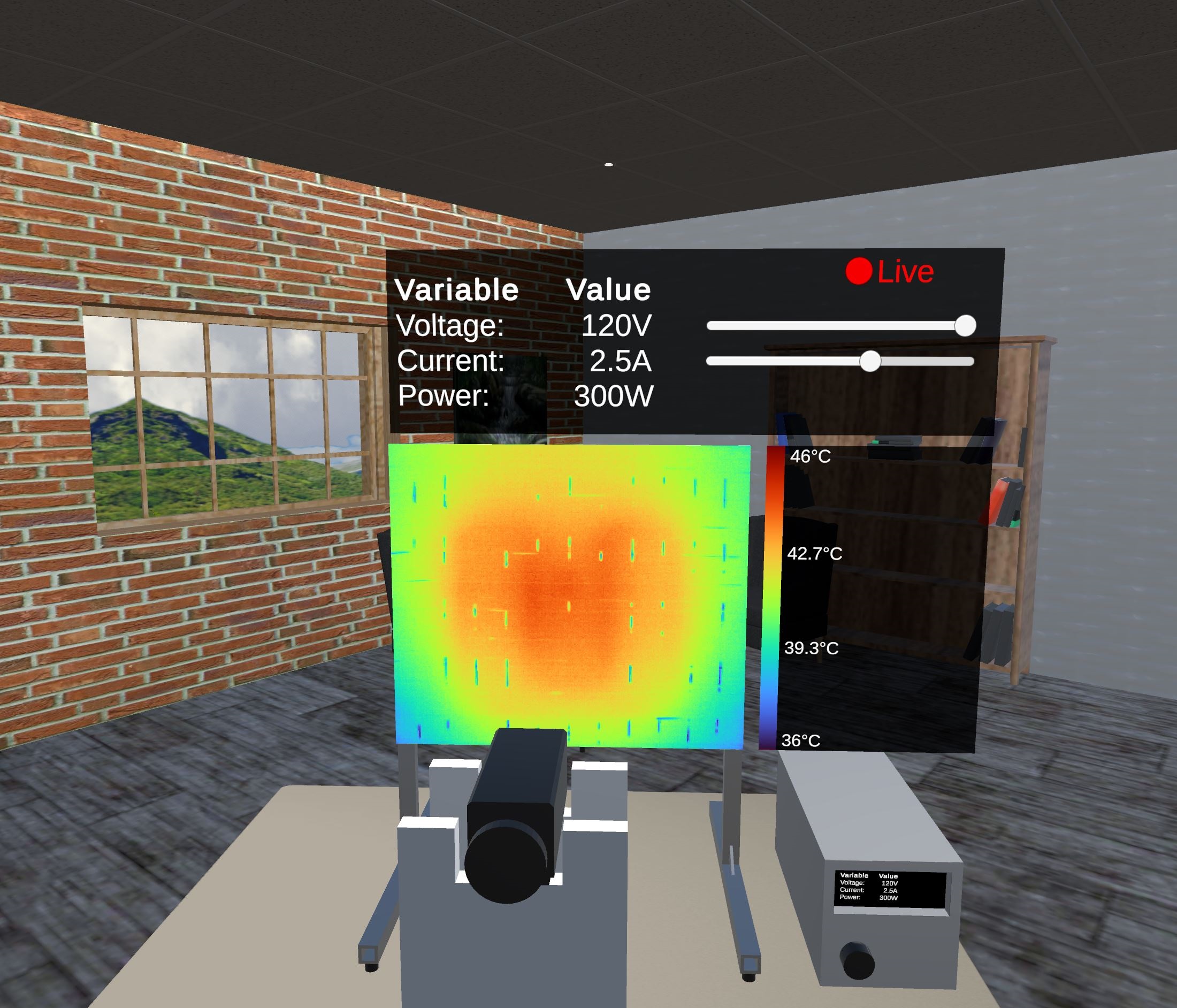}
        \caption{A screenshot of the virtual reality interface that features real-time monitoring and control of the experimental setup.}
        \label{fig:VR}
    \end{figure}

        The human-machine interfaces have been constructed according to the description in Section~\ref{ssec:interface} and have proven useful for remote operation not only throughout the project itself but also in communication with both technical and interdisciplinary colleagues. A screenshot of the virtual reality application is shown in Fig.~\ref{fig:VR}. The application can be directly deployed onto a virtual reality headset and does not require any additional hardware. It has been successfully used to intuitively communicate both the three-dimensional experimental setup and its operation, as well as the results. This approach has been especially valuable when communicating with partners unfamiliar with the physical setup.

\section{Conclusion, wider context, and future work}
\label{sec:conclusionandfuturework}
This work has successfully demonstrated the development and practical application of a diagnostic and predictive digital twin, employing advanced mathematical models and thermal imaging for effective condition monitoring. The key contributions of this work include the following.

\begin{itemize}
    \item \textit{Development of a Digital Twin-Ready Physical framework for condition monitoring:} We developed a comprehensive digital twin ready physical framework for condition monitoring, featuring key components such as sensors (thermocouples for calibration and thermal cameras), computing devices (microcontroller, PC, cloud computing), a switch, a remotely controllable variac, a heating coil, and a plate representing the monitored asset. This setup generates large volumes of high dimensional real-time data, which can be used for model validation and benchmarking, enhancing predictive maintenance and system diagnostics. The framework is adaptable and can be expanded to include various other types of sensors.
    \item \textit{Creation of a Real-Time Diagnostic and Predictive Digital Twin Framework:} We developed a digital twin of the physical setup with integrated diagnostic and predictive capabilities, fully synchronized with its physical counterpart through real-time data transfer and on-the-fly analysis. Our novel prediction and anomaly detection algorithms, based on Proper Orthogonal Decomposition (POD), Robust Principal Component Analysis (RPCA), and Dynamic Mode Decomposition (DMD), drive this process. The modular and extendable framework allows for simple adjustment of these algorithms, enabling the integration of other advanced methods, such as Autoencoders, Long Short-Term Memory (LSTM) networks, and Transformers, enabling the testing of new approaches for enhanced applicability.
    \item \textit{Establishment of an Intuitive Human-Machine Interface:} Although not the focus of this work, we also briefly demonstrated how such a physical framework can be operated remotely using the digital twin in virtual reality. This interface blurs the distinction between reality and the virtual world, creating a seamless experience that helps users engage with the digital twin in a more effective and immersive way. It also shields users from complex underlying details, distilling vast amounts of data into accessible, actionable insights. 
\end{itemize}

The methodologies and technologies developed here have significant potential for application across various high-stake industrial sectors. For instance, in autonomous shipping, this digital twin technology could be pivotal for monitoring critical components like engines or gearboxes, enhancing preventive maintenance, and ensuring operational safety. Similarly, in the wind energy sector, applying this technology to monitor gearbox conditions in wind turbines could drastically reduce downtime and maintenance costs. Additionally, in the metallurgical industry, this approach could revolutionize process monitoring, optimizing operations and enhancing quality control. Lastly, it could also be applied to monitor cables and electrical substations, improving reliability and performance in energy distribution networks. Moving forward, we aim to adapt and scale the digital twin framework to these diverse applications, driving further innovation in predictive maintenance and asset management. This expansion will leverage the predictive capabilities of our digital twin technology, potentially transforming how industries manage and maintain their critical systems and infrastructure.

\section*{Data availability}
Will be made available on request.

\section*{Declaration of competing interest}
    The authors declare that they have no known competing financial interests or personal relationships that could have appeared to influence the work reported in this paper.

\section*{Acknowledgements}    
    This publication has been prepared as part of NorthWind (Norwegian Research Centre on Wind Energy)\citep{FMENorthWindnrc} (project code 321954) and SFI AutoShip \citep{SFIAutoship}(project code 309230) which are co-financed by the Research Council of Norway, industry, and research partners. In addition we also acknowledge the European Union’s Horizon 2020 research and innovation PERSEUS doctoral program under the Marie Skłodowska-Curie grant agreement number 101034240. 

\bibliographystyle{elsarticle-harv} 
\bibliography{refs}


\end{document}